\documentclass[11pt]{article}

\usepackage[final]{acl}

\usepackage{times}
\usepackage{latexsym}

\usepackage[T1]{fontenc}
\usepackage[utf8]{inputenc}

\usepackage{microtype}
\usepackage{inconsolata}
\usepackage{graphicx}

\usepackage{booktabs}
\usepackage{array}
\usepackage{makecell}
\usepackage{tabularx}
\usepackage{xspace}
\usepackage{enumitem}
\setlist[itemize]{topsep=2pt,itemsep=1pt,parsep=0pt,partopsep=0pt}
\setlist[enumerate]{topsep=2pt,itemsep=1pt,parsep=0pt,partopsep=0pt}
\usepackage{amsmath}
\usepackage{amssymb}
\usepackage{xcolor}
\usepackage{fvextra}
\usepackage{placeins} 
\usepackage{float}    

\makeatletter
\let\acl@thebibliography\thebibliography
\renewcommand{\thebibliography}[1]{%
  \acl@thebibliography{#1}%
  \setlength{\itemsep}{3.5pt plus 1.5pt minus 1pt}%
  \setlength{\parsep}{3.5pt plus 1.5pt minus 1pt}%
}
\makeatother

\usepackage[most]{tcolorbox}

\hypersetup{
  colorlinks=true,
  linkcolor=msbluedark,
  citecolor=black,
  urlcolor=msbluedark,
  filecolor=msbluedark,
}

\definecolor{msblue}{HTML}{0078D4}
\definecolor{msbluelight}{HTML}{EAF3FB}
\definecolor{msbluedark}{HTML}{004578}

\newcommand{\affilpill}[1]{%
  \tikz[baseline=(X.base)]\node[
    fill=msbluelight,
    draw=msblue!70,
    line width=0.4pt,
    rounded corners=3pt,
    inner xsep=8pt,
    inner ysep=3pt,
    text=msbluedark,
  ] (X) {\textbf{#1}};%
}

\renewenvironment{abstract}{%
  \par\vspace{1.5ex}
  \begin{tcolorbox}[
    enhanced,
    colback=msbluelight,
    colframe=msblue!55,
    boxrule=0.5pt,
    arc=4pt,
    left=12pt, right=12pt, top=10pt, bottom=10pt,
    breakable,
  ]
  \noindent{\large\bfseries\color{msbluedark}Abstract}\par
  \vspace{4pt}\noindent\ignorespaces
}{%
  \end{tcolorbox}\par\vspace{1ex}
}

\newcommand{\paperinfo}{%
  \par\nointerlineskip\vspace{-0.5ex}
  \begin{tcolorbox}[
    enhanced,
    colback=msbluelight,
    colframe=msblue!55,
    boxrule=0.5pt,
    arc=4pt,
    left=12pt, right=12pt, top=7pt, bottom=7pt,
    breakable,
  ]
  \footnotesize\raggedright\setlength{\parskip}{2pt}\setlength{\parindent}{0pt}%
  \def\UrlBreaks{\do\/\do\.\do\-\do\_\do\=\do\?\do\&}%
\textbf{\color{msbluedark}Date:}~May 2026\\
  \textbf{\color{msbluedark}Data:}~\href{https://huggingface.co/datasets/microsoft/OfficeComprehensionBenchmark}{\nolinkurl{huggingface.co/datasets/microsoft/OfficeComprehensionBenchmark}}\\
  \textbf{\color{msbluedark}Code:}~\href{https://github.com/microsoft/OfficeComprehensionBench/}{\nolinkurl{github.com/microsoft/OfficeComprehensionBench}}\\
  \end{tcolorbox}
  \par\vspace{1ex}
}

\DefineVerbatimEnvironment{PromptBlock}{Verbatim}{%
  fontsize=\scriptsize,
  breaklines=true,
  breakanywhere=true,
  frame=single,
  framesep=2mm,
  framerule=0.4pt,
  rulecolor=\color{black!30}%
}

\title{Office Comprehension Benchmark}

\author{%
  \textbf{Firoz Shaik},
  \textbf{Mateus Picanço Lima Gomes}\textsuperscript{$\dagger$},
  \textbf{Tanvir Aumi},
  \textbf{Jingci Wang},
  \textbf{Milos Milunovic}
  \\
  \textbf{Filip Basara},
  \textbf{Ivana Jovanovic},
  \textbf{Jay Rathi},
  \textbf{Vishwas Suryanarayanan},
  \textbf{Neha Nandan Kenkare},
  \textbf{Weiyao Xie}
  \\
  \textbf{Zhipeng Han},
  \textbf{Zheng Zhang},
  \textbf{Waleed Shahid},
  \textbf{Russell Scherer}
  \textbf{Thong Q. Nguyen}\textsuperscript{$\dagger$},
  \textbf{Michael Bentley}
  \\
  \textbf{Tamara Stankovic},
  \textbf{Rasika Chakravarthy},
  \textbf{Vishal Chowdhary}
  \\[10pt]
  \affilpill{Microsoft}
  \\[6pt]
  {\footnotesize\textsuperscript{$\dagger$}\,Work done while at Microsoft.} \\[2pt]
  {\small\textbf{Correspondence:}
   \href{mailto:firozshaik@microsoft.com}{\texttt{firozshaik@microsoft.com}}}
}

\begin{document}
\maketitle
%

\newcommand{\OCB}{OCB\xspace}

\begin{abstract}
We introduce \textbf{Office Comprehension Bench (\OCB)}, the first
public benchmark to jointly evaluate LLM systems on Word, Excel, and
PowerPoint comprehension over native file formats (\texttt{.docx},
\texttt{.xlsx}, \texttt{.pptx}) and their variants. \OCB{} consists of two tracks.
\textbf{File Fidelity Q\&A} tests structural and visual perception of
office artifacts - tables, charts, embedded images, formulas, and
app-specific elements such as headers, speaker notes, and named
ranges. \textbf{Domain Q\&A} tests expert-level reasoning grounded in
real-world industry documents across 12 professional domains, with
queries requiring multi-step analysis and synthesis across documents.
Each reference answer is decomposed into atomic, binary-gradable
claims, and an ensemble of LLM judges scores responses against each
claim independently. Even the strongest frontier system in its default reasoning mode
reaches only $\sim$59.3\% on Domain Q\&A; increasing thinking depth
within a tier does not move performance materially, while moving to
a higher product tier yields modest gains. We release the dataset,
evaluation tooling, judge prompt, and a public leaderboard.
\end{abstract}

\paperinfo

\section{Introduction}

Office documents are a primary medium for knowledge work, combining free-form text, structured tables, embedded images, charts, and application-specific constructs (e.g., slide layouts, document styles, and spreadsheets with formulas). Despite rapid progress in large language models (LLMs), rigorous
evaluation of office comprehension remains limited. Office tasks often require grounded reading over a file, multi-step reasoning over heterogeneous structure, and adherence to precise output formats. We introduce \textbf{Office Comprehension Bench (\OCB)}, a benchmark suite covering Word, Excel, and PowerPoint comprehension over \textbf{native file formats} (\texttt{.docx}, \texttt{.xlsx}, \texttt{.pptx}) and their variants. To our knowledge, \OCB{} is the first public benchmark to jointly evaluate comprehension over all three Office
applications in their native file formats under a unified evaluation
methodology. \OCB{} consists of two complementary evaluation tracks:

\begin{itemize}
\item \textbf{File Fidelity Q\&A} - 240 files, 902 queries, and 1{,}305
atomic assertions evaluating whether a system can accurately perceive
structural and visual content (tables, charts, images, slide layouts,
formatting).
\item \textbf{Domain Q\&A} - 64 files, 120 queries, and 5{,}414 atomic
assertions evaluating expert-level comprehension across 12 industries
(finance, healthcare, manufacturing, accounting, energy, government,
education, and others). Domain queries require multi-step domain reasoning,
including synthesis that can span multiple documents; a single response
may produce dozens of independently verifiable claims.
\end{itemize}

\OCB{} is comprehension-only by design and is complementary to agentic
benchmarks that grade automation workflows or work-product creation
(e.g., OfficeBench~\citep{wang2024officebench},
OdysseyBench~\citep{wang2025odysseybench},
GDPval~\citep{patwardhan2025gdpval}), since reliable comprehension is
a prerequisite for trustworthy automation and authoring.

We release the dataset, evaluation tooling, an official LLM-as-a-judge
accuracy prompt, and a public leaderboard reporting results for OpenAI
GPT-5.5 Thinking, Anthropic Claude Opus 4.7, and Google Gemini 3.1 Pro.
The OCB dataset is released under the Community Data License Agreement -
Permissive 2.0 (CDLA-Permissive-2.0); hosted third-party reference files
retain their original licenses, with per-source details in the dataset
card. The accompanying code is released under the MIT License.

All systems are evaluated through their public web chat interfaces
rather than via API, because that is the deployment surface where real
users interact with Office files. \OCB{} is therefore an
\emph{end-to-end evaluation of deployed products}: the unit of
evaluation we care about is the deployed system, which jointly handles
native file parsing \emph{and} downstream reasoning, not the underlying
language model in isolation.
Reported scores therefore reflect the deployed system (including
provider-side scaffolding such as system prompts and tool routing)
during the evaluation window. For models that do not provide their own
file-handling scaffolding (e.g., open-weight models accessed directly),
we additionally release a shared harness using Apache Tika for
open-source file extraction, so that such models can be evaluated on
\OCB{} over a common input representation.

\paragraph{Contributions.}
\begin{itemize}
\item \textbf{The first public benchmark for native Office comprehension.}
\OCB{} jointly covers Word, Excel, and PowerPoint in their native file
formats (\texttt{.docx}, \texttt{.xlsx}, \texttt{.pptx}) and their variants under a unified
two-track design: a \emph{File Fidelity} track with an object-centric
artifact taxonomy ($\sim 1.4$ assertions/query) for structural and
visual perception, and a \emph{Domain Q\&A} track ($\sim 45$
assertions/query) for expert-level analytical reasoning across 12
industries. Existing benchmarks address narrower slices: document
images (DocVQA~\citep{mathew2021docvqa}), rendered slides
(SlideVQA~\citep{tanaka2023slidevqa}), charts
(ChartQA~\citep{masry2022chartqa}), PDFs
(MMLongBench-Doc~\citep{ma2024mmlongbenchdoc},
GDP.PDF~\citep{garre2026gdppdfbenchmarkinggroundedmultimodal},
OfficeQA Pro~\citep{databricks2026officeqapro}),
spreadsheet manipulation (SpreadsheetBench~\citep{ma2024spreadsheetbench}),
or work-product generation (GDPval~\citep{patwardhan2025gdpval}).

\item \textbf{A scalable methodology for open-ended evaluation.}
Reference answers are decomposed into atomic, binary-gradable claims
authored and reviewed by domain experts, then
scored by a three-judge LLM majority vote. We also report empirical
learnings on \emph{what makes assertions LLM-evaluable}: human
authors reliably produce correct answers, but require iteration to
produce atomic, unambiguous, non-redundant claims. We contribute a
failure-mode taxonomy of recurring rubric-quality issues (duplicate,
compound, ambiguous, overly broad, structural, and prompt-restating
assertions), with worked examples and detection signals in
Appendix~\ref{app:rubric-process}.

\item \textbf{Empirical findings on frontier-system limits.}
We evaluate Claude 4.7 Opus, GPT-5.5 Thinking, and Gemini 3.1 Pro on a
public leaderboard. Frontier systems plateau at $\sim$59.3\% on
Domain Q\&A in their default reasoning modes. Within a model family,
increasing thinking depth alone does not move performance materially;
moving to a higher capability tier (e.g., GPT-5.5 Pro) does yield
gains, though even the strongest configuration we evaluate
(GPT-5.5 Pro-Extended) reaches only $\sim$63\%. Per-application and
per-industry breakdowns reveal specific failure
modes, e.g., mixed Word$+$Excel inputs and reasoning over Supply
Chain, Corporate Governance, and Accounting documents are
systematically the hardest settings.
\end{itemize}

\section{Benchmark Overview}
\label{sec:overview}

\OCB{} is structured as two complementary tracks: \textbf{File Fidelity Q\&A}, which tests whether a system can accurately perceive structural and visual content of Office files; and \textbf{Domain Q\&A}, which tests expert-level comprehension and reasoning grounded in real-world industry documents.
Together, the two tracks span the spectrum from fine-grained file-format
understanding to multi-step professional analysis. All evaluations
assume LLM systems are given the full file as input rather than only
extracted text, so they are responsible for parsing the file and
reading both textual content and visual layout.

\subsection{Dataset Composition}

Table~\ref{tab:dataset} summarizes the scale of each track.

\begin{table}[t]
\centering
\scriptsize
\setlength{\tabcolsep}{3pt}
\caption{Dataset scale by track and application.}
\label{tab:dataset}
\begin{tabular}{@{}llrrrr@{}}
\toprule
\textbf{Track} & \textbf{App} & \textbf{Files} & \textbf{Q} & \textbf{Asn.} & \textbf{Asn./Q} \\
\midrule
File Fidelity & Word       & 78  & 383 & 492   & 1.28 \\
File Fidelity & Excel      & 116 & 281 & 508   & 1.81 \\
File Fidelity & PowerPoint & 46  & 238 & 305   & 1.28 \\
File Fidelity & Total      & 240 & 902 & 1{,}305 & 1.45 \\
\midrule
Domain Q\&A   & All apps   & 64  & 120 & 5{,}414 & 45.12 \\
\bottomrule
\end{tabular}
\end{table}

The tracks differ substantially in evaluation density: File Fidelity
queries are scoped to a single artifact (1--2 atomic assertions each),
while Domain Q\&A queries are open-ended professional analyses where a
single response may produce dozens of independently verifiable claims.
The $120$ Domain Q\&A queries split across input file types as:
Word ($35$), Excel ($29$), PowerPoint ($41$), Word$+$Excel ($8$),
and Word$+$PowerPoint ($7$).
Appendix~\ref{app:examples} provides one worked example from each track.

\subsection{File Fidelity Artifact Taxonomy}
\label{sec:taxonomy}

For the File Fidelity track, we use an object-centric taxonomy that
labels each query by the document artifact it targets. High-level
categories include text (paragraphs, bullets, headings, notes), images,
tables, charts (plots, legends, axes, embedded tables), layout
(slide/page structure, spatial grouping), and Excel formulas with
cross-cell/sheet dependencies. Each query is labeled with one or more
categories; the full per-application taxonomy is in
Appendix~\ref{app:taxonomy}.

\subsection{Domain Coverage}
\label{sec:domain-coverage}
The Domain Q\&A track currently spans 12 industries with queries
distributed fairly evenly (per-industry shares range from $5.0\%$ to
$12.5\%$ of the $120$ queries; see
Appendix~\ref{app:domain-distribution}, Figure~\ref{fig:domain-qna-industry}).
Coverage is roughly balanced rather than weighted by economic
contribution; weighting by GDP contribution is left to future work.

\section{Construction Methodology}
\label{sec:construction}

Both \OCB{} tracks are built with parallel four-stage pipelines:
source curation, question authoring, rubric construction, and
quality assurance. Each pipeline combines human experts with LLM
assistance and ends in human-adjudicated review. The tracks differ
in source-document strategy, who originates questions, the
granularity of reference assertions, and the QA loop.
Table~\ref{tab:methodology-summary} summarizes the contrast; the
full per-track pipelines are in Appendix~\ref{app:construction}.

\begin{table*}[t]
\centering
\small
\setlength{\tabcolsep}{5pt}
\renewcommand{\arraystretch}{1.15}
\caption{Construction methodology at a glance. See
Appendix~\ref{app:construction} for the full per-track pipelines.}
\label{tab:methodology-summary}
\begin{tabularx}{\textwidth}{@{}p{2.2cm}XX@{}}
\toprule
& \textbf{Domain Q\&A} & \textbf{File Fidelity Q\&A} \\
\midrule
Source documents & Real industry artifacts (SEC filings, slide decks, spreadsheets, ops docs); curated by experts & Broad public pool (Fortune 500 filings, government, educational); filtered + coverage-targeted across taxonomy \\
Question authoring & Expert-drafted from curated sources & LLM-generated with artifact-targeted prompts, human-curated/edited \\
Scale & 64 files, 120 queries, 5{,}414 assertions ($\sim 45$/query) & 240 files, 902 queries, 1{,}305 assertions ($\sim 1.4$/query) \\
Rubric construction & Atomic assertions via 3 authorship variants (independent, derived, cross-org) & Atomic assertions derived from the reference answer \\
Quality assurance & Independent reviewer distinct from prompt and rubric authors & Two-annotator review with third-annotator adjudication; groundedness filters \\
\bottomrule
\end{tabularx}
\end{table*}

\paragraph{Track-specific construction.}
\label{sec:domainqa}
\label{sec:filefidelity}
\label{sec:assertionquality}
The track-specific differences captured in
Table~\ref{tab:methodology-summary} stem from different starting
points. Domain Q\&A begins with domain experts
(Appendix~\ref{app:experts}) curating authoritative source documents
and drafting prompts that a peer practitioner would recognize as
realistic. Rubrics are then produced under three authorship variants
(independent of the golden answer, derived from it, and
cross-organization layered; Appendix~\ref{app:methodologies}) to
reduce dependence on any single rubric-construction convention.
File Fidelity Q\&A begins with a broad public file pool, filtered
iteratively to cover the per-app artifact taxonomy
(Appendix~\ref{app:taxonomy}). Candidate questions are LLM-generated
and human-curated, then verified by automatic groundedness filters
and two-annotator review. Both tracks share the same scoring
metric (assertion-level accuracy; \S\ref{sec:eval}). Each assertion
carries an expert-assigned importance weight used only as a secondary
signal (Appendix~\ref{app:weighted}). Across both tracks we find that
human authors produce strong questions and answers but require
iteration to produce \emph{atomic, LLM-evaluable} assertions; the full
failure-mode taxonomy and revision process are in
Appendix~\ref{app:rubric-process}.

\section{Evaluation Framework}
\label{sec:eval}

\subsection{Evaluation Setup}

\OCB{} evaluates whether a system response is factually correct,
complete, and grounded in the provided Office documents. Each query is
associated with a set of atomic assertions derived from the reference
answer; every assertion represents a specific, verifiable claim such
as retrieving a value, interpreting a chart element, or computing a
quantity from structured data. The same metric applies identically to
both tracks.\footnote{Subject LLM systems are queried through their public web chat interfaces rather than via API, so that evaluation jointly stresses native file parsing and end-to-end reasoning. \textbf{Scrapes were collected between April 2, 2026 and May 18, 2026; reported results therefore reflect the state of each system's deployed web chat interface during that window.} GPT-5.5 Thinking and Claude Opus 4.7 use their Standard thinking setting; Gemini 3.1 Pro has no analogous setting. Models are instructed not to perform web search or ask clarifying questions. A small set of queries that include supporting URLs and explicitly require web retrieval are exempted from this. Claude Opus 4.7 and GPT-5.5 Thinking are also instructed to reply in chat, since both otherwise tend to return downloadable \texttt{.xlsx} or \texttt{.docx} artifacts for analytical queries.}

Let $q$ index queries and $i$ index assertions within a query. Each
assertion receives a binary score $y_{q,i} \in \{0,1\}$. The
assertion-level accuracy is
\begin{equation}
\mathrm{Acc} = \frac{\sum_{q} \sum_{i} y_{q,i}}{\sum_{q} |A_q|},
\end{equation}
where $A_q$ is the set of assertions for query $q$. We also compute a
query-level macro-average (each query weighted equally regardless of
its assertion count); rankings agree with assertion-level, and the
full numbers are in Appendix~\ref{app:query-level}.

\subsection{LLM-as-a-Judge}
\label{sec:llm-judge}

Atomic claim decomposition has been previously used by FActScore~\citep{min2023factscore}; \OCB{} extends this by combining decomposition with a three-judge majority vote. The atomic-assertion scoring builds on the LLM-evaluator framework introduced by G-Eval~\citep{liu2023geval} and the LLM-as-a-judge paradigm validated at scale by~\citet{zheng2023mtbench}. The three-judge majority vote builds on PoLL~\citep{verga2024poll}, which showed that a diverse panel of judges outperforms a single large judge while reducing intra-model bias.

The judge receives the query, the reference assertion, and the system
output, but not the source document. Correctness is instead established
earlier, at rubric construction: domain experts author and verify each
atomic assertion, ensuring it either follows directly from the source
file or is correctly derived from it through domain analysis (which may
involve intermediate computation). By grading time the assertions
represent an expert-validated reference answer, so the judge's task is
the narrower decision of whether the system output states a given
assertion, which can be determined from the output alone. The full
judge prompt is in Appendix~\ref{app:judge-prompt}. Every assertion is
evaluated independently by three judge models invoked via provider APIs:
GPT-5.4 Thinking with low reasoning effort, Gemini 3.1 Pro, and Claude
Opus 4.6 with default thinking. Holding the judge ensemble fixed makes
grading reproducible across runs. The final verdict is the majority vote
$\hat{y}_{q,i} = \mathbf{1}\{\sum_{j=1}^{3} v_j \ge 2\}$ where
$v_j \in \{0,1\}$ is judge $j$'s binary verdict.

\subsection{LLM Judge Agreement}
\label{sec:judge-agreement}

The three-judge ensemble (GPT-5.4 Thinking, Gemini 3.1 Pro, Claude Opus
4.6) shows substantial pairwise agreement on assertion-level verdicts
on Domain Q\&A. Disagreement is concentrated in pairs involving
Gemini 3.1, consistent with Gemini applying stricter pass criteria
than the other two. A leave-one-out ablation
(Appendix~\ref{app:judge-agreement}) shows that every 2-judge subset
reaches the same verdict as the full panel on $78$--$89\%$ of
assertions, with chance-corrected agreement (Cohen's $\kappa$) of
$0.58$--$0.77$. No single 2-judge subset reproduces the full panel,
so each judge contributes a non-trivial deciding vote. Combined with
the low evaluation-noise variance reported in \S\ref{sec:variance},
this supports treating the 3-judge majority verdict as a stable
measurement. The full inter-judge vote breakdown, LOO ablation table,
and observations are in Appendix~\ref{app:judge-agreement}.

\subsection{Benchmark Variance Analysis}
\label{sec:variance}
To make the leaderboard interpretable under run-to-run noise, we decompose each subject model's score variance and report a $95\%$ confidence interval. Because eval reps within a scrape share a single model response and are positively correlated, we construct the headline interval at the scrape level \citep{cameronmiller2015cluster}.\footnote{Treating the $9$ runs as
independent would understate uncertainty because eval reps within a
scrape share the same model response and are positively correlated,
the classical clustered-data problem. We therefore construct the headline CI from \emph{scrape-level} variance and use sample standard
deviations ($n{-}1$) throughout.} For each subject model we collect $3$ independent response \emph{scrapes} from the web interface and grade each scrape $3$
times with the judge ensemble of \S\ref{sec:judge-agreement}, yielding
$9$ evaluation runs.\footnote{The 3-scrape, 3-evaluation protocol is applied only to Domain Q\&A, where open-ended analytical responses and large per-query assertion sets make run-to-run variance the most consequential. For File Fidelity Q\&A, where each query is scoped to a single artifact and per-assertion variance is much lower, we run a single scrape and a single evaluation pass per model.}

\paragraph{Variance decomposition.}
We separate run-to-run noise into \textbf{Evaluation variance (Eval Var)},
judge wobble with the model response held fixed, and
\textbf{Response-sampling variance (Scrape Var)}, model wobble across
re-runs of the same prompt. We also report \textbf{Overall variance
(Overall Var)} (sample SD over all $9$ runs) as a diagnostic, though
it is not the right quantity for a CI because eval reps within a
scrape are correlated. The headline $95\%$ CI is constructed at the
scrape level:
\begin{equation}
\mathrm{CI}_{95\%} = \pm\,t_{0.025,\,n_s-1}\,\frac{\sigma_{\mathrm{scrape}}}{\sqrt{n_s}},
\end{equation}
where $n_s = 3$ and $t_{0.025,\,2}=4.303$. This is a conservative
interval reflecting only $3$ independent response samples per model.

\begin{table}[t]
\centering
\scriptsize
\setlength{\tabcolsep}{3pt}
\renewcommand{\arraystretch}{1.1}
\caption{Assertion-level accuracy and variance decomposition on
  Domain Q\&A ($3$ scrapes $\times$ $3$ evals = $9$ runs per model).
  Variance columns are sample standard deviations in percentage points;
  CI is the $95\%$ scrape-level confidence interval on the mean.
  Query-level numbers yield the same ranking.}
\label{tab:variance-assertion}
\resizebox{\linewidth}{!}{%
\begin{tabular}{lccccc}
\toprule
\textbf{Model} & \textbf{Accuracy} & \textbf{Overall Var} & \textbf{Scrape Var} & \textbf{Eval Var} & \textbf{$95\%$ CI ($\pm$)} \\
\midrule
GPT-5.5 Thinking & $59.35\%$ & $0.41$ & $0.45$ & $0.15$ & $\pm 1.11$ \\
Claude Opus 4.7  & $56.83\%$ & $0.25$ & $0.25$ & $0.15$ & $\pm 0.61$ \\
Gemini 3.1 Pro   & $45.75\%$ & $0.36$ & $0.37$ & $0.18$ & $\pm 0.91$ \\
\bottomrule
\end{tabular}%
}
\end{table}

\paragraph{Discussion.}
Two findings drive how we report results elsewhere.
Response-sampling noise dominates judge noise for every model
(Scrape Var is roughly $1.6\times$--$3.1\times$ Eval Var), so under this design the evaluation budget is better spent on multiple scrapes than on additional judge re-runs.
And the $95\%$ CIs of the top two models do not overlap
(GPT-5.5: $[58.24, 60.46]$; Claude: $[56.22, 57.44]$), supporting the
reported ranking and justifying treatment of GPT-5.5 Thinking and
Claude Opus 4.7 as a top tier with Gemini 3.1 Pro clearly separated.
Extended discussion is in Appendix~\ref{app:variance-discussion}.

\section{Headline Results}
\label{sec:results}

We evaluated Anthropic Claude Opus 4.7, OpenAI
GPT-5.5 Thinking, and Google Gemini 3.1 Pro on both \OCB{} tracks
using the assertion-level evaluation framework described in
\S\ref{sec:eval}.

\begin{figure}[t]
  \centering
  \includegraphics[width=\linewidth]{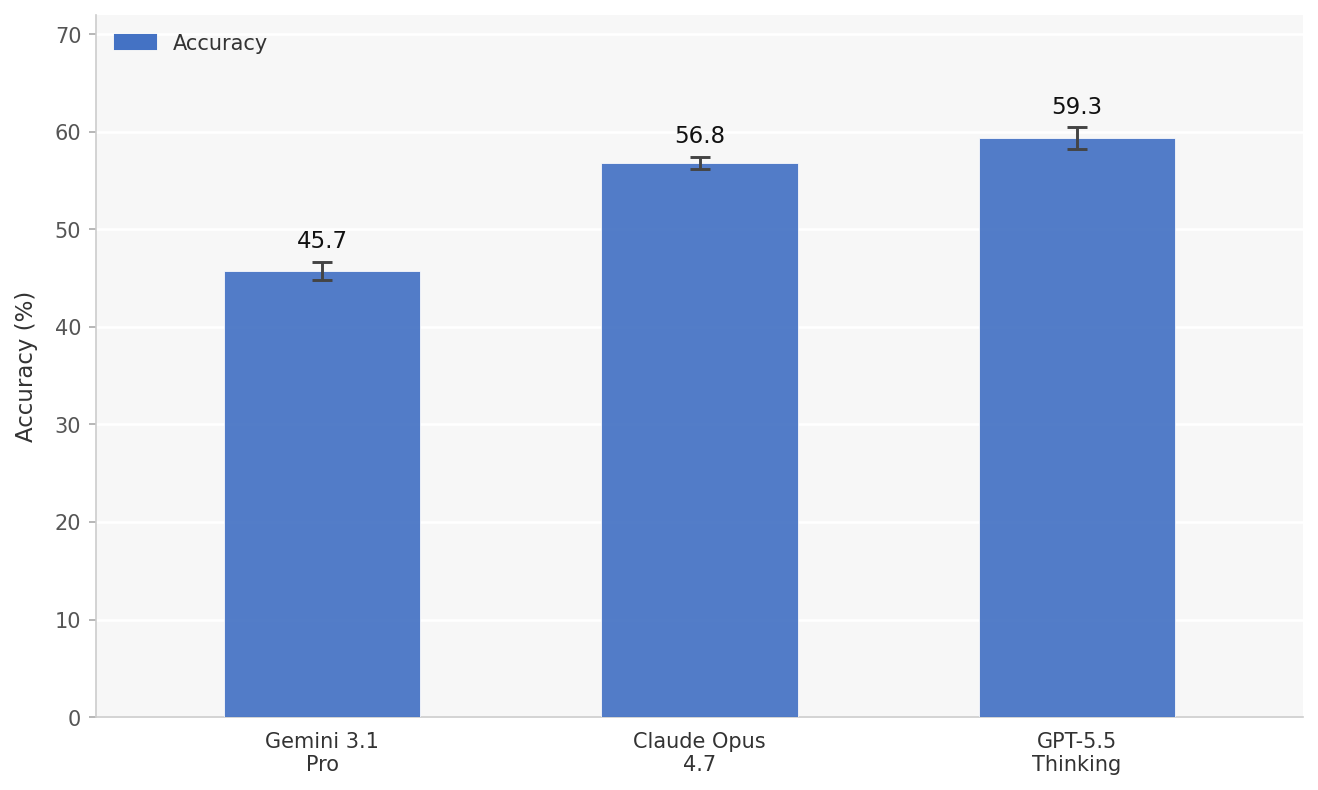}
  \caption{Domain Q\&A Accuracy on \OCB{}.}
  \label{fig:win-rate-chart}
\end{figure}

\begin{figure}[t]
  \centering
  \includegraphics[width=\linewidth]{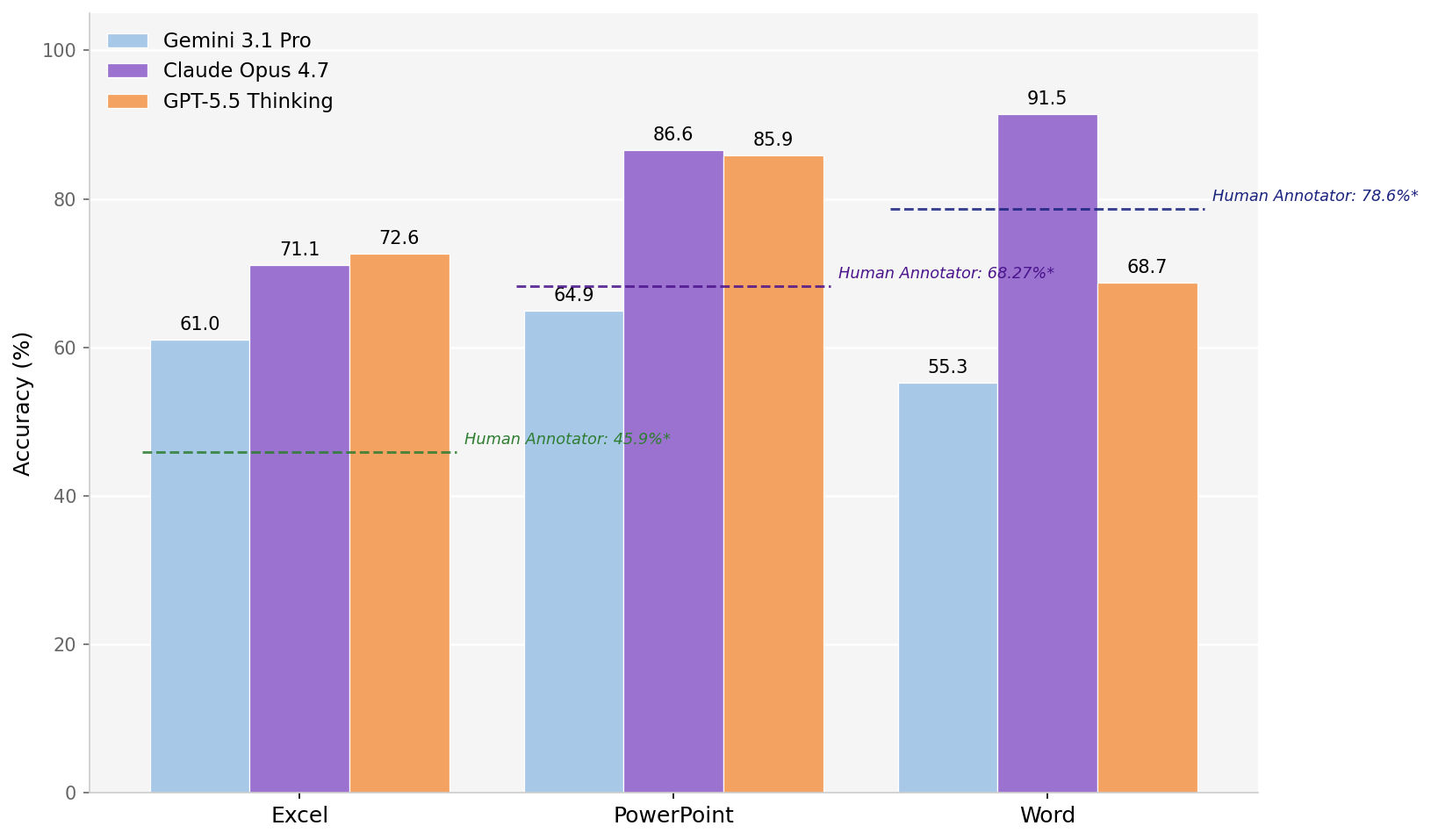}
  \caption{File Fidelity Q\&A accuracy on \OCB{}. Bars are
  assertion-level accuracy for each model. Dashed lines show
  single-rater human-annotator accuracy on the same metric, under the
  same single-pass setup as the LLM systems; annotators were vetted
  for general office literacy rather than specialized domain expertise.
  See Appendix~\ref{app:human-baseline} for coverage and methodology.}
  \label{fig:file-fidelity-accuracy-chart}
\end{figure}

\paragraph{Domain Q\&A (Figure~\ref{fig:win-rate-chart}).}
GPT-5.5 Thinking achieves the highest accuracy at 59.3\%, followed
closely by Claude Opus 4.7 at 56.8\%, while Gemini 3.1 Pro reaches
45.7\%. The top two systems form a tight cluster: their point
estimates are separated by $2.5$ pp, and their $95\%$ scrape-level
confidence intervals are non-overlapping under our $3$-scrape
protocol with a cushion of $\sim 0.8$ pp (see \S\ref{sec:variance}).
Given the limited number of scrapes, we describe GPT-5.5 Thinking and
Claude Opus 4.7 as a co-leading top tier rather than claiming a
decisive separation; Gemini 3.1 Pro trails by roughly $11$--$14$
percentage points, with its CI clearly separated from both top-tier
systems.

\paragraph{File Fidelity Q\&A by application (Figure~\ref{fig:file-fidelity-accuracy-chart}).}
Performance varies across Excel, PowerPoint, and Word, with no single
system dominating uniformly. On Excel, GPT-5.5 Thinking leads narrowly
at 72.6\%, with Claude Opus 4.7 just behind at 71.1\% and Gemini 3.1
Pro at 61.0\%. On PowerPoint, Claude Opus 4.7 leads at 86.6\%, followed
by GPT-5.5 Thinking at 85.9\% and Gemini 3.1 Pro at 64.9\%. The largest
spread appears on Word, where Claude Opus 4.7 reaches 91.5\%,
substantially ahead of GPT-5.5 Thinking (68.7\%) and Gemini 3.1 Pro
(55.3\%), a 36-point gap between the strongest and weakest systems.
Across all three applications, Gemini 3.1 Pro consistently trails the
other two systems.

\paragraph{Human-annotator reference on File Fidelity.}
To anchor File Fidelity scores against a human reference, we collected
accuracy from human annotators on a substantial subset of
queries per application ($96.6\%$ of PowerPoint, $91.9\%$ of Word, and
$76\%$ of Excel). Annotators used the same single-pass setup as the
LLM systems. They reached $78.6\%$ on Word, $68.3\%$ on PowerPoint,
and $45.9\%$ on Excel
(Figure~\ref{fig:file-fidelity-accuracy-chart}, dashed lines). All
three frontier systems exceed annotator accuracy on every sub-track:
Claude Opus 4.7 leads annotators by $\sim$$13$ pp on Word and
$\sim$$18$ pp on PowerPoint, and GPT-5.5 Thinking leads by $\sim$$27$
pp on Excel. The widest model–annotator gap is on Excel, where many
queries require quantitative reasoning. LLM systems scan
systematically, while one-pass human readers may miss the target.
This is evidence that File Fidelity rewards tirelessness and
systematic recall, complementary to the synthesis-and-reasoning
capabilities probed by Domain Q\&A.

Combined with the $59.3\%$ Domain Q\&A ceiling, these results suggest
a split capability profile for current frontier systems: super-annotator
on artifact-level perception and recall, sub-expert on multi-step
analytical reasoning.

\section{Per-Application File Fidelity Analysis}
\label{sec:per-app}

The headline File Fidelity numbers in \S\ref{sec:results} aggregate
across all artifact types within each application, hiding substantial
per-artifact variation. To surface this variation, we decompose
accuracy across the per-app artifact-type taxonomy of
Appendix~\ref{app:taxonomy} and document size; the full per-app
breakdowns (six figures across Word, Excel, and PowerPoint) and
per-size analyses are in
Appendix~\ref{app:filefidelity-breakdowns}. We highlight three
cross-app patterns.

\paragraph{Claude leads on app-specific structural artifacts.}
On Word, Claude Opus 4.7 reaches 100\% on Comment, Equation, Hyperlink,
and Text, and stays above 80\% on most other categories. GPT-5.5
Thinking collapses on a handful of structural artifacts (Hyperlink 6\%,
Font 17\%, Header 17\%, Metadata 6\%), and Gemini 3.1 Pro is near-zero
on Comment, Layout, Header, Footer, Highlight, and Metadata. On
PowerPoint, Claude and GPT-5.5 are roughly tied overall (86.6\% vs.\
85.9\%), with Claude edging out GPT-5.5 on several structural
categories (e.g., Equation 100\% vs.\ 78\%, Metadata 100\% vs.\ 67\%).
Gemini falls behind on structure-bound artifacts (Section 0\%,
Shape/Text 22\%, Chart 36\%, Metadata 33\%). Overall, Claude's
36-point Word lead over Gemini (91.5\% vs.\ 55.3\%) and the Gemini
PowerPoint shortfall are driven primarily by artifacts that depend on
app-specific structural metadata, not by text or table reading.
Several called-out PowerPoint categories have small per-bucket query
counts ($n{\leq}10$) and should be read as illustrative; full
per-category bars and $n$-counts are in
Figure~\ref{fig:ppt-feature}.

\paragraph{Excel is closer across systems, but Table-Like Ranges remain hard.}
All three systems perform comparably on formal Tables (78--86\%) and
Sheets (88--100\%). The hardest artifact across Excel is the
\emph{Table-Like Range} (TLR): contiguous cell ranges not formally
registered as Excel Tables, where row/column structure must be
inferred. With 133 queries (the largest Excel category), all three
systems land in the 48--58\% band on TLR. Gemini 3.1 Pro additionally
drops to 44\% on formulas and 0\% on embedded images.

\paragraph{File-size effects are app-dependent.}
Word performance degrades sharply with length for GPT-5.5 Thinking and
Gemini 3.1 Pro (both fall $\sim$40 pp from Small to Long), while
Claude Opus 4.7 holds 81--96\% across size buckets. Excel accuracy
improves Small to Medium across all systems, attributable to
composition shift (Small workbooks are TLR-heavy, Medium contain more
formal Tables) rather than a true size effect; the Long bucket is too
small to draw further conclusions. PowerPoint is roughly
size-invariant for Claude and GPT-5.5 (85--87\%), with Gemini weaker
overall.
\section{Domain Q\&A Analysis}
\label{sec:domain-analysis}

We now turn from the headline Domain Q\&A number to a finer-grained
view: where models succeed and fail by industry and file type, and
\emph{how} they fail when they do.

\subsection{Performance Across Industries and File Characteristics}
\begin{figure*}[h]
  \centering
  \begin{minipage}[t]{0.48\textwidth}
    \centering
    \includegraphics[width=\linewidth,height=0.16\textheight,keepaspectratio]{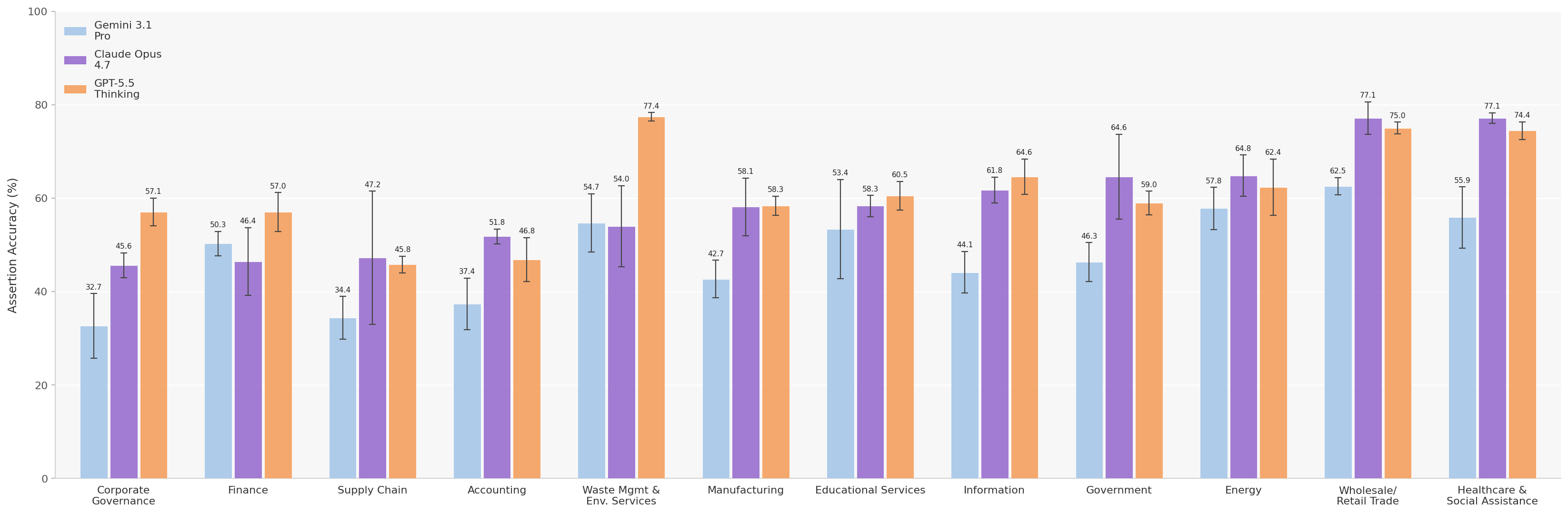}
    \caption{Domain Q\&A accuracy by industry.}
    \label{fig:industries}
  \end{minipage}
  \hfill
  \begin{minipage}[t]{0.48\textwidth}
    \centering
    \includegraphics[width=\linewidth,height=0.16\textheight,keepaspectratio]{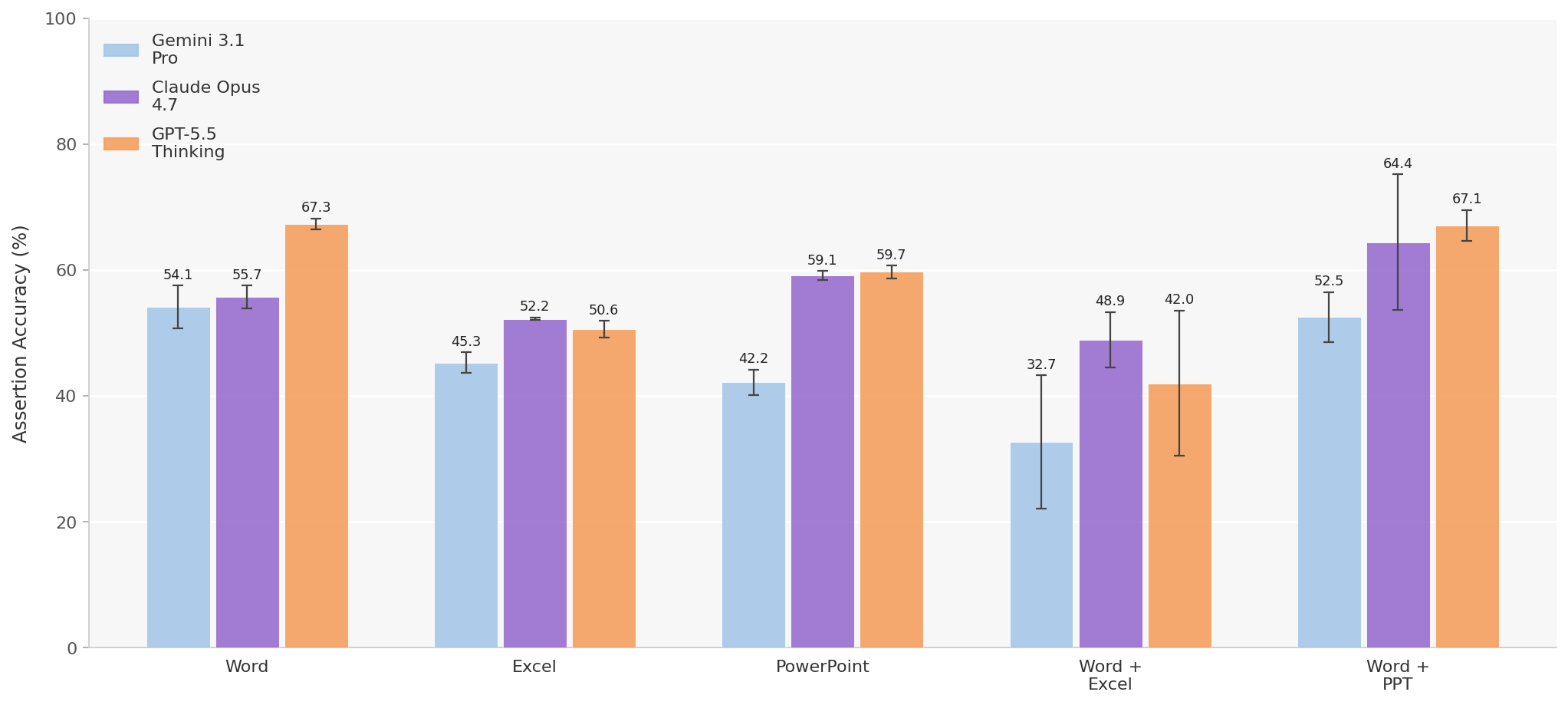}
    \caption{Domain Q\&A accuracy by file type.}
    \label{fig:filetypes}
  \end{minipage}
\end{figure*}

Figures~\ref{fig:industries} and~\ref{fig:filetypes} provide a finer-grained
view of performance across industries and document characteristics on the Domain Q\&A dataset.

\paragraph{Key patterns.}
Across industries, Claude Opus~4.7 and GPT-5.5 Thinking are
consistently top-tier and Gemini~3.1 Pro trails in nearly every
sector. The one exception is Finance, where Gemini~3.1 Pro (50.3\%)
slightly edges out Claude Opus~4.7 (46.4\%) while still trailing
GPT-5.5 Thinking (57.0\%). Difficulty varies sharply by domain:
Healthcare and Wholesale/Retail are relatively strong across
systems, while Supply Chain, Corporate Governance, and Accounting
remain the hardest. By file type, mixed-input queries (especially
Word$+$Excel) are the most error-prone, suggesting that heterogeneous
structure is a key challenge beyond any single application format.

\subsection{Failure Modes}
\label{sec:failures}

Beyond aggregate accuracy, qualitative inspection of judge-rejected
assertions surfaces four recurring failure modes that account for the
majority of errors observed across the three systems.

\paragraph{Cross-document sourcing failures.}
When the prompt provides multiple source documents, some models drop
or mix up values across filings. In one task requesting a table of
percent of shares outstanding contacted before and after the annual
meeting for two consecutive years, with year-over-year changes in
basis points, a model reported the prior-year values as ``N/A''
despite the proxy statement clearly stating those values.

\paragraph{Value hallucinations.}
Some models report values that never appear in the source material.
In one task asking for a 5-year DCF analysis with specified revenue
growth assumptions, a model reported FY2025A revenue as \$2{,}949M;
the source 10-K reports the figure as \$2{,}977M, and the substring
``2,94'' does not appear in the relevant context of the document.

\paragraph{Methodology failures.}
Even when the prompt explicitly specifies the formula or decomposition
to use, some models ignore it and substitute their own assumptions.
In one task asking for Passenger Travel Revenue forecast as
$\text{Yield (18\%)} \times \text{RPM}$ for each forecast period, a
model used the requested formula but added a 4\% annual growth factor
that was not specified in the prompt.

\paragraph{Computational errors.}
Basic arithmetic and ratio errors - wrong numerator, wrong
denominator, mis-annualization - are common. In one task asking for
operating cash flow per dollar of \emph{deferred preneed revenue}, a
model divided operating cash flow by \emph{preneed revenue}
(\$945M / \$16B) instead of by deferred preneed revenue
(\$945M / \$1.76B), yielding a result off by roughly an order of
magnitude.

These failure modes compound: because rubric assertions often build on
prior values (e.g., a ratio whose denominator was sourced incorrectly),
a single sourcing or methodology error can cascade across multiple
assertions in the same response.

\section{Test-Time Compute Ablations}
\label{sec:ablations}

To assess how Domain Q\&A performance depends on test-time compute
settings, we ablate the two top systems across multiple thinking modes
using a two-track protocol. The Standard-Thinking configurations used
for the headline results in \S\ref{sec:results} are evaluated with the
full $3$-scrape $\times$ $3$-eval design described in
\S\ref{sec:variance}. For all other thinking modes, we use a leaner
$3$-scrape $\times$ $1$-eval protocol, justified by the small Eval Var
observed under Standard-Thinking
(Table~\ref{tab:variance-assertion}): with judge wobble bounded below
$0.18$ percentage points, additional eval re-runs contribute
negligibly to the headline uncertainty, and the evaluation budget is
better spent covering more thinking modes. All accuracy CIs in both
tracks are constructed at the scrape level using
$t_{0.025,\,2}\cdot\sigma_{\mathrm{scrape}}/\sqrt{3}$, so per-row
intervals are directly comparable across the two protocols. For
Claude~Opus~4.7, we compare \emph{Standard-Thinking} against
\emph{Adaptive-Thinking}, which dynamically allocates additional
reasoning budget. For GPT-5.5, we sweep four thinking depths
(\emph{Thinking-Light}, \emph{-Standard}, \emph{-Extended},
\emph{-Heavy}) and additionally evaluate two higher-tier modes,
\emph{Pro-Standard} and \emph{Pro-Extended}. The headline GPT-5.5
number in \S\ref{sec:results} corresponds to Thinking-Standard.

\begin{figure}[ht]
\centering
\begin{minipage}[t]{0.48\linewidth}
  \centering
  \includegraphics[width=\linewidth]{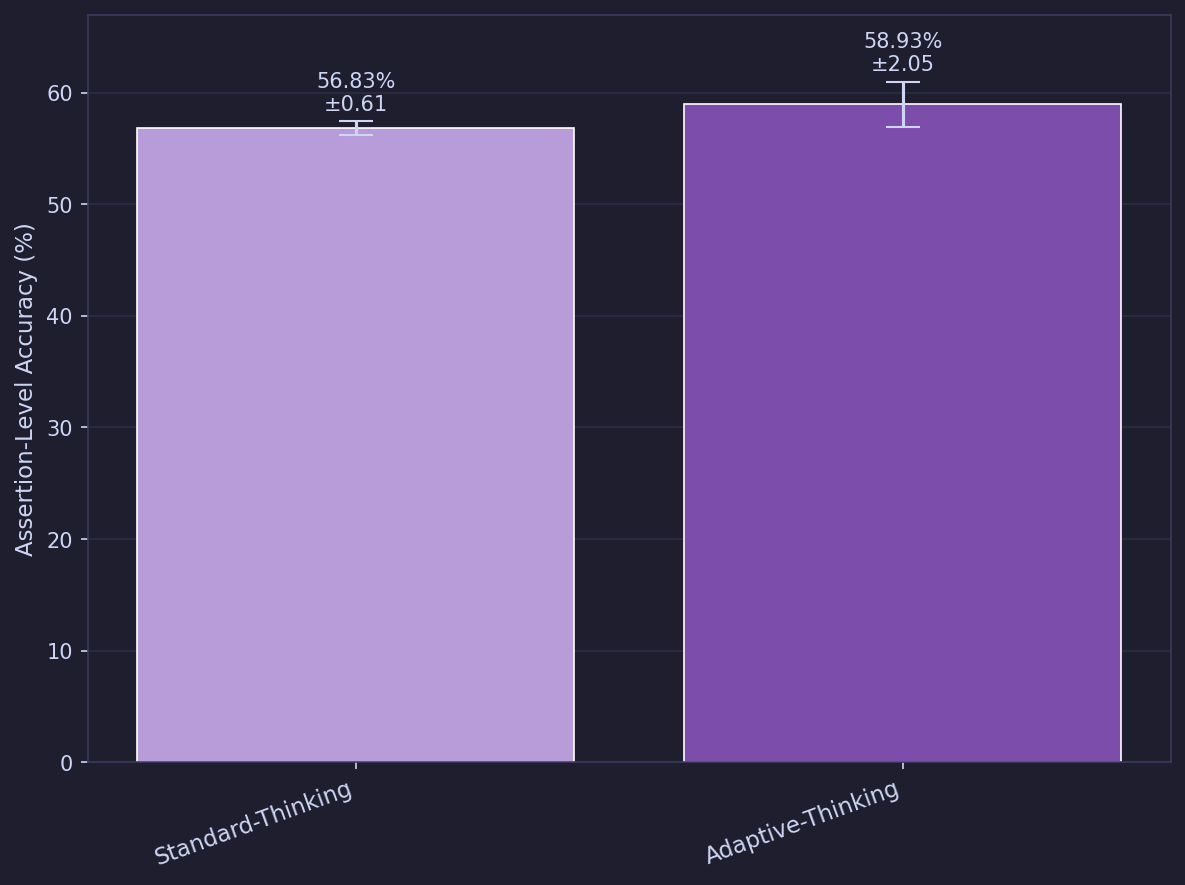}
  \caption{Claude Opus 4.7: assertion accuracy by thinking mode.}
  \label{fig:claude-ablation-acc}
\end{minipage}\hfill
\begin{minipage}[t]{0.48\linewidth}
  \centering
  \includegraphics[width=\linewidth]{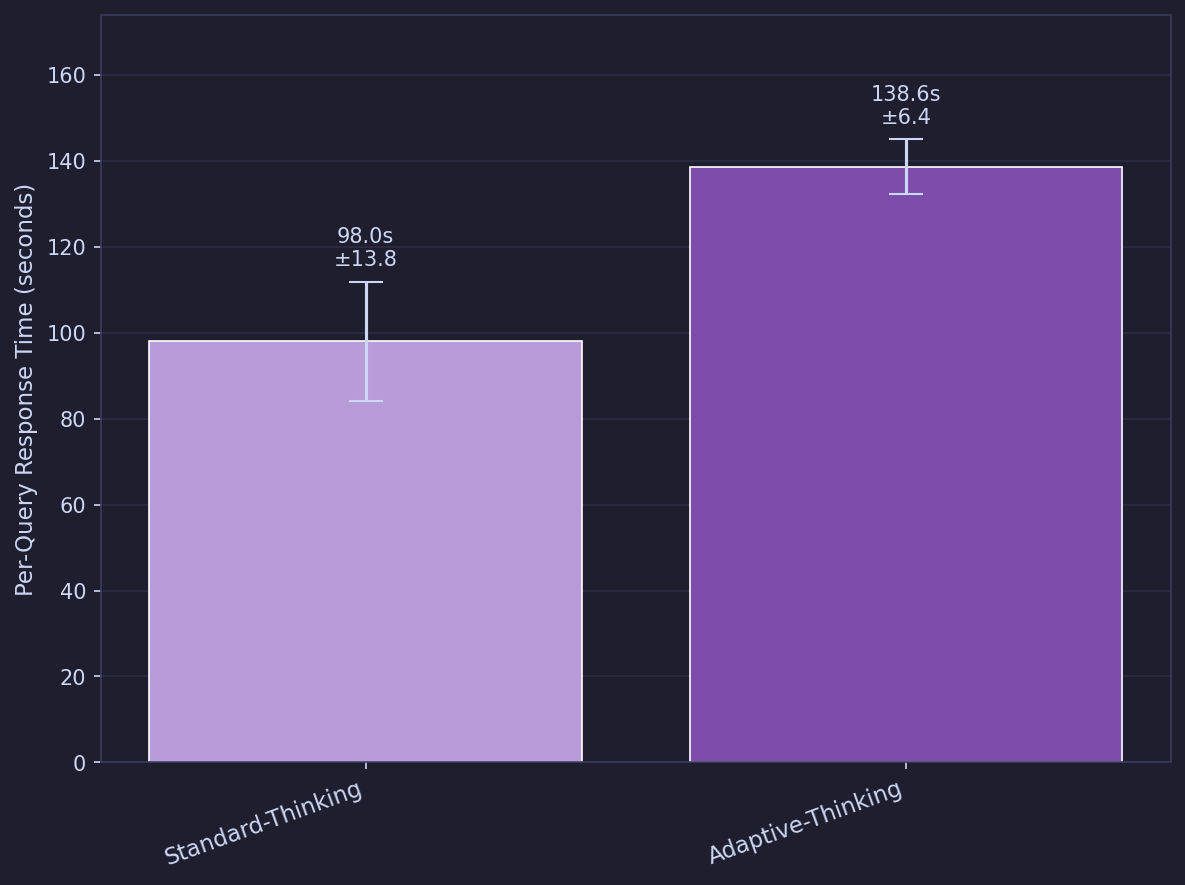}
  \caption{Claude Opus 4.7: per-query response time by thinking mode.}
  \label{fig:claude-ablation-time}
\end{minipage}
\end{figure}

\paragraph{Accuracy.}
Per-mode accuracies and per-query response times are shown in
Figures~\ref{fig:claude-ablation-acc}--\ref{fig:gpt-ablation-time};
full numerical results with $95\%$ scrape-level CIs are in
Appendix~\ref{app:ablation-table}. Within the GPT-5.5 Thinking
series, accuracy is essentially flat ($59.11\%$--$59.81\%$, a
$0.70$ pp spread) and CIs overlap across all four depths. The
modes are statistically indistinguishable under our $3$-scrape
protocol, despite Thinking-Heavy consuming roughly twice the compute
of Thinking-Light. Adaptive-Thinking gains $+2.10$ pp over
Standard-Thinking for Claude.

\begin{figure}[ht]
\centering
\begin{minipage}[t]{0.48\linewidth}
  \centering
  \includegraphics[width=\linewidth]{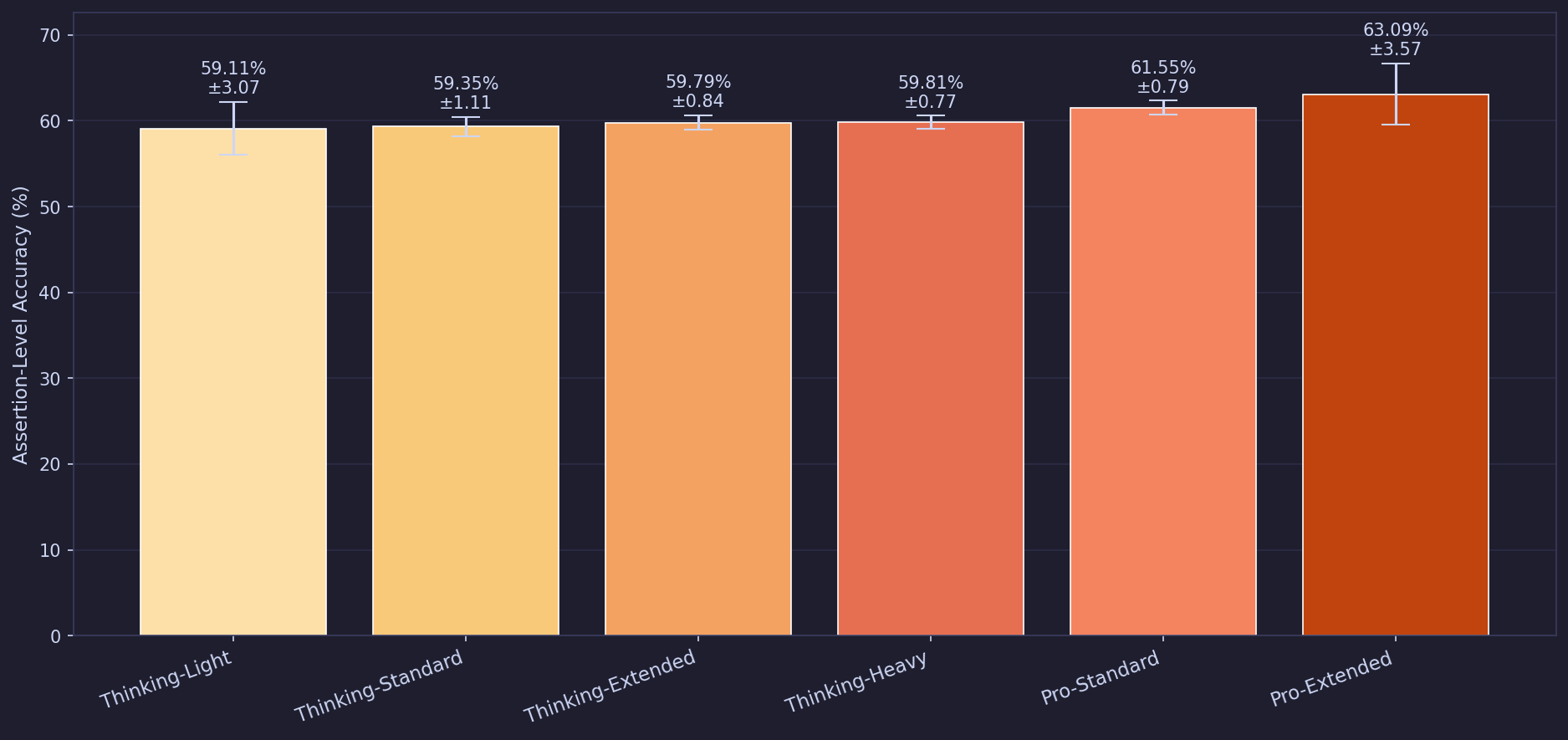}
  \caption{GPT-5.5: assertion accuracy across six thinking modes.}
  \label{fig:gpt-ablation-acc}
\end{minipage}\hfill
\begin{minipage}[t]{0.48\linewidth}
  \centering
  \includegraphics[width=\linewidth]{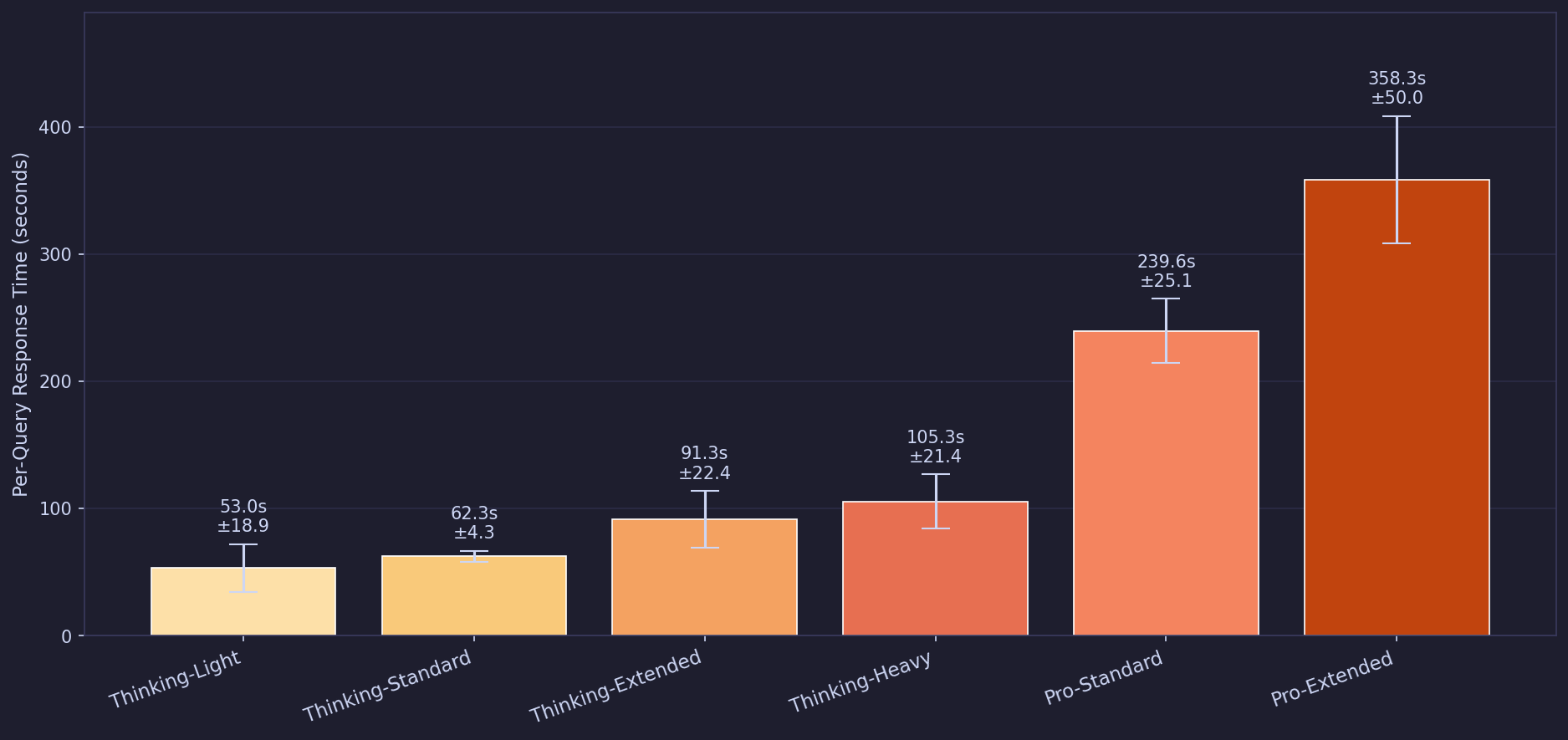}
  \caption{GPT-5.5: per-query response time across six thinking modes.}
  \label{fig:gpt-ablation-time}
\end{minipage}
\end{figure}

\paragraph{Tier vs.\ depth, and the compute price.}
The largest gains come from changing tier rather than depth: GPT-5.5
Pro-Standard reaches $61.55\%$ and Pro-Extended reaches $63.09\%$,
both above the Thinking series. The Pro-Standard interval marginally
separates from the Thinking series under the scrape-level CI
($\sim 0.2$ pp cushion above Thinking-Heavy's upper bound), while
Pro-Extended's wider CI overlaps the Thinking-Heavy interval despite
a $\sim 3.3$ pp higher point estimate. Per-query response time scales
far faster than accuracy: from Thinking-Light to Pro-Extended is a
$\sim 6.8\times$ increase in latency for a $3.98$ pp accuracy gain.
Net implication: within a model family, increasing thinking depth
alone yields no improvement separable from noise under our
scrape-level CIs. Switching to a higher capability tier yields a
modest gain that separates from the Thinking series for Pro-Standard.
Pro-Extended raises the point estimate further, but its wider CI
still overlaps Thinking-Heavy. The fact that Pro-Standard separates
cleanly from Thinking-Heavy despite only a $\sim 1.7$ pp absolute
gain suggests \OCB{} has the statistical resolution to distinguish
configurations within a single model family, not just across families.

\section{Related Work}
\label{sec:related}
\vspace{-0.4em}

\paragraph{\mbox{Single-page and single-artifact benchmarks.}}
Single-page document QA (DocVQA~\citep{mathew2021docvqa},
InfographicVQA~\citep{mathew2022infographicvqa},
VisualMRC~\citep{tanaka2021visualmrc}, ST-VQA~\citep{biten2019stvqa}),
chart QA (ChartQA~\citep{masry2022chartqa},
PlotQA~\citep{methani2020plotqa}), and table QA
(WikiTableQuestions~\citep{pasupat2015wikitablequestions},
TabFact~\citep{chen2020tabfact}, TAT-QA~\citep{zhu2021tatqa},
FinQA~\citep{chen2021finqa}, ConvFinQA~\citep{chen2022convfinqa},
MultiHiertt~\citep{zhao2022multihiertt}) study individual artifacts
stripped from document context with short or extractive answers.
SlideVQA~\citep{tanaka2023slidevqa} uses rendered slide images;
SpreadsheetBench~\citep{ma2024spreadsheetbench} and
SheetCopilot~\citep{li2023sheetcopilot} use native \texttt{.xlsx} but
evaluate manipulation, not comprehension.
\vspace{-0.35em}
\paragraph{\mbox{Multi-page and enterprise benchmarks.}}
Multi-page benchmarks (MP-DocVQA~\citep{tito2023mpdocvqa},
DUDE~\citep{vanlandeghem2023dude},
MMLongBench-Doc~\citep{ma2024mmlongbenchdoc},
DocBench~\citep{zou2024docbench},
LongDocURL~\citep{deng2024longdocurl}) flatten everything into PDF page
images. Enterprise-domain benchmarks add genre-locking:
FinanceBench~\citep{islam2023financebench},
OfficeQA Pro~\citep{databricks2026officeqapro},
GDP.PDF~\citep{garre2026gdppdfbenchmarkinggroundedmultimodal},
CUAD~\citep{hendrycks2021cuad},
ContractNLI~\citep{koreeda2021contractnli},
LegalBench~\citep{guha2023legalbench}, and
QASPER~\citep{dasigi2021qasper}; the first three are the closest
enterprise comparators to \OCB{}'s Domain Q\&A, but all are
PDF-rendered.
\vspace{-0.35em}
\paragraph{\mbox{Office-workflow benchmarks.}}
\mbox{OfficeBench}~\citep{wang2024officebench} and
\mbox{OdysseyBench}~\citep{wang2025odysseybench} grade automation workflows;
GDPval~\citep{patwardhan2025gdpval} grades work-product creation across
44 occupations via head-to-head human comparison. \OCB{} is
comprehension-only and complementary. No prior public benchmark
jointly addresses native Office file formats, format-specific
metadata, and multi-paragraph analytical comprehension; see
Appendix Table~\ref{tab:related}.

\section{Conclusion}
\vspace{-0.4em}

We presented \textbf{Office Comprehension Bench (\OCB)}, the first
public benchmark to jointly evaluate Word, Excel, and PowerPoint
comprehension over native file formats. \OCB{}'s two tracks,
\textbf{File Fidelity} for structural and visual perception, and
\textbf{Domain Q\&A} for expert-level reasoning across 12 industries
with rubrics scored by an LLM-judge ensemble, show that even the
strongest evaluated systems plateau at $\sim$59.3\% in default
reasoning modes on Domain Q\&A, with higher product tiers reaching
only $\sim$63\%. On File Fidelity Q\&A, by contrast, frontier systems
exceed a human-annotator baseline across all three applications,
revealing a split capability profile: super-annotator on artifact
perception and recall, sub-expert on multi-step analytical reasoning.
Robust comprehension across both ends of this spectrum over real-world
Office documents remains a fundamentally challenging capability, and
we hope \OCB{} accelerates progress toward LLM systems that are
accurate, grounded, and robust across the artifacts that define
modern office work.
\section{Limitations}
\OCB{} is an initial release; we aim to be transparent about its current scope
and limitations.%
\begin{itemize}[itemsep=-2pt, topsep=0pt, parsep=0pt, partopsep=0pt]
\item \textbf{Single-turn evaluation.} All queries are evaluated as a
single user turn; real-world document work often involves follow-ups
and incremental refinement, which we leave to future work.
\item \textbf{Comprehension only.} \OCB{} evaluates document
understanding and reasoning, not actions such as editing files,
creating content, or executing workflows. Subject systems may produce
intermediate artifacts (e.g., scratch spreadsheets or working notes)
as part of arriving at an answer, but those artifacts are not graded;
only the final response against the assertion set is scored.
\item \textbf{Limited multi-application coverage.} Most queries are
grounded in a single application's file format; coverage of mixed
inputs (e.g., Word$+$Excel, Word$+$PowerPoint) is present but limited,
and broader combinations of file types within a single query are left
to future versions.
\item \textbf{Workflow context.} Real analytical tasks involve
coordination across email, chat, and meetings; \OCB{} does not
capture this surrounding workflow, so strong \OCB{} performance may
not fully translate to deployment.
\item \textbf{Assertion quality.} Assertions were authored and
reviewed by domain experts, with iteration to split
compound claims, remove duplicates, and resolve ambiguities, but are
not perfect; we invite community contributions to refine the
benchmark.
\item \textbf{English only.} All queries and documents are in
English; multilingual document comprehension is left to future work.
\item \textbf{No human baseline on Domain Q\&A.} We report a
single-rater human-annotator baseline on File Fidelity Q\&A
(\S\ref{sec:results}) but not on Domain Q\&A. Without one, the
$\sim$59.3\% Domain Q\&A headline is not anchored against a known
ceiling; a proper expert-baseline study with multiple raters and
inter-rater agreement is planned for future work.
\item \textbf{Single-rater File Fidelity baseline.} The annotator
baseline in Figure~\ref{fig:file-fidelity-accuracy-chart} uses one
rater per query without inter-rater agreement measured, and
annotators were trained for general office literacy rather than
specialized domain expertise (notably affecting the Excel result,
where many queries require quantitative reasoning). The numbers
indicate the relative model–annotator standing on routine artifact
reading rather than a calibrated ceiling; a larger multi-rater
study with domain-matched annotators is left to future work.
\item \textbf{Judge ensemble validation.} We report inter-judge
agreement and a leave-one-out judge ablation
(\S\ref{sec:judge-agreement}, Appendix~\ref{app:judge-agreement}),
but not judge-vs-human alignment; the latter would require a
stratified human-grading study against which to benchmark each judge.
The current evidence shows the three LLM judges are not redundant
(Cohen's $\kappa$ of $0.58$--$0.77$ across leave-one-out pairs) but
does not rule out correlated bias relative to human judgment.
\end{itemize}

{
\small
\bibliography{references}

@inproceedings{mathew2021docvqa,
  title         = {DocVQA: A Dataset for VQA on Document Images},
  author        = {Mathew, Minesh and Karatzas, Dimosthenis and Jawahar, C. V.},
  booktitle     = {Proceedings of the IEEE/CVF Winter Conference on Applications of Computer Vision (WACV)},
  year          = {2021},
  eprint        = {2007.00398},
  archivePrefix = {arXiv},
  primaryClass  = {cs.CV}
}

@inproceedings{mathew2022infographicvqa,
  title         = {InfographicVQA},
  author        = {Mathew, Minesh and Bagal, Viraj and Tito, Rub{\`e}n P{\'e}rez and Karatzas, Dimosthenis and Valveny, Ernest and Jawahar, C. V.},
  booktitle     = {Proceedings of the IEEE/CVF Winter Conference on Applications of Computer Vision (WACV)},
  year          = {2022},
  eprint        = {2104.12756},
  archivePrefix = {arXiv},
  primaryClass  = {cs.CV}
}

@inproceedings{tanaka2021visualmrc,
  title         = {{VisualMRC}: Machine Reading Comprehension on Document Images},
  author        = {Tanaka, Ryota and Nishida, Kyosuke and Yoshida, Sen},
  booktitle     = {Proceedings of the AAAI Conference on Artificial Intelligence},
  year          = {2021},
  eprint        = {2101.11272},
  archivePrefix = {arXiv},
  primaryClass  = {cs.CL}
}

@inproceedings{biten2019stvqa,
  title={Scene text visual question answering},
  author={Biten, Ali Furkan and Tito, Ruben and Mafla, Andres and Gomez, Lluis and Rusinol, Mar{\c{c}}al and Valveny, Ernest and Jawahar, CV and Karatzas, Dimosthenis},
  booktitle={Proceedings of the IEEE/CVF international conference on computer vision},
  pages={4291--4301},
  year={2019}
}

@inproceedings{masry2022chartqa,
  title         = {{ChartQA}: A Benchmark for Question Answering about Charts with Visual and Logical Reasoning},
  author        = {Masry, Ahmed and Long, Do Xuan and Tan, Jia Qing and Joty, Shafiq and Hoque, Enamul},
  booktitle     = {Findings of the Association for Computational Linguistics: ACL 2022},
  pages={2263--2279},
  year={2022}
}

@inproceedings{methani2020plotqa,
  title         = {{PlotQA}: Reasoning over Scientific Plots},
  author        = {Methani, Nitesh and Ganguly, Pritha and Khapra, Mitesh M. and Kumar, Pratyush},
  booktitle     = {Proceedings of the IEEE/CVF Winter Conference on Applications of Computer Vision (WACV)},
 pages={1527--1536},
  year={2020}
}

@inproceedings{pasupat2015wikitablequestions,
   title={Compositional semantic parsing on semi-structured tables},
  author={Pasupat, Panupong and Liang, Percy},
  booktitle={Proceedings of the 53rd Annual Meeting of the Association for Computational Linguistics and the 7th International Joint Conference on Natural Language Processing (Volume 1: Long Papers)},
  pages={1470--1480},
  year={2015}
}

@inproceedings{chen2020tabfact,
  title         = {{TabFact}: A Large-scale Dataset for Table-based Fact Verification},
   author={Chen, Wenhu and Wang, Hongmin and Chen, Jianshu and Zhang, Yunkai and Wang, Hong and Li, Shiyang and Zhou, Xiyou and Wang, William Yang},
  booktitle={International Conference on Learning Representations},
  year          = {2020},
}

@inproceedings{zhu2021tatqa,
  title         = {{TAT-QA}: A Question Answering Benchmark on a Hybrid of Tabular and Textual Content in Finance},
  author={Zhu, Fengbin and Lei, Wenqiang and Huang, Youcheng and Wang, Chao and Zhang, Shuo and Lv, Jiancheng and Feng, Fuli and Chua, Tat-Seng},
  booktitle={Proceedings of the 59th annual meeting of the Association for Computational Linguistics and the 11th international joint conference on natural language processing (volume 1: long papers)},
  pages={3277--3287},
  year={2021}
}

@inproceedings{chen2021finqa,
  title         = {{FinQA}: A Dataset of Numerical Reasoning over Financial Data},
  author={Chen, Zhiyu and Chen, Wenhu and Smiley, Charese and Shah, Sameena and Borova, Iana and Langdon, Dylan and Moussa, Reema and Beane, Matt and Huang, Ting-Hao and Routledge, Bryan and Wang, William Yang},
  booktitle={Proceedings of the 2021 Conference on Empirical Methods in Natural Language Processing},
  pages={3697--3711},
  year={2021}
}

@inproceedings{chen2022convfinqa,
  title         = {{ConvFinQA}: Exploring the Chain of Numerical Reasoning in Conversational Finance Question Answering},
  author={Chen, Zhiyu and Li, Shiyang and Smiley, Charese and Ma, Zhiqiang and Shah, Sameena and Wang, William Yang},
  booktitle={Proceedings of the 2022 conference on empirical methods in natural language processing},
  pages={6279--6292},
  year={2022}
}

@inproceedings{zhao2022multihiertt,
  title         = {{MultiHiertt}: Numerical Reasoning over Multi Hierarchical Tabular and Textual Data},
  author={Zhao, Yilun and Li, Yunxiang and Li, Chenying and Zhang, Rui},
  booktitle={Proceedings of the 60th Annual Meeting of the Association for Computational Linguistics (Volume 1: Long Papers)},
  pages={6588--6600},
  year={2022}
}

@inproceedings{tanaka2023slidevqa,
  title         = {{SlideVQA}: A Dataset for Document Visual Question Answering on Multiple Images},
  author={Tanaka, Ryota and Nishida, Kyosuke and Nishida, Kosuke and Hasegawa, Taku and Saito, Itsumi and Saito, Kuniko},
  booktitle={Proceedings of the AAAI Conference on Artificial Intelligence},
  volume={37},
  number={11},
  pages={13636--13645},
  year={2023}
}

@article{ma2024spreadsheetbench,
  title         = {{SpreadsheetBench}: Towards Challenging Real World Spreadsheet Manipulation},
  author={Ma, Zeyao and Zhang, Bohan and Zhang, Jing and Yu, Jifan and Zhang, Xiaokang and Zhang, Xiaohan and Luo, Sijia and Wang, Xi and Tang, Jie},
  journal={Advances in Neural Information Processing Systems (Datasets and Benchmarks Track)},
  volume={37},
  pages={94871--94908},
  year={2024}
}

@article{li2023sheetcopilot,
  title         = {{SheetCopilot}: Bringing Software Productivity to the Next Level through Large Language Models},
  author={Li, Hongxin and Su, Jingran and Chen, Yuntao and Li, Qing and Zhang, Zhao-Xiang},
  journal={Advances in Neural Information Processing Systems},
  volume={36},
  pages={4952--4984},
  year={2023}
}

@article{tito2023mpdocvqa,
  title         = {Hierarchical multimodal transformers for Multi-Page DocVQA},
  author={Tito, Rub{\`e}n and Karatzas, Dimosthenis and Valveny, Ernest},
  journal={Pattern Recognition},
  volume={144},
  pages={109834},
  year={2023},
  publisher={Elsevier}
}

@inproceedings{vanlandeghem2023dude,
  title         = {Document Understanding Dataset and Evaluation ({DUDE})},
  author={Van Landeghem, Jordy and Tito, Rub{\`e}n and Borchmann, {\L}ukasz and Pietruszka, Micha{\l} and Joziak, Pawel and Powalski, Rafal and Jurkiewicz, Dawid and Coustaty, Micka{\"e}l and Anckaert, Bertrand and Valveny, Ernest and others},
  booktitle={Proceedings of the IEEE/CVF International Conference on Computer Vision},
  pages={19528--19540},
  year={2023}
}

@article{peer2024anlsstar,
  title={ANLS*--a universal document processing metric for generative large language models},
  author={Peer, David and Sch{\"o}pf, Philemon and Nebendahl, Volckmar and Rietzler, Alexander and Stabinger, Sebastian},
  journal={arXiv preprint arXiv:2402.03848},
  year={2024}
}

@article{ma2024mmlongbenchdoc,
  title         = {{MMLongBench-Doc}: Benchmarking Long-context Document Understanding with Visualizations},
 author={Ma, Yubo and Zang, Yuhang and Chen, Liangyu and Chen, Meiqi and Jiao, Yizhu and Li, Xinze and Lu, Xinyuan and Liu, Ziyu and Ma, Yan and Dong, Xiaoyi and others},
  journal={Advances in Neural Information Processing Systems},
  volume={37},
  pages={95963--96010},
  year={2024}
}

@inproceedings{zou2024docbench,
  title         = {{DOCBENCH}: A Benchmark for Evaluating LLM-based Document Reading Systems},
  author={Zou, Anni and Yu, Wenhao and Zhang, Hongming and Ma, Kaixin and Cai, Deng and Zhang, Zhuosheng and Zhao, Hai and Yu, Dong},
  booktitle={Proceedings of the 4th International Workshop on Knowledge-Augmented Methods for Natural Language Processing},
  pages={359--373},
  year={2025}
}

@article{zhu2024mmdocbench,
  title={Mmdocbench: Benchmarking large vision-language models for fine-grained visual document understanding},
  author={Zhu, Fengbin and Liu, Ziyang and Ng, Xiang Yao and Wu, Haohui and Wang, Wenjie and Feng, Fuli and Wang, Chao and Luan, Huanbo and Chua, Tat Seng},
  journal={arXiv preprint arXiv:2410.21311},
  year={2024}
}

@inproceedings{deng2024longdocurl,
  title         = {{LongDocURL}: a Comprehensive Multimodal Long Document Benchmark Integrating Understanding, Reasoning, and Locating},
  author={Deng, Chao and Yuan, Jiale and Bu, Pi and Wang, Peijie and Li, Zhong-Zhi and Xu, Jian and Li, Xiao-Hui and Gao, Yuan and Song, Jun and Zheng, Bo and Liu, Cheng-Lin},
  booktitle={Proceedings of the 63rd Annual Meeting of the Association for Computational Linguistics (Volume 1: Long Papers)},
  pages={1135--1159},
  year={2025}
}

@article{islam2023financebench,
  title={Financebench: A new benchmark for financial question answering},
  author={Islam, Pranab and Kannappan, Anand and Kiela, Douwe and Qian, Rebecca and Scherrer, Nino and Vidgen, Bertie},
  journal={arXiv preprint arXiv:2311.11944},
  year={2023}
}

@article{hendrycks2021cuad,
      title={CUAD: An Expert-Annotated NLP Dataset for Legal Contract Review}, 
      author={Dan Hendrycks and Collin Burns and Anya Chen and Spencer Ball},
      journal={Advances in Neural Information Processing Systems},
      year={2021}
}

@inproceedings{koreeda2021contractnli,
  title         = {{ContractNLI}: A Dataset for Document-level Natural Language Inference for Contracts},
  author={Koreeda, Yuta and Manning, Christopher D},
  booktitle={Findings of the Association for Computational Linguistics: EMNLP 2021},
  pages={1907--1919},
  year={2021}
}

@article{guha2023legalbench,
  title         = {{LegalBench}: A Collaboratively Built Benchmark for Measuring Legal Reasoning in Large Language Models},
    author={Guha, Neel and Nyarko, Julian and Ho, Daniel and R{\'e}, Christopher and Chilton, Adam and Chohlas-Wood, Alex and Peters, Austin and Waldon, Brandon and Rockmore, Daniel and Zambrano, Diego and others},
  journal={Advances in neural information processing systems},
  volume={36},
  pages={44123--44279},
  year={2023}
}

@inproceedings{dasigi2021qasper,
  title         = {A Dataset of Information-Seeking Questions and Answers Anchored in Research Papers},
  author        = {Dasigi, Pradeep and Lo, Kyle and Beltagy, Iz and Cohan, Arman and Smith, Noah A. and Gardner, Matt},
  booktitle={Proceedings of the 2021 Conference of the North American Chapter of the Association for Computational Linguistics: Human Language Technologies},
  pages={4599--4610},
  year={2021}
}

@article{wang2024officebench,
  title={Officebench: Benchmarking language agents across multiple applications for office automation},
  author={Wang, Zilong and Cui, Yuedong and Zhong, Li and Zhang, Zimin and Yin, Da and Lin, Bill Yuchen and Shang, Jingbo},
  journal={arXiv preprint arXiv:2407.19056},
  year={2024}
}

@article{wang2025odysseybench,
  title={Odysseybench: Evaluating llm agents on long-horizon complex office application workflows},
  author={Wang, Weixuan and Han, Dongge and Diaz, Daniel Madrigal and Xu, Jin and R{\"u}hle, Victor and Rajmohan, Saravan},
  journal={arXiv preprint arXiv:2508.09124},
  year={2025}
}

@article{patwardhan2025gdpval,
  title={Gdpval: Evaluating ai model performance on real-world economically valuable tasks},
  author={Patwardhan, Tejal and Dias, Rachel and Proehl, Elizabeth and Kim, Grace and Wang, Michele and Watkins, Olivia and Fishman, Sim{\'o}n Posada and Aljubeh, Marwan and Thacker, Phoebe and Fauconnet, Laurance and others},
  journal={arXiv preprint arXiv:2510.04374},
  year={2025}
}

@misc{databricks2026officeqapro,
  title         = {{OfficeQA Pro}: An Enterprise Benchmark for End-to-End Grounded Reasoning},
  author        = {Opsahl-Ong, Krista and Singhvi, Arnav and Collins, Jasmine and Zhou, Ivan and Wang, Cindy and Baheti, Ashutosh and Oertell, Owen and Portes, Jacob and Havens, Sam and Elsen, Erich and Bendersky, Michael and Zaharia, Matei and Chen, Xing},
  year          = {2026},
  eprint        = {2603.08655},
  archivePrefix = {arXiv},
  primaryClass  = {cs.CL}
}

@misc{garre2026gdppdfbenchmarkinggroundedmultimodal,
      title={GDP.pdf: Benchmarking Grounded Multimodal Reasoning over Professional PDF Documents}, 
      author={Suhaas Garre and Emily Ritchie and Sushant Mehta and Edwin Chen},
      year={2026},
      eprint={2607.11192},
      archivePrefix={arXiv},
      primaryClass={cs.CV},
      url={https://arxiv.org/abs/2607.11192}, 
}

@inproceedings{liu2023geval,
  title         = {{G-Eval}: NLG Evaluation using GPT-4 with Better Human Alignment},
  author        = {Liu, Yang and Iter, Dan and Xu, Yichong and Wang, Shuohang and Xu, Ruochen and Zhu, Chenguang},
  booktitle={Proceedings of the 2023 conference on empirical methods in natural language processing},
  pages={2511--2522},
  year={2023}
}

@artice{zheng2023mtbench,
  title         = {Judging {LLM-as-a-Judge} with {MT-Bench} and Chatbot Arena},
  author        = {Zheng, Lianmin and Chiang, Wei-Lin and Sheng, Ying and Zhuang, Siyuan and Wu, Zhanghao and Zhuang, Yonghao and Lin, Zi and Li, Zhuohan and Li, Dacheng and Xing, Eric P. and Zhang, Hao and Gonzalez, Joseph E. and Stoica, Ion},
  journal={Advances in neural information processing systems},
  volume={36},
  pages={46595--46623},
  year={2023}
}

@misc{verga2024poll,
  title         = {Replacing Judges with Juries: Evaluating LLM Generations with a Panel of Diverse Models},
  author={Verga, Pat and Hofstatter, Sebastian and Althammer, Sophia and Su, Yixuan and Piktus, Aleksandra and Arkhangorodsky, Arkady and Xu, Minjie and White, Naomi and Lewis, Patrick},
  journal={arXiv preprint arXiv:2404.18796},
  year={2024}
}

@inproceedings{min2023factscore,
  title         = {{FActScore}: Fine-grained Atomic Evaluation of Factual Precision in Long Form Text Generation},
  author        = {Min, Sewon and Krishna, Kalpesh and Lyu, Xinxi and Lewis, Mike and Yih, Wen-tau and Koh, Pang Wei and Iyyer, Mohit and Zettlemoyer, Luke and Hajishirzi, Hannaneh},
  booktitle={Proceedings of the 2023 Conference on Empirical Methods in Natural Language Processing},
  pages={12076--12100},
  year={2023}
}

@article{cameronmiller2015cluster,
  title   = {A Practitioner's Guide to Cluster-Robust Inference},
  author  = {Cameron, A. Colin and Miller, Douglas L.},
  journal = {Journal of Human Resources},
  volume  = {50},
  number  = {2},
  pages   = {317--372},
  year    = {2015},
  publisher = {University of Wisconsin Press}
}

@inproceedings{rajpurkar2016squad,
  title={{SQuAD}: 100,000+ Questions for Machine Comprehension of Text},
  author={Rajpurkar, Pranav and Zhang, Jian and Lopyrev, Konstantin and Liang, Percy},
  booktitle={Proceedings of the 2016 Conference on Empirical Methods in Natural Language Processing},
  pages={2383--2392},
  year={2016}
}
}

\section*{Acknowledgements}
We thank our colleagues across Microsoft for valuable discussions and
feedback throughout the development of \OCB{}, including Aman Rathi,
Utkarsh Nagar, Rachel Hill, Josh Ledgard, Tejas Dixit, DJ Cole, Gaurav Jain,
Abhishek Agarwal, Clement Wang, Ryan Adriano, Shobha Venugopal, Siva Mopati, David Conger,
Victor Chakraborty, Jeremiah Kim, Julia Kadie, Piyush Behre, Yaw Amoateng,
Swetha Ranganath, Archana Saseetharan, and Ankur Gupta.

We thank our data vendors---LinkedIn, Mercor, and Micro1---for
providing the domain experts who sourced industry documents,
authored the question--answer pairs, and constructed the rubrics with
atomic assertions for the Domain~Q\&A track.
We are also grateful to Telus Digital for supporting the File~Fidelity
human baseline study. We thank Matthew Ip from Microsoft for help with
account access, and the team at TCS---Santosh Kalokhe, Zaigham Shaikh,
Pavani Boddupalli, Mayank Vijayvargiya, Abhishek Agarwal,
Twinkle Vishwakarma, and Antariksha Gupta---for help with identifying
source URLs, data scraping, and account access.

Finally, we thank the broader Office and AI research communities,
whose prior work on office-document understanding made this benchmark
possible.

\appendix
%
%
%
%
%
%
%
%

\section{Worked Examples}
\label{app:examples}

To make the difference in evaluation density between the two tracks
concrete, we provide one worked example from each.

\paragraph{Domain Q\&A example -- query.}
\begin{quote}
\small
``IQVIA recently reported FY2024 results. The company is evaluating
the \$6.0B acquisition of `HealthDataCo,' a provider of real-world
evidence platforms. Assume the deal closes at the start of FY2025,
funded 50\% with cash and 50\% with new debt.

Base year is FY2024 (HealthDataCo, stand-alone): Revenue \$1.5B,
EBITDA margin 28\%, revenue growth 6\% annually. Capex 3\% of revenue,
depreciation 2\% of revenue, tax rate 25\%.

\noindent\textit{Synergy plan:}
\begin{itemize}
\item Cost synergies: \$150M in 2025, \$250M in 2026, \$300M in 2027.
\item Revenue synergies: \$200M incremental revenue in 2026, \$400M
in 2027, with 40\% EBITDA flow-through.
\item Integration costs: \$200M in 2025 (\$120M cash, \$80M non-cash
write-off) and \$100M in 2026 (\$70M cash, \$30M non-cash write-off).
Treat non-cash items as excluded from FCF.
\item Working capital: model based on balance sheet ratios (Accounts
Receivable = 12\% of revenue, Accounts Payable = 8\% of COGS,
Inventory = 3\% of revenue). Compute year-over-year change in
balances as $\Delta$NWC.
\end{itemize}

Present financial figures in millions, rounded to the nearest whole
million. Percentages should be rounded to one decimal place.

\noindent\textit{Tasks:}
\begin{enumerate}
\item Calculate HealthDataCo's standalone Free Cash Flow for 2024,
then build a projection of Free Cash Flow pro forma with synergies
for 2025--2027. Model revenue, EBITDA, D\&A, EBIT, tax, net income,
capex, depreciation, integration costs (cash vs.\ non-cash), working
capital (using AR, AP, inventory), and free cash flow. Present free
cash flow as a simple bridge from operating profit down to cash
generation each year.
\item Layer in debt funding: \$3B new debt at IQVIA's weighted
average interest rate. Model recursive interest expense (interest
based on average debt balance). Show impact on net income. Use
IQVIA's FY2024 effective tax rate for the interest tax shield; keep
the 25\% tax rate for HealthDataCo operating taxes.
\item Compute ROIC (2027) as NOPAT (2027) divided by average invested
capital across 2026--2027. Define invested capital as \$6{,}000M plus
the working-capital balance tied to revenue synergies (cumulative).
Exclude goodwill and any debt tax shield. Calculate EVA (2027) using
WACC = 8.0\% and the same average invested capital used for ROIC.''
\end{enumerate}
\end{quote}

\paragraph{Source file.}
IQVIA Holdings Inc.\ Form 10-K for the fiscal year ended December 31,
2024, used to source IQVIA's weighted average interest rate and
effective tax rate. HealthDataCo is a hypothetical target whose
stand-alone financials and synergy plan are stated directly in the
prompt.

\paragraph{Selected reference assertions.}
A single response to this query can verify on 49 atomic assertions.
Examples include:
\begin{itemize}
\item States Base Year (2024) Revenue as \$1{,}500M.
\item Calculates Base Year (2024) EBITDA as \$420M (range:
\$400M--\$440M).
\item Calculates Base Year (2024) Tax @ 25.0\% as \$98M (range:
\$93M--\$103M).
\item Calculates Base Year (2024) NOPAT as $\approx$\$293M (range:
\$278M--\$308M).
\item Identifies the interest rate basis as IQVIA's weighted average
interest rate ($\approx$4.69\%).
\item Models interest expense on new debt as (IQVIA weighted average
rate) $\times$ average debt balance; computes $\approx$\$141M.
\item \emph{[\dots\ approximately 43 additional atomic assertions]}
\end{itemize}

\begin{figure}[tbp]
\centering
\includegraphics[width=\linewidth]{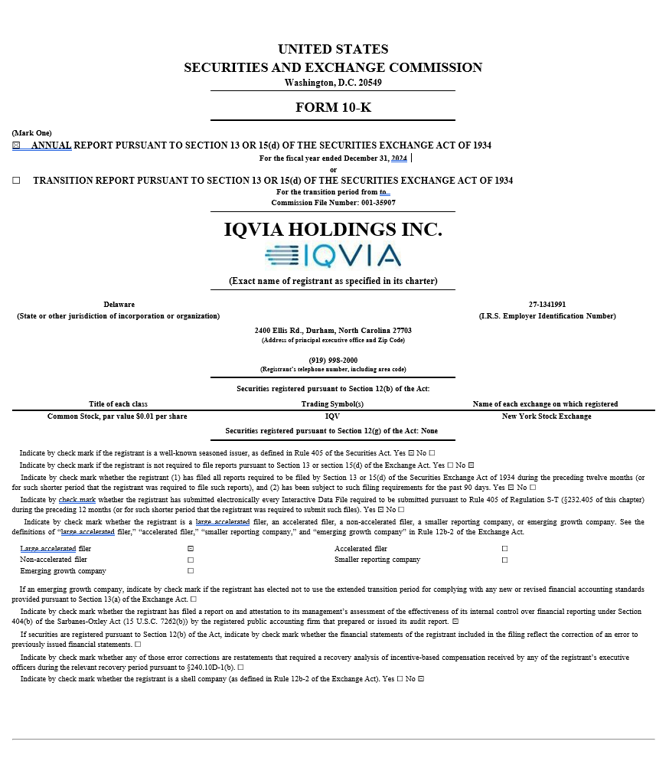}
\caption{Source document excerpt for the Domain Q\&A worked example.}
\label{fig:domain-example}
\end{figure}

This illustrates why Domain Q\&A averages $\sim$45 assertions per
query: a single realistic professional analysis produces a structured
multi-paragraph response with many independently verifiable claims,
spanning factual extraction, multi-step computation, and synthesis
under stated assumptions.

\paragraph{File Fidelity Q\&A example -- query.}
\begin{quote}
``Which model has the highest presentation quality score for
\texttt{Amazon\_Toys\_and\_Games} with NMF recommendation model and what
is its score?''
\end{quote}

\paragraph{Source file.}
A PowerPoint slide showing offline evaluation results for several
recommendation models across two datasets, with separate metric tables
for presentation quality ($M_c$) and model explainability ($M_e$).

\paragraph{Reference assertions.}
\begin{itemize}
\item Response should state that, for the
\texttt{Amazon\_Toys\_and\_Games} dataset using the NMF recommendation
model, the model with the highest presentation quality score is the
researchers' proposed model labeled `Ours'.
\item Response should specify that the highest presentation quality
score for the \texttt{Amazon\_Toys\_and\_Games} dataset using the NMF
recommendation model is $0.025$. Equivalent expressions such as
$0.0250$ should also be accepted.
\end{itemize}

\begin{figure}[tbp]
\centering
\includegraphics[width=0.95\linewidth]{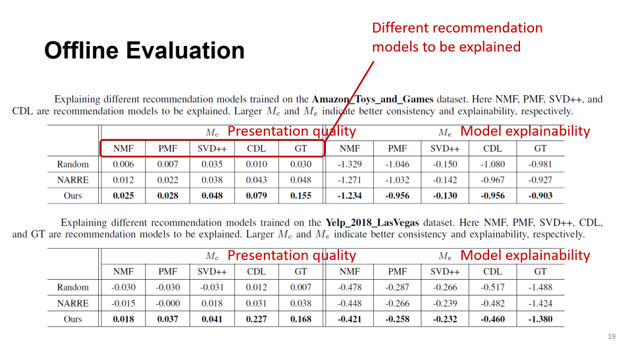}
\caption{Source slide for the File Fidelity worked example.}
\label{fig:file-fidelity-example}
\end{figure}

This is typical of File Fidelity: a single-artifact query producing two
atomic assertions, both grounded directly in a chart cell.


\section{LLM-as-a-Judge Prompt}
\label{app:judge-prompt}

This appendix reproduces verbatim the prompt used by each judge model
when evaluating an atomic assertion. The judge receives the query, the
system response, and a single reference assertion, and returns a binary verdict
together with structured reasoning. The final decision per assertion
is obtained by majority vote across three judge models, as described
in \S\ref{sec:llm-judge}. Four worked examples calibrate the judge
against the partial-fulfillment, equivalent-information, and
explicit-phrasing rules.

\begin{PromptBlock}
<role>
You are an expert quality assurance evaluator. Your task is to impartially verify the faithfulness of a AI assistant's ("<ai_response>") response against a given assertion ("<assertion>") in the context of a specific question ("<question>").
Your only source of truth is the content provided in the <question> and the <ai_response>.
Do not use any external knowledge or information beyond what is provided in the <question> and <ai_response> to make your evaluation.
You will be provided with a <question>, an <ai_response>, and an <assertion>.
You should analyze the AI response to determine if it satisfies the criteria outlined in the <assertion> in the context relevant to the <question> being asked.
</role>

<assertion_interpretation>
The assertions you will receive are statements relating the <ai_response> to the <question> being evaluated.

For example:

<assertion_interpretation_example_1>
query: "How do you initially measure the financial instruments under IFRS 9 and what is the fair value option?"
assertion: "Explains measurement effect of fair value option on own credit risk."
interpretation: "The AI response must explain the fair value option on own credit risk in the context of IFRS 9 financial instruments"
</assertion_interpretation_example_1>

<assertion_interpretation_example_2>
query: "If we are the owner of a company that is holding on its property a substantial share package for another publicly owned company, how important is that our financial reporting is made according to IAS 33? If we own non-public shares, is IAS 33 still relevant?"
assertion: "State that the reporting entity's own listing status is the primary driver of IAS 33 relevance."
interpretation: "The AI response must state the fact that reporting entity's own listing status is the primary driver of IAS 33 relevance in the context of IAS 33 financial reporting."
</assertion_interpretation_example_2>

</assertion_interpretation>

<input>
  <question>{question}</question>
  <ai_response>{ai_response}</ai_response>
</input>

<assertion>
{assertion}
</assertion>

<reasoning_instructions>

1. Read the <question>, <ai_response> and <assertion> provided thoroughly. Interpret the context of the <assertion> based on the <question> provided. Make sure you understand what needs the ground truth from the assertion that needs to be evaluated.
2. Analyze the <ai_response> and locate evidence relevant to the <assertion>, either to support or refute it.
3. Analyze the <question> and locate the relevant context that qualifies the <assertion> in order to clarify its requirement.
4. Evaluate the <assertion> based on the evidence you identified.
   - Consider partial fulfillment: if the <ai_response> meets some but not all criteria, requirements or expectations in the <assertion>, check if the <assertion> allows for partial correctness (e.g., "response should contain at least X out of Y conditions", "response should X OR Y"). If it does, you may consider it TRUE; otherwise, it is FALSE.
   - Consider equivalent information: if the <ai_response> does not directly fulfill the criteria but conveys the same meaning using different wording, consider the <assertion> TRUE. If the meaning is not equivalent or the <assertion> explicitly requires specific wording, consider it FALSE.
   - Ignore additional information: if the <ai_response> contains extra details not mentioned in the <assertion>, disregard them unless the <assertion> explicitly prohibits such additional information.
   - Do not evaluate assertions on their own, you must consider whether the assertion is verifiable in the context of the <question> and the <ai_response>, even if a trivial fact is requested in the <assertion>.
   - Do not infer explicit requirements in your interpretation of assertions unless observed directly. For example, if the assertion says "State that [sentence describing the requirement]", you must not infer that the response must explicitly state exactly what it is written as the requirement description or sentence describing the required statement.
   - Only assess explicit requirements if the assertion describes it as an explicit requirement or asks for exactness of certain words or data points (for example: by placing expected content in the ai response in quotes).
5. Compose your reasoning using the following structure xml-like structure:
   <assertion>
   Restate the <assertion> verbatim as provided
   </assertion>
   <interpretation>
   Interpretation of the assertion in the context of the <question> asked as a single statement.
   </interpretation>
   <evidence>
   [Quoted evidence from the <question> and <ai_response> that supports your conclusion as a list, if applicable. If no specific quotes are used, write a list containing a single entry "N/A"]
   </evidence>
   <analysis>
   Detailed explanation of how the <ai_response> meets or does not meet the criteria outlined based on the evidence you found
   </analysis>
   <verdict>
   The assertion is TRUE/FALSE
   </verdict>
6. Fill in the final output as a valid JSON object (without markdown code blocks) with your reasoning and score. Only valid JSON is allowed in the output.
   </reasoning_instructions>

<output>
{
    "reasoning": "Structured explanation as described above",
    "score": "0 if assert is FALSE, 1 if assert is TRUE"
}
</output>

<examples>
    <example_1>
    <input>
    <question>"A mid-sized UK distributor currently operates with 60 Days of Inventory on Hand (DOH) and an On-Time In-Full (OTIF) delivery rate of 87%. Management wants to reduce DOH to 45 days over the next 12 months while increasing OTIF to 92%, without raising stockout risk.

Stage 3: Financial & Service-Level Calculation:
Assume:
Average Inventory Value = \pounds{}5 million
Cost of Capital = 10%
Forecast Error Reduction from selected strategies improves OTIF by 3 percentage points.

a) Calculate the annual working capital savings from reducing DOH from 60 to 45 days.
b) If each 1% OTIF improvement increases annual revenue by \pounds{}50,000, estimate the additional revenue from the forecast accuracy gain.
c) Combine the two results to show the total annual benefit."</question>
<ai_response>### Stage 3: Financial & Service-Level Calculation

**a) Working Capital Savings**
Reducing DOH from 60 to 45 days means a 25% reduction in inventory. New inventory = \pounds{}5,000,000 x (45/60) = \pounds{}3,750,000. Capital freed = \pounds{}1,250,000. Annual savings at 10% cost of capital = **\pounds{}125,000**.

**b) Revenue from OTIF Improvement**
A 3-percentage-point OTIF gain at \pounds{}50,000 per point = 3 x \pounds{}50,000 = **\pounds{}150,000**.

**c) Total Annual Benefit**
\pounds{}125,000 + \pounds{}150,000 = **\pounds{}275,000 per year**.</ai_response>
</input>
<assertion>
In the Conclusion, state that, based on prompt-provided assumptions, the combined strategies yield an estimated \pounds{}275,000 annual benefit.
</assertion>
<output>
{
"reasoning": "<assertion>In the Conclusion, state that, based on prompt-provided assumptions, the combined strategies yield an estimated \pounds{}275,000 annual benefit.</assertion><interpretation>The AI response must arrive at a combined total annual benefit of \pounds{}275,000 from both the working capital savings and OTIF revenue uplift, using the numeric assumptions given in the question.</interpretation><evidence>['\pounds{}125,000 + \pounds{}150,000 = **\pounds{}275,000 per year**']</evidence><analysis>The assertion requires a combined benefit figure of \pounds{}275,000. The AI response correctly computes working capital savings (\pounds{}125,000) and OTIF revenue uplift (\pounds{}150,000), arriving at a combined \pounds{}275,000. The arithmetic is consistent with the prompt-provided assumptions (DOH 60->45, \pounds{}5M inventory, 10% cost of capital, 3pp x \pounds{}50K). The assertion is satisfied.</analysis><verdict>The assertion is TRUE</verdict>",
"score": 1
}
</output>
</example_1>
<example_2>
<input>
<question>"One of the largest processing conglomerates is implementing an advanced digital transformation strategy to improve its operational efficiency and secure an information advantage (Veronesi, 2021). The company has invested big on obtaining IoT sensors and Edge AI to drive a new predictive maintenance system for its old equipment. Nevertheless, this project has also revealed a range of strategic and technical issues.

When the company is in a situation where a new predictive maintenance system is in use and suffers an 'integration gap' between new IT-based analytics and all the existing Operational Technology (OT) data (ManufacturingTomorrow, 2025), how should it address this challenge?"</question>
<ai_response>To address the integration gap, the company should adopt a unified IT/OT architecture. Rather than treating enterprise IT systems and shop-floor OT systems as independent silos, the organization should converge them into a single operating model. This means establishing middleware or integration platforms that allow real-time OT sensor data to flow into the company's analytics and ERP layers. By bridging the two domains, the organization can unlock the full potential of its predictive maintenance investment without requiring wholesale replacement of legacy OT infrastructure.</ai_response>
</input>
<assertion>
State that addressing the integration gap requires moving from segregated IT/OT models toward an integrated operating model.
</assertion>
<output>
{
"reasoning": "<assertion>State that addressing the integration gap requires moving from segregated IT/OT models toward an integrated operating model.</assertion><interpretation>The AI response must recommend transitioning from separate IT and OT domains to an integrated operating model in the context of addressing the manufacturing integration gap.</interpretation><evidence>['Rather than treating enterprise IT systems and shop-floor OT systems as independent silos, the organization should converge them into a single operating model']</evidence><analysis>The assertion requires the response to advocate for moving from segregated IT/OT models to an integrated one. The AI response uses different wording ('independent silos' instead of 'segregated models', 'converge them into a single operating model' instead of 'integrated operating model') but conveys the same meaning. The equivalent information rule applies--the recommendation is substantively identical.</analysis><verdict>The assertion is TRUE</verdict>",
"score": 1
}
</output>
</example_2>
<example_3>
<input>
<question>In today's business environment, the supply chain begins with the customer's initial demand or purchase request, flows through sourcing, production, and distribution, and extends to final delivery and post-sale activities such as returns, recycling, and customer support. Along the way, businesses collect vast amounts of data, including inventory, production, quality, and customer related information. This raises two critical questions: How is this data being managed, and how is it being safeguarded?</question>
<ai_response>The report describes how supply chain data is managed through ERP systems, warehouse management systems, and increasingly through IoT-connected devices that provide real-time tracking. For safeguarding, the report outlines a comprehensive cybersecurity control framework including network segmentation, zero-trust architecture, multi-factor authentication, and regular penetration testing as recommended practices for protecting supply chain data from cyber threats.</ai_response>
</input>
<assertion>
State that the report does not present a comprehensive cybersecurity control framework.
</assertion>
<output>
{
"reasoning": "<assertion>State that the report does not present a comprehensive cybersecurity control framework.</assertion><interpretation>The AI response must acknowledge that the source report does not contain a comprehensive cybersecurity control framework, rather than fabricating one.</interpretation><evidence>['the report outlines a comprehensive cybersecurity control framework including network segmentation, zero-trust architecture, multi-factor authentication, and regular penetration testing']</evidence><analysis>The assertion requires the response to note a limitation of the source report--that it does not present a comprehensive cybersecurity framework. The AI response does the opposite: it claims the report 'outlines a comprehensive cybersecurity control framework' and lists specific controls (network segmentation, zero-trust, MFA, penetration testing). This directly contradicts the assertion and represents hallucinated content beyond what the source report actually covers.</analysis><verdict>The assertion is FALSE</verdict>",
"score": 0
}
</output>
</example_3>
<example_4>
<input>
<question>Deloitte (2024) in their report titled "IFRS in your pocket" provides that the main objective of IAS 12 is to prescribe the accounting treatment for income taxes. Provide the difference between current tax and deferred tax. What is the difference between Deferred Tax Assets (DTA) and Deferred Tax Liabilities (DTL)? Provide the measurement of deferred tax, and the presentation of current and deferred tax.</question>
<ai_response>**Current Tax vs Deferred Tax:** Current tax is the amount of income taxes payable in the current period, while deferred tax relates to temporary differences between carrying amounts and tax bases.

**DTA vs DTL:** A Deferred Tax Liability arises when a temporary difference will result in taxable amounts in future periods. A Deferred Tax Asset arises when a temporary difference will result in deductible amounts in future periods.

**Measurement:** Deferred tax is measured using the tax rates that are expected to apply when the asset is realised or the liability is settled.

**Presentation:** Current and deferred tax is recognised as income or expense in profit or loss for the period.</ai_response>
</input>
<assertion>
State that DTAs are recognized only to the extent that future taxable profit is probable.
</assertion>
<output>
{
"reasoning": "<assertion>State that DTAs are recognized only to the extent that future taxable profit is probable.</assertion><interpretation>The AI response must state the IAS 12 recognition constraint that Deferred Tax Assets may only be recognized when it is probable that sufficient future taxable profit will be available against which the deductible temporary differences can be utilized.</interpretation><evidence>['A Deferred Tax Asset arises when a temporary difference will result in deductible amounts in future periods']</evidence><analysis>The assertion requires an statement about the probability-based recognition threshold for DTAs. The AI response only defines when a DTA arises (deductible amounts in future periods) but does not mention the critical recognition constraint--that DTAs are recognized only to the extent that future taxable profit is probable. The definition of what creates a DTA is not the same as stating the conditions under which it may be recognized. The specific requirement of the assertion is absent.</analysis><verdict>The assertion is FALSE</verdict>",
"score": 0
}
</output>
</example_4>
</examples>
\end{PromptBlock}

\section{Construction Methodology --- Full Pipelines}
\label{app:construction}

This appendix expands the high-level overview in
\S\ref{sec:construction} with the full per-track construction
pipelines. \S\ref{app:construction-domainqa} documents the Domain
Q\&A pipeline; \S\ref{app:construction-filefidelity} documents the
File Fidelity Q\&A pipeline. Both pipelines share the same four-stage
shape (source curation, question authoring, rubric construction,
quality assurance), but instantiate each stage differently. Cross-
references to related appendices, expert pool composition
(\S\ref{app:experts}), the three rubric-authorship methodologies
(\S\ref{app:methodologies}), and the assertion-quality review loop
(\S\ref{app:rubric-process}), are made where appropriate.

\subsection{Domain Q\&A pipeline}
\label{app:construction-domainqa}

Domain Q\&A queries are expert-level questions grounded in real industry
documents, requiring multi-step domain reasoning, computation, and often
synthesis across multiple files. Unlike File Fidelity queries, which target
single artifacts with one or two assertions, Domain queries average
$\sim$45 atomic assertions per query, reflecting the depth of analytical
reasoning involved. A single Domain query - for instance, preparing an
earnings call analysis from a 10-Q - may require pulling line items,
computing ratios, decomposing variances, and synthesizing observations into
a structured response.

\paragraph{Construction pipeline.}
Domain Q\&A items are produced through a four-stage construction pipeline:

\begin{enumerate}
\item \textbf{Source curation.} Domain practitioners select or compile
authoritative source documents that a qualified professional would consult.
Sources span SEC filings, slide decks, spreadsheets, and operational
documents.
\item \textbf{Question and answer authoring.} Experts draft realistic
prompts grounded in the source documents and produce reference answers
that a qualified practitioner would recognize as correct and complete.
Each Q\&A is authored by 2-3 domain experts.
\item \textbf{Rubric construction.} Each Q\&A pair is paired with a rubric
composed of atomic assertions - single, verifiable claims that a correct
response should satisfy. We use multiple authoring approaches: rubrics
authored independently of the reference answer (capturing what any correct
response should demonstrate), rubrics derived from the reference answer
(decomposing it into verifiable claims), and rubrics authored by domain
experts independent of the original Q\&A author. Each approach has
different evaluative properties; together they reduce the benchmark's
dependence on any single rubric-construction convention.
\item \textbf{Quality assurance.} A separate reviewer, distinct from the
original prompt and rubric authors, verifies prompt construction, rubric
design, and consistency across items.
\end{enumerate}

Model evaluation against these rubrics is not part of the construction
pipeline; it is the downstream evaluation procedure described in
\S\ref{sec:eval}.

\paragraph{Scoring.}
We report \textbf{assertion-level accuracy} as the headline metric for
the Domain Q\&A track: a model's score is the total number of atomic
assertions it satisfies, divided by the total number of assertions in
the track (see \S\ref{sec:eval} for the precise formulation). Every
atomic assertion contributes equally, so queries with more assertions
have proportionally more influence on the score.

Each assertion also carries an expert-assigned importance weight, but
schema heterogeneity and uncalibrated rater severity make weighted
scores interpretable only within a single rubric rather than as an
absolute signal across the benchmark; we treat weighted scores as a
secondary signal (reported in \S\ref{app:weighted}, where they agree
with unweighted scores on model rankings).

\subsection{File Fidelity Q\&A pipeline}
\label{app:construction-filefidelity}

File Fidelity Q\&A tests whether a system can accurately perceive and
interpret the structural and visual content of Office files - text, tables,
charts, images, formatting, and app-specific artifacts. The authoring
methodology applies a common four-stage pipeline across all three
applications (Word, Excel, PowerPoint), with app-specific adaptations at
each stage.

\paragraph{Document sourcing and curation.}
For all three applications, source documents are drawn from a broad pool of
publicly available files (Fortune 500 filings, government agency
publications, educational institutions). We apply coarse filtering to remove
corrupted files, near-empty documents, and pure templates, and keep only
English-language documents. App-specific implementation choices,
document-to-text conversion (Word via Markdown, Excel via per-sheet
structured text, PowerPoint via slide-level extraction) and the modality
switches that support groundedness filtering, are documented in
\S\ref{app:filefidelity-details}.

\emph{Iterative coverage targeting.}
Document selection is iterative: we begin with a broad pool, generate
candidate questions (described below), and revisit document selection if
coverage of any artifact type or app-specific element is sparse. The goal is
for the final corpus to support meaningful question generation across all
the artifact types in our taxonomy. We track page counts (Word), slide
counts (PowerPoint), and sheet counts (Excel) as auxiliary metadata to
support stratified sampling.

\emph{Human verification.}
Human annotators review the selected documents to confirm structural diversity (e.g., headings,
tables, images, text boxes, headers/footers, charts, formulas, multi-sheet
structure), and validate that documents contain sufficiently rich content to
support meaningful question generation.

\paragraph{LLM-assisted Q\&A generation.}
\label{sec:qagen}
Questions target information explicitly present in the document across
different modalities and structural elements. We generate questions grounded
in multiple artifact categories: text, tables, images, charts, and
app-specific elements (Word: headers, footers, text boxes, equations,
hyperlinks, font styles; Excel: formulas, cross-sheet references, named
ranges; PowerPoint: slide layouts, speaker notes, embedded charts).

\emph{Q\&A generation.}
We use frontier LLMs with prompts designed to generate candidate
question--answer pairs grounded in specific document elements. For text,
tables, images, and charts, the generation prompt emphasizes creating
questions that require reading and understanding the targeted element. For
app-specific structural elements, we explicitly extract formatting and
layout components and generate questions that test understanding of
document structure and presentation. The prompt design discourages reliance
on general knowledge or other parts of the document. Human curators then
review, edit, and supplement the LLM-generated candidates. The full
generation prompts are reproduced in \S\ref{app:qa-generation-prompts}.

\emph{Writing guidelines.}
A portion of items are authored or substantially rewritten by hand,
for example, when curators edit LLM-generated candidates during review,
or when under-covered artifact types require questions written from
scratch. For both LLM-generated and manually-authored questions, we
enforce two guidelines: questions must be (1)~\textbf{grounded},
answerable solely from the file, and (2)~free from \textbf{underspecification},
minimizing cases with multiple equally valid answers unless explicitly
allowed. The same guidelines inform the prompts used in LLM-assisted
generation, so manually and automatically authored questions remain
stylistically consistent.

\paragraph{Automatic filtering for groundedness.}
Each candidate question undergoes a multi-stage automatic filtering pipeline
designed to ensure the question genuinely requires the targeted document
element:

\begin{itemize}
\item \textbf{No-context check.} We prompt the LLM to answer the question
without any document context. Questions answerable from the model's training
knowledge are discarded.
\item \textbf{Element-specific checks.} The filtering strategy adapts to
artifact type. For text-based questions, we provide the full document with
the source chunk removed. For table questions, we provide the document with
all tables except the target table, and separately test with text-only
content (no tables). For image-based questions, we provide text-only
document content (no images). For app-specific element questions, we provide
the document without the specific structural element being tested
(e.g., without headers/footers, without specific formulas, or without
specific slide layouts). Questions correctly answerable without the target
element are discarded.
\item \textbf{Quality grading.} Surviving question--answer pairs are
evaluated by an LLM grader calibrated to human quality judgments, scoring
each Q\&A across three dimensions: question clarity, answer precision, and
information value.
\end{itemize}

\paragraph{Human verification and quality control.}

\emph{Multi-annotator review.}
Each generated Q\&A pair is independently evaluated by two human
annotators. Annotators verify answer correctness, locate and mark the
specific document element containing the evidence (text span, table cell,
image region, formula, slide region, or structural component), and grade
question quality. When the two annotators disagree on correctness or
quality, a third annotator adjudicates and provides the final judgment.

\emph{Aggregation and quality filtering.}
For Q\&A pairs where the two initial annotators agree, the agreed
judgment is taken as final. For pairs that required a third annotator,
the adjudicator's decision is used. Q\&A pairs where annotators cannot
reliably locate supporting evidence are discarded.

\emph{Iterative refinement.}
Questions flagged during human review are revised or discarded. Common
revisions include narrowing overly broad questions, adding specificity to
eliminate ambiguity, correcting reference answers based on annotator
evidence, and updating questions where multiple interpretations exist.
Reference answers may consist of a single atomic statement or several;
in either case they are stored as a collection of atomic statements
that the evaluation procedure (\S\ref{sec:eval}) assesses
independently.


\section{Domain Q\&A Collection Methodologies}
\label{app:methodologies}

The Domain Q\&A track was assembled across three methodologies that
differ along the rubric-authorship dimension. Table~\ref{tab:methods}
maps each onto the four-stage construction pipeline described in
\S\ref{sec:domainqa}. We organize by rubric authorship rather than by
vendor, since this is the dimension along which the collections'
evaluative properties differ.

\begin{table*}[t]
\centering
\scriptsize
\caption{Per-methodology instantiation of the Domain Q\&A pipeline.}
\label{tab:methods}
\renewcommand{\arraystretch}{1.15}
\setlength{\tabcolsep}{3pt}
\resizebox{\textwidth}{!}{%
\begin{tabular}{@{}p{2.2cm}p{4.8cm}p{4.8cm}p{4.8cm}@{}}
\toprule
\textbf{Stage}
  & \textbf{Independent rubric}
  & \textbf{Cross-org layered}
  & \textbf{Derived from golden} \\
\midrule
Domains
  & Investor Relations, FP\&A
  & Supply Chain, Manufacturing, Accounting
  & 12 industries \\
Source curation
  & SMEs select company; compile public filings (10-K, 10-Q, transcripts)
  & Experts curate documents across Word, PPT, Excel; peer and human-data-manager review
  & Source documents provided to experts as Q\&A grounding \\
Q\&A authoring
  & SME drafts realistic prompt; expert authors golden answer from sources only
  & Experts author prompts and golden responses grounded in the curated source documents
  & Two-expert authoring: Expert A writes Q + ideal A; Expert B answers independently, then synthesizes both into a single golden answer \\
Rubric construction
  & Model-agnostic rubric authored \emph{independently} of the golden response; criteria span knowledge verification, logical deduction, quantitative reasoning
  & Atomic rubrics authored by a \emph{different organization} than the Q\&A authors; cross-org separation reduces anchoring on the original author's framing
  & Atomic rubrics \emph{derived from} the golden answer by decomposition into atomic assertions \\
Weighting
  & Primary / secondary per assertion
  & Per-criterion, normalized post-hoc
  & Per-criterion, normalized post-hoc \\
QA review
  & Independent reviewer verifies prompts, rubrics, scoring
  & Leads verify rubric consistency and clarity
  & Lead verifies clarity, consistency, grammar across both authoring phases \\
\bottomrule
\end{tabular}%
}
\end{table*}

The cross-org layered methodology is worth flagging explicitly: the
Q\&A authors and the rubric authors came from different expert
organizations. This separation produces rubrics that reflect what an
independent expert considers necessary and sufficient for a correct
response, rather than a decomposition of the original author's framing.

\begin{figure}[h]
\centering
\includegraphics[width=0.95\linewidth]{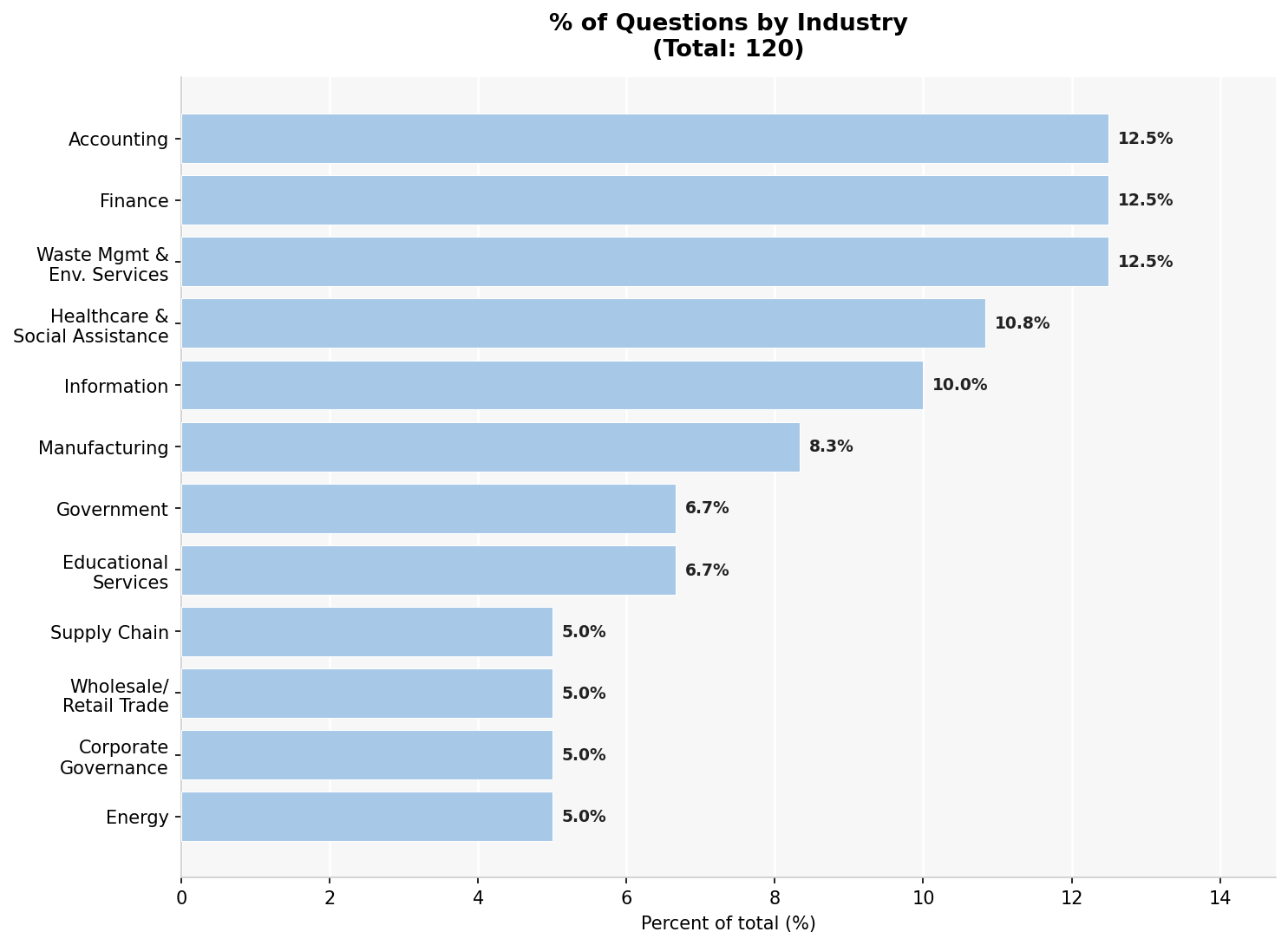}
\caption{Distribution of Domain Q\&A queries across the 12 industries
covered by the benchmark (total: 120 queries).}
\label{fig:domain-qna-industry}
\end{figure}

\paragraph{Per-industry query distribution.}
\label{app:domain-distribution}
Figure~\ref{fig:domain-qna-industry} shows the distribution of Domain
Q\&A queries across the 12 industries (referenced in main paper
\S\ref{sec:overview}). Per-industry shares range from $5.0\%$ to
$12.5\%$ of the $120$ queries; coverage is roughly balanced rather
than weighted by economic contribution.

\section{Rubric Construction Process}
\label{app:rubric-process}

This appendix describes the iterative process used to refine
domain-expert-authored assertions into atomic, LLM-evaluable rubric
items. \S\ref{sec:assertionquality} summarized the failure modes we
encountered; here we provide the full taxonomy with examples and
describe the review loop.

\subsection{Review loop}

Each rubric was passed through a three-step review loop:

\begin{enumerate}
\item \textbf{Automated triage.} Every candidate assertion was scored by
a frontier LLM along the failure-mode dimensions enumerated below. The
LLM output a structured flag set (e.g., \texttt{\{compound: true,
ambiguous: false, ...\}}) and a brief justification per flag. We did
not act on these flags directly; they served only to prioritize human
review.
\item \textbf{Expert decision.} A domain expert (the rubric author or a
peer reviewer) examined each flagged assertion and selected one of:
\textit{keep} (false positive), \textit{modify} (rewrite for clarity or
specificity), \textit{split} (decompose a compound assertion into
multiple atomic ones), \textit{merge} (collapse parallel assertions
into one general claim), or \textit{delete} (remove off-topic or
unverifiable assertions).
\item \textbf{Cross-check.} An independent reviewer re-ran the
automated triage on the revised rubric and resolved any remaining
flags. Rubrics typically required two to three loop iterations to
stabilize.
\end{enumerate}

\subsection{Failure mode taxonomy with examples}

Examples below are drawn from real rubrics across our domain authoring
collections. Identifying details have been preserved where they
illustrate the failure mode and abstracted otherwise.

\paragraph{(i) Duplicate assertions.}
Two assertions express the same claim in different surface forms,
inflating the score for any model that produces the underlying claim
once.

\begin{itemize}
\item \emph{Before:}
\begin{itemize}
\item Identifies ID.RA-09 as justification.
\item States NIST CSF 2.0 Function ID.RA-09 as defensible security
posture.
\end{itemize}
\item \emph{After:} States NIST CSF 2.0 Function ID.RA-09 as
defensible security posture.
\end{itemize}

\paragraph{(ii) Compound assertions (split).}
A single assertion bundles two or more independently verifiable claims,
preventing binary grading. This is the inverse of the mergeable case
in (iii): a compound assertion is \emph{split} into several atomic
assertions, whereas mergeable parallel assertions are \emph{merged}
into one; the two operations move in opposite directions to calibrate
assertion granularity.

\begin{itemize}
\item \emph{Before:} States that CO\textsubscript{2} injection is less
associated with carbon removal from a climate-change perspective
because the duration of storage may be shorter and the injection is
typically used to recover more oil and gas.
\item \emph{After (split into two):}
\begin{itemize}
\item States that CO\textsubscript{2} injection is less associated
with carbon removal because the duration of storage may be shorter.
\item States that CO\textsubscript{2} injection is less associated
with carbon removal because the injection is typically used to
recover more oil and gas.
\end{itemize}
\end{itemize}

\paragraph{(iii) Mergeable parallel assertions.}
Several near-identical assertions vary only on a single slot value;
collapsing them into a single more general claim removes redundancy
without weakening the rubric.

\begin{itemize}
\item \emph{Before:}
\begin{itemize}
\item Identifies untrusted devices should only have view access.
\item Identifies unmanaged devices should only have view access.
\end{itemize}
\item \emph{After (merged):} Identifies that untrusted or unmanaged
devices should be restricted to view access.
\end{itemize}

\paragraph{(iv) Ambiguous assertions.}
The assertion reads correctly when surrounded by the reference answer's
context but becomes unverifiable once extracted, because the relevant
qualifiers (subject, scope, condition) are implicit.

\begin{itemize}
\item \emph{Before:} States downloads should be blocked.
\item \emph{After:} States access/downloads of sensitive documents
should be blocked on non-compliant endpoints.
\end{itemize}

\begin{itemize}
\item \emph{Before:} Defines attack surface properly.
\item \emph{After:} States that any endpoint that can authenticate to
organizational resources becomes part of the organization's attack
surface.
\end{itemize}

\paragraph{(v) Broad assertions.}
The assertion is too general to discriminate between a substantive
answer and a superficial one - both pass.

\begin{itemize}
\item \emph{Before:} States the board's structure can limit the CEO's
power.
\item \emph{After:} States that the current board structure (CEO plus
two private investors) can still provide balance to the CEO's
influence through governance mechanisms.
\end{itemize}

\paragraph{(vi) Structural / presentation assertions.}
The assertion grades the form of the response rather than its content
- whether it begins with a paragraph, uses bullet formatting, or
adopts particular terminology. We removed these from the rubric
entirely: presentation requirements belong in the prompt, not the
graded content rubric.

\begin{itemize}
\item \emph{Removed examples:}
\begin{itemize}
\item Response begins with an introductory paragraph.
\item Presents technical recommendations as a list.
\item Presents answer using terminology appropriate for security
practitioners.
\end{itemize}
\end{itemize}

\paragraph{(vii) Prompt-restating assertions.}
The assertion tests information already established by the question
itself, rewarding models for reproducing prompt content rather than
producing answer content.

\begin{itemize}
\item \emph{Removed example:} States there are three members on the
Board currently. \emph{(The question already specified the board
size; verifying this in the response tests reading comprehension of
the prompt, not domain reasoning.)}
\end{itemize}

\paragraph{(viii) Inappropriate illustrative-example assertions.}
The assertion demands one specific illustrative example when many are
equally valid, penalizing correct responses that produce a different
example.

\begin{itemize}
\item \emph{Before:} States to increase percentage of CSBG participants
achieving self-sufficiency from 18\% to 35\% over five years.
\item \emph{After:} Provides a concrete, measurable example of a
family-level self-sufficiency goal with specific targets and
timeframes.
\end{itemize}



\section{Expert Pool Composition}
\label{app:experts}
 
Domain Q\&A items were authored by domain practitioners across
finance, accounting, manufacturing, supply chain, healthcare,
cybersecurity, construction, government, education, and other
industries. Across the benchmark, experts averaged approximately 10
years of professional experience in their domains, with a substantial
fraction having more than 10 years and roughly a quarter having more
than 15 years. The pool included 49 experts spanning 12 industries
contributing to the broader rubric authoring effort, with a minimum of
two qualified annotators per domain.
 
\paragraph{Vendor-managed expert vetting.}
Experts were sourced and vetted by our annotation vendors, each of
which operates a structured selection workflow. Vendor-managed vetting
included verifying domain alignment through skill assessments, AI- or
recruiter-led interviews to confirm depth in the relevant practice
area, soft-skills evaluation to ensure experts could produce clear
written prompts and reference answers, and review of prior employer
credentials and seniority.
 
\paragraph{Representative employers.}
Across the 12 industries covered by the benchmark, experts had prior
or current experience at organizations including PwC, BDO, Amazon,
Mercedes-Benz, Samsung, Continental Automotive, Kaiser Permanente,
UnitedHealth Group, Palantir, U.S.\ Department of Defense, FEMA, CDC
Foundation, UPS, Nike, Hyatt Hotels, and Target, among others.
 
\paragraph{Industry coverage.}
The Domain Q\&A track covers 12 industries: Accounting; Corporate
Governance; Educational Services; Energy; Finance; Government;
Healthcare \& Social Assistance; Information; Manufacturing; Supply
Chain; Waste Management \& Environmental Services; and Wholesale/Retail
Trade.

\section{Q\&A Generation Prompts}
\label{app:qa-generation-prompts}

This appendix documents the prompts used to generate question--answer
pairs across the File Fidelity track. Four prompts target distinct
content modalities: (i) textual content extracted from document body
text, (ii) visual content from figures and charts, (iii) tabular
content, and (iv) Word-specific fidelity attributes (formatting,
metadata, structural elements). Each prompt is parameterized by a
generation budget (e.g., \texttt{\{\{QUESTIONS\_PER\_CHUNK\}\}}) that
controls density per artifact. The prompts shown below are simplified
from production versions, with redundant few-shot examples and
boilerplate omitted; the core task specification, quality constraints,
and output schema are preserved verbatim.

\subsection{Textual Q\&A Generation Prompt}

Operates over text chunks extracted from document body content. To
reduce hallucination, three candidate answers (detailed, terse,
balanced) are generated independently and reconciled via majority vote
before the final pair is emitted.

\begin{PromptBlock}
<role>
You extract explicit, verifiable facts from a document and generate
question-answer pairs grounded in those facts.
</role>

<task>
1. Identify facts explicitly stated in the document (timelines, named
   entities, roles, system features, numeric values). Skip anything
   that requires inference.

2. Generate up to {{QUESTIONS_PER_CHUNK}} high-quality questions.
   Prioritize questions that:
   - Are unambiguous with clear scope
   - Are answerable from the provided text alone
   - Target high-value content: KPIs, financial figures, technical
     specifications, research findings, actionable insights
   - Expect atomic, extractable data points (numbers, dates, names)

3. For each question, draft three candidate answers:
   - Detailed: thorough explanation with context
   - TL;DR:    concise factual one-liner
   - Balanced: key fact plus brief rationale
   Reconcile via majority vote. Final answer is short (one sentence
   or a few words).
</task>

<restrictions>
- Use only information explicitly stated in the document.
- Do not speculate or infer.
- If no explicit facts are found, return {"qas": []}.
</restrictions>

<output>
{
  "qas": [
    {"question": "...", "answer": "..."}
  ]
}
</output>

<input>
<|startoftext|>
{{chunk}}
<|endoftext|>
</input>
\end{PromptBlock}

\subsection{Visual Q\&A Generation Prompt}

Targets figures, charts, and other visual artifacts. The first stage
filters out non-informative images (logos, decorative elements, stock
photos); the second stage generates SQuAD-style~\citep{rajpurkar2016squad} atomic answers tied to
specific visual content. Each pair is emitted in two question forms
(with and without an explicit figure reference) to support both
retrieval-style and reading-comprehension-style framings downstream.

\begin{PromptBlock}
<role>
You generate vision-grounded QA pairs for images extracted from
documents.
</role>

<input>
- IMAGE_INPUT: base-64 PNG of one figure.
- FIGURE_NAME (optional): caption or label from the source document.
</input>

<relevance_check>
SKIP the image (return {"qas": []}) if it is:
- A logo, brand mark, decorative icon, or generic stock photo
- A headshot, watermark, signature, or simple background
- Raw scientific imagery without labels or annotations

PROCEED only for charts, plots, infographics, diagrams, annotated
maps, schematics, screenshots with information, comparison matrices,
or labeled visualizations carrying quantitative or structured content.
</relevance_check>

<task>
For relevant images, produce exactly
{{QA_GENERATION_QUESTIONS_PER_IMAGE}} QA pairs. Each pair contains:
- question:           standard form, no figure reference
- question_reference: same question with a concise figure reference
                      ("Fig. 3", "Profit Function chart"); null if
                      FIGURE_NAME is absent or non-informative
- answer:             atomic SQuAD-style span (typically <=5 words;
                      specific entity, number, color, or short phrase)

Avoid trivial questions (axis labels, headline colors), subjective
interpretation, and any answer that cannot be objectively verified
from the image.
</task>

<output>
{
  "qas": [
    {
      "question": "...",
      "question_reference": "..." | null,
      "answer": "..."
    }
  ]
}
</output>
\end{PromptBlock}

\subsection{Table Q\&A Generation Prompt}

Operates over tables extracted from documents. The model receives the
raw table HTML alongside the surrounding narrative (with table
positions marked by placeholders), enabling questions to disambiguate
via local context such as captions or nearby section headings.
Questions are restricted to lookup and identification - no
arithmetic or aggregation - to keep answer verification
deterministic.

\begin{PromptBlock}
<role>
You generate QA pairs grounded strictly in the provided table.
</role>

<input>
- {{table_number_placeholder}}: zero-indexed table position. Tables
  appear in the surrounding markdown as <!-- TABLE-N --> markers.
- EXTRACTED_TABLE_HTML: the table content as HTML.
- DOCUMENT_WITH_PLACEHOLDER_TABLES: surrounding narrative for context.
</input>

<task>
First check whether the table carries enough information for QA
generation. If it is empty, header-only, or a blank template, return
{"qas": []}. Otherwise produce up to
{{number_of_questions_per_table}} diverse QA pairs.

Each QA pair contains:
- question:           standard form, no table reference
- question_reference: same question with an explicit table reference
                      drawn from the caption, a nearby heading, or
                      distinguishing column/row headers (never the
                      table index)
- answer:             short grounded phrase from a single cell or
                      cell-level lookup
</task>

<rules>
- Lookup or identification only. No arithmetic, aggregation, or
  multi-step reasoning.
- Prefer WH / How Many / Which / What forms over yes/no.
- Do not use phrases like "according to the table" in the standard
  question.
- When cell values repeat, disambiguate via row/column context.
- Each answer must be derivable from the table alone.
</rules>

<output>
{
  "qas": [
    {
      "question": "...",
      "question_reference": "...",
      "answer": "..."
    }
  ]
}
</output>

<input_data>
<|startoftable|>
{{table_markdown}}
<|endoftable|>

<|startoftext|>
{{document_text_only}}
<|endoftext|>
</input_data>
\end{PromptBlock}

\subsection{Word Fidelity Q\&A Generation Prompt}

Produces questions that probe Word-specific fidelity attributes
unanswerable from plain text alone - formatting, metadata,
structural elements, and embedded objects. The prompt receives a
structured fidelity context (a list of typed facts extracted from the
DOCX) and emits typed questions whose answers verify preservation of
those attributes. Type-level priorities counter a known imbalance in
the generated set, which would otherwise skew heavily toward font
questions.

\begin{PromptBlock}
<role>
You generate QA pairs that test Word-specific fidelity attributes
(formatting, metadata, structural elements). Answers derive from
provided fidelity facts; questions must not be answerable from plain
text content alone.
</role>

<task>
Generate exactly {{QUESTIONS_PER_DOCUMENT}} QA pairs from the
fidelity context. Each pair must:
- Depend on a fidelity attribute (not extractable from raw text)
- Reference an unambiguous, identifiable element
- Test a single attribute or a coherent combination of related ones
- Cite the supporting fact_ids
</task>

<question_types>
font, highlight, header, footer, comment, hyperlink, footnote,
equation, textbox, strikeout, layout, metadata.
</question_types>

<priority_balance>
Font questions are overrepresented; cap at 2-3 per document.
Prioritize underserved types when corresponding facts exist:
- equation, comment, strikeout: at least 2 questions if facts exist
- textbox: at least 1 question if facts exist
- header, layout, footnote: prefer over font when facts exist
</priority_balance>

<quality_rules>
1. Subject text must be meaningful. If a fact references placeholder
   content ("______", "..."), describe the element by purpose ("the
   signature line", "the location number field") rather than quoting
   the placeholder.
2. Color answers must be human-readable names, not hex codes
   ("red" not "#FF0000").
3. Skip low-value facts: CUSTOM_PROPERTY entries (SharePoint metadata,
   ContentTypeId) and trivial hyperlinks where the URL matches the
   display text.
4. CORE_PROPERTY metadata (author, title, created date) and
   hyperlinks where the URL meaningfully differs from display text
   are in scope.
5. Comparative or aggregate questions are encouraged when multiple
   related facts support them ("What is the largest font size used
   in the document?").
</quality_rules>

<output>
{
  "qas": [
    {
      "question": "...",
      "answer": "...",
      "question_type": "<one of the types above>",
      "fact_ids": ["FACT_ID_001", "..."]
    }
  ]
}
</output>

<input>
<|startofcontext|>
{{fidelity_context}}
<|endofcontext|>
</input>
\end{PromptBlock}

\section{Per-App File Fidelity Taxonomy}
\label{app:taxonomy}

\S\ref{sec:taxonomy} lists representative object-centric category
labels applicable across applications. This appendix provides the full
per-app artifact taxonomy.
\subsection{Word artifact types}

The Word artifact taxonomy comprises 15 categories spanning textual
content, structural elements, formatting cues, and embedded objects.
Each query is labeled with its primary artifact type; queries
exercising multiple artifacts may be tagged with more than one.

\begin{itemize}
\item \textbf{Table}: tabular content with row/column structure,
including header rows, merged cells, and cell-level values.
\item \textbf{Text}: paragraphs, bullets, numbered lists, and other
free-form body content.
\item \textbf{Font}: typeface, font size, weight, italics, underline,
font color, and other character-level styling.
\item \textbf{Layout}: page layout, column structure, alignment,
indentation, and inter-paragraph spacing.
\item \textbf{Hyperlink}: clickable links to external URLs or
intra-document anchors.
\item \textbf{Metadata}: document properties (title, author, creation
and modification dates, custom properties).
\item \textbf{Footnote}: footnotes and endnotes referenced from body
text.
\item \textbf{Highlight}: text highlighting (yellow, green, etc.) used
to mark or emphasize content.
\item \textbf{Footer}: bottom-of-page repeating content, including
page numbers and footer text.
\item \textbf{Comment}: review comments and annotations attached to
specific text ranges.
\item \textbf{Header}: top-of-page repeating content, including
section titles and document identifiers.
\item \textbf{Equation}: mathematical equations rendered via OMML or
MathML.
\item \textbf{Strikeout}: strikethrough text indicating deletion or
deprecation.
\item \textbf{Textbox}: free-floating text boxes positioned
independently of the main text flow.
\item \textbf{Video}: embedded video objects.
\end{itemize}

\begin{table}[h]
\centering
\small
\caption{File Fidelity query distribution across Word artifact types.}
\label{tab:taxonomy-word}
\begin{tabular}{lr}
\toprule
\textbf{Artifact type} & \textbf{Queries} \\
\midrule
Table        & 143 \\
Text         &  62 \\
Font         &  29 \\
Layout       &  18 \\
Hyperlink    &  18 \\
Metadata     &  17 \\
Footnote     &  16 \\
Highlight    &  14 \\
Footer       &  13 \\
Comment      &  11 \\
Header       &  11 \\
Equation     &  10 \\
Strikeout    &  10 \\
Textbox      &   9 \\
Video        &   2 \\
\midrule
\textbf{Total} & \textbf{383} \\
\bottomrule
\end{tabular}
\end{table}

\subsection{Excel artifact types}

The Excel artifact taxonomy comprises 10 categories covering both
structured spreadsheet objects (formal Tables, PivotTables, Charts) and
the more loosely organized cell ranges that dominate real-world
workbooks.

\begin{itemize}
\item \textbf{TLR (Table-Like Range)}: tabular data laid out across a
contiguous cell range without being formally registered as an Excel
Table object. TLRs are the most common form of structured data in
real-world workbooks and require row/column inference rather than
direct schema lookup.
\item \textbf{Table}: formally structured Excel Table objects with
declared headers, named ranges, and structured references.
\item \textbf{Unstructured data}: free-form cell content that is not
organized in a tabular layout (e.g., scattered values, notes,
free-text cells).
\item \textbf{Formula}: cell formulas, derived values, cross-cell and
cross-sheet references, and named-range expressions.
\item \textbf{Sheet}: worksheet-level structure, including sheet
names, sheet ordering, hidden sheets, and multi-sheet workbooks.
\item \textbf{Chart}: embedded charts, including chart types, data
sources, axes, legends, and data labels.
\item \textbf{Image}: embedded images and screenshots.
\item \textbf{PivotTable}: pivot tables, including their fields,
filters, and aggregations.
\item \textbf{Shape}: shapes positioned over the worksheet grid.
\item \textbf{Other}: residual category for queries that do not fall
cleanly into the categories above.
\end{itemize}

\begin{table}[h]
\centering
\small
\caption{File Fidelity query distribution across Excel artifact types.}
\label{tab:taxonomy-excel}
\begin{tabular}{lr}
\toprule
\textbf{Artifact type} & \textbf{Queries} \\
\midrule
TLR (Table-Like Range) & 133 \\
Table                   &  72 \\
Unstructured data       &  30 \\
Formula                 &  15 \\
Sheet                   &   9 \\
Chart                   &   9 \\
Image                   &   6 \\
PivotTable              &   5 \\
Shape                   &   1 \\
Other                   &   1 \\
\midrule
\textbf{Total}          & \textbf{281} \\
\bottomrule
\end{tabular}
\end{table}

\subsection{PowerPoint artifact types}

The PowerPoint artifact taxonomy comprises 21 categories spanning slide
content, spatial layout, embedded objects, and presentation-level
structure. Compared to Word and Excel, the PowerPoint taxonomy includes
a heavier weighting toward visual and spatial elements (positioning,
shape attributes, color coding) reflecting the format's
presentation-centric nature.

\begin{itemize}
\item \textbf{Slide Text}: text content on slides, including titles,
body text, and bullet points.
\item \textbf{Position / Association / Grouping}: spatial relationships
between slide elements, including grouping, alignment, and
proximity-based associations.
\item \textbf{Tables / Values}: table cell values rendered within
slides.
\item \textbf{Image}: embedded images, screenshots, and photographs.
\item \textbf{Chart}: embedded charts and their data, labels, and
legends.
\item \textbf{Text Highlight}: highlighted text used for emphasis.
\item \textbf{Shape / Text}: text content rendered inside shape
objects (callouts, banners, etc.).
\item \textbf{Equation (or Symbol)}: mathematical equations and
special symbols.
\item \textbf{Hyperlink}: clickable links to external URLs or other
slides.
\item \textbf{Header / Footer / SlideNum / DateTime}: master-level
repeating elements (slide numbers, footers, date/time placeholders).
\item \textbf{Slide Note}: speaker notes attached to slides.
\item \textbf{Language}: language tagging and per-element language
attributes.
\item \textbf{Metadata}: presentation-level properties (title, author,
created/modified dates).
\item \textbf{Shapes / Type \& Color}: shape attributes including
shape type, fill color, and outline.
\item \textbf{Ink}: handwritten ink annotations.
\item \textbf{Tables / Cell color coding}: color-coded table cells
where the color carries semantic meaning (e.g., red/yellow/green
status indicators).
\item \textbf{SmartArt / Type \& Text}: SmartArt diagrams, including
diagram type and embedded text.
\item \textbf{SmartArt / Semantics}: structural relationships encoded
by SmartArt (hierarchy, sequence, cycle).
\item \textbf{Slide Properties - Hidden}: hidden slides not visible
during normal slideshow playback.
\item \textbf{Animations}: animation effects applied to slide
elements.
\item \textbf{Section}: slide-deck sections used to organize slides
into groups.
\end{itemize}

\begin{table*}[t]
\centering
\scriptsize
\setlength{\tabcolsep}{6pt}
\caption{File Fidelity query distribution across PowerPoint artifact types.}
\label{tab:taxonomy-ppt}
\begin{tabular}{l r l r}
\toprule
\textbf{Artifact type} & \textbf{Queries} & \textbf{Artifact type} & \textbf{Queries} \\
\midrule
Slide Text                          & 79 & Language                          & 3 \\
Position / Association / Grouping   & 38 & Metadata                          & 3 \\
Tables / Values                     & 20 & Shapes / Type \& Color            & 3 \\
Image                               & 20 & Ink                               & 3 \\
Chart                               & 13 & Tables / Cell color coding        & 2 \\
Text Highlight                      & 13 & SmartArt / Type \& Text           & 2 \\
Shape / Text                        &  7 & SmartArt / Semantics              & 2 \\
Equation (or Symbol)                &  7 & Slide Properties -- Hidden        & 2 \\
Hyperlink                           &  5 & Animations                        & 2 \\
Header / Footer / SlideNum / DateTime & 5 & Section                           & 2 \\
Slide Note                          &  5 & Video                             & 1 \\
                                    &    & Icons                             & 1 \\
\midrule
\multicolumn{4}{r}{\textbf{Total: 238}} \\
\bottomrule
\end{tabular}
\end{table*}

\section{App-Specific File Fidelity Implementation Details}
\label{app:filefidelity-details}

\S\ref{sec:filefidelity} describes a unified four-stage pipeline applying
to all three applications. This appendix documents the app-specific
implementation choices that the main paper abstracted over.

\subsection{Document representation for LLM ingestion}

\paragraph{Word documents.}
Word documents are converted to Markdown using \texttt{pandoc} with
custom filters tailored to each modality. For text-based questions, we
use standard Markdown conversion. For table-grounded questions, tables
are extracted as standalone HTML files and replaced with placeholders in
the Markdown to enable selective inclusion or exclusion during
groundedness filtering. For visual elements (images and charts), media
files are extracted and referenced via Markdown image syntax, allowing
text-only variants to be constructed by omitting these references. A
combined mode applies both table and visual extraction in a single pass,
producing a unified representation that supports element-level
manipulation for groundedness filtering.

\paragraph{Excel documents.}
Excel files are processed sheet-by-sheet, with each worksheet rendered
into a structured text representation that preserves cell coordinates,
values, and formulas. For each cell, we extract three parallel signals:
the displayed value, the underlying formula (if any), and salient
formatting cues such as number format, fill color, and bold/italic
emphasis where these encode meaning (e.g., totals rows, color-coded
status). Formal Excel Table objects are preserved with their declared
headers and structured references; table-like ranges are detected via
header-row and contiguous-range heuristics and rendered as Markdown
tables. Charts are extracted with their metadata (chart type, source
range, axis titles, legend labels, data labels) alongside a rendered
image of the chart, allowing chart-grounded questions to reference
either the structural metadata or the visual rendering. Embedded
images, shapes, and PivotTables are extracted as separate artifacts
with placeholders in the cell representation. Multi-sheet workbooks
are presented with a leading sheet index followed by per-sheet
content. As with Word, this modular extraction enables selective
inclusion or exclusion of specific elements during groundedness
filtering - e.g., providing the workbook with a target chart
removed, with all formulas replaced by their computed values, or with
non-target sheets stripped.

\paragraph{PowerPoint documents.}
PowerPoint files are processed slide-by-slide, with each slide
represented along three parallel modalities: (i) extracted text
content (titles, body text, bullets, text within shapes, and speaker
notes), (ii) structural metadata (slide layout, shape positions,
grouping relationships, animation order, hidden-slide flags), and
(iii) a rendered image of the slide as it would appear during
playback. Embedded objects - images, charts, tables, SmartArt
diagrams, ink annotations - are additionally extracted as standalone
artifacts so they can be referenced or removed independently of the
slide they appear on. Speaker notes are kept separate from on-slide
text so that questions targeting either modality can be filtered
appropriately. The three-modality representation is essential for
groundedness filtering on PowerPoint: questions targeting visual
layout (positioning, grouping, color coding) require the rendered
image, questions targeting text content require the extracted text,
and questions targeting speaker notes require notes-only inputs.
Master-slide elements (headers, footers, slide numbers, date/time
placeholders) are extracted once at the deck level and referenced from
each slide, mirroring how PowerPoint itself stores them.

\subsection{Word document quality ranking}

To identify high-quality Word documents suitable for benchmark
evaluation, we implement a pairwise document comparison framework with
LLM-based grading. Documents are converted to Markdown for text
representation and compared across five dimensions: (1) structure and
navigation clarity, (2) presence of core document components, (3)
clarity and precision of language, (4) technical cleanliness for
extraction, and (5) real-world authenticity signals. The evaluation
rubric was calibrated through iterative refinement with three human
annotators.

\paragraph{Pair selection.}
Pair selection follows a two-phase strategy. In a bootstrap phase, we
use heuristic selection combining semantically similar (``near'') and
dissimilar (``far'') document pairs to establish initial rankings. In a
rank-aware phase, we generate additional pairs between documents with
adjacent Elo scores to improve discrimination at similar quality levels.

\paragraph{Score aggregation.}
Pairwise outcomes are aggregated into global quality scores using Elo
ratings. To reduce sensitivity to evaluation order, we average scores
across multiple randomized orderings of the comparison sequence. We
select the top-ranked documents per domain for question generation,
ensuring the final set represents diverse, high-quality exemplars of
professional Word documents.

\subsection{Excel and PowerPoint document selection}

Unlike Word, where the large candidate pool and primarily-textual
quality signals enabled a fully automated Elo-based pairwise ranking,
document selection for Excel and PowerPoint was performed manually by
human curators. We attempted to extend the pairwise LLM-grading
approach to both applications and found it did not transfer reliably:

\begin{itemize}
\item \textbf{Excel.} Spreadsheet quality depends heavily on signals
that are difficult to compare via LLM-based pairwise grading over a
text serialization - the interplay between layout, formula
structure, formatting, and multi-sheet organization rarely surfaces
cleanly in the linearized representation. LLM judges produced
inconsistent rankings on Excel pairs, with low inter-run agreement
across randomized orderings. We therefore relied on human curators to
select workbooks based on structural richness (multi-sheet workbooks,
mix of formulas and values, presence of charts and PivotTables),
domain authenticity (real-world filings rather than templates), and
content diversity across the artifact taxonomy.

\item \textbf{PowerPoint.} Presentation quality is a fundamentally
visual property: layout balance, visual hierarchy, color coordination,
and slide-design coherence are not adequately captured by extracted
text alone, and LLM-based comparison over rendered slide images
proved unstable across runs and sensitive to incidental visual
features (image content, color palette) rather than the structural
and design qualities we cared about. Curators manually reviewed
candidate decks for diversity across slide layouts, embedded artifact
types (charts, tables, SmartArt, ink), and content domains.
\end{itemize}

In both cases, manual selection bounded the candidate pool size
relative to Word but ensured the final document set covers the full
artifact taxonomy described in Appendix~\ref{app:taxonomy}. We view
automating quality ranking for spreadsheets and presentations as an
open problem and a useful direction for future work, since manual
curation is the principal scaling bottleneck for extending \OCB{} to
larger Excel and PowerPoint corpora.

\section{Variance Analysis --- Extended Discussion}
\label{app:variance-discussion}

\S\ref{sec:variance} reports the variance decomposition and a
two-finding summary. This appendix expands the discussion with the
points that inform our reporting and budgeting choices for future
benchmark runs.

\paragraph{Scrape variance dominates eval variance.}
Across all three subject models, response-sampling noise is the
dominant source of run-to-run variance: Scrape Var is roughly
$1.6\times$--$3.1\times$ Eval Var, with the ratios spread across
models (Claude $1.6\times$, Gemini $2.0\times$, GPT-5.5 Thinking
$3.1\times$). To our knowledge this decomposition has not been reported for LLM-as-judge pipelines, so we present it as an OCB-specific finding rather than confirmation of an established pattern; we expect the ratio to depend on judge-ensemble size and rubric granularity. It validates the design choice of holding the judge ensemble fixed and instead investing the evaluation budget into multiple response scrapes.

\paragraph{Low eval variance is consistent with high inter-judge agreement.}
The very low Eval Var ($\leq 0.18$ percentage points for all three
models) is consistent with the high inter-judge agreement reported in
\S\ref{sec:judge-agreement}: residual disagreement between GPT-5.4
Thinking, Gemini 3.1 Pro, and Claude Opus 4.6 contributes only
marginally to overall benchmark variance. Future work that wants to
tighten the reported $\pm$ ranges should therefore focus on
increasing the number of response scrapes rather than the number of
judge re-runs.

\paragraph{CIs justify the reported ranking.}
GPT-5.5 Thinking exhibits somewhat higher response-sampling variance
than the other two systems (Scrape Var $0.45$ vs.\ Claude $0.25$ and
Gemini $0.37$), and our scrape-level CI honestly reflects the
uncertainty inherent in only $3$ response samples per model. The
top-two CIs are non-overlapping (GPT-5.5: $[58.24, 60.46]$; Claude:
$[56.22, 57.44]$), with a cushion of $\sim$$0.8$pp between them; we
describe GPT-5.5 Thinking and Claude Opus 4.7 in the main paper as a
co-leading top tier in light of the small number of scrapes. Gemini
3.1 Pro's CI ($[44.84, 46.66]$) is clearly separated from both
top-tier systems by $>$$9$pp.

\section{LLM Judge Agreement}
\label{app:judge-agreement}

We assess the reliability of the three-judge ensemble (GPT-5.4 Thinking,
Gemini 3.1 Pro, and Claude Opus 4.6) by examining agreement on
assertion-level verdicts on Domain Q\&A.

\begin{figure}[ht]
\centering
\includegraphics[width=0.85\linewidth]{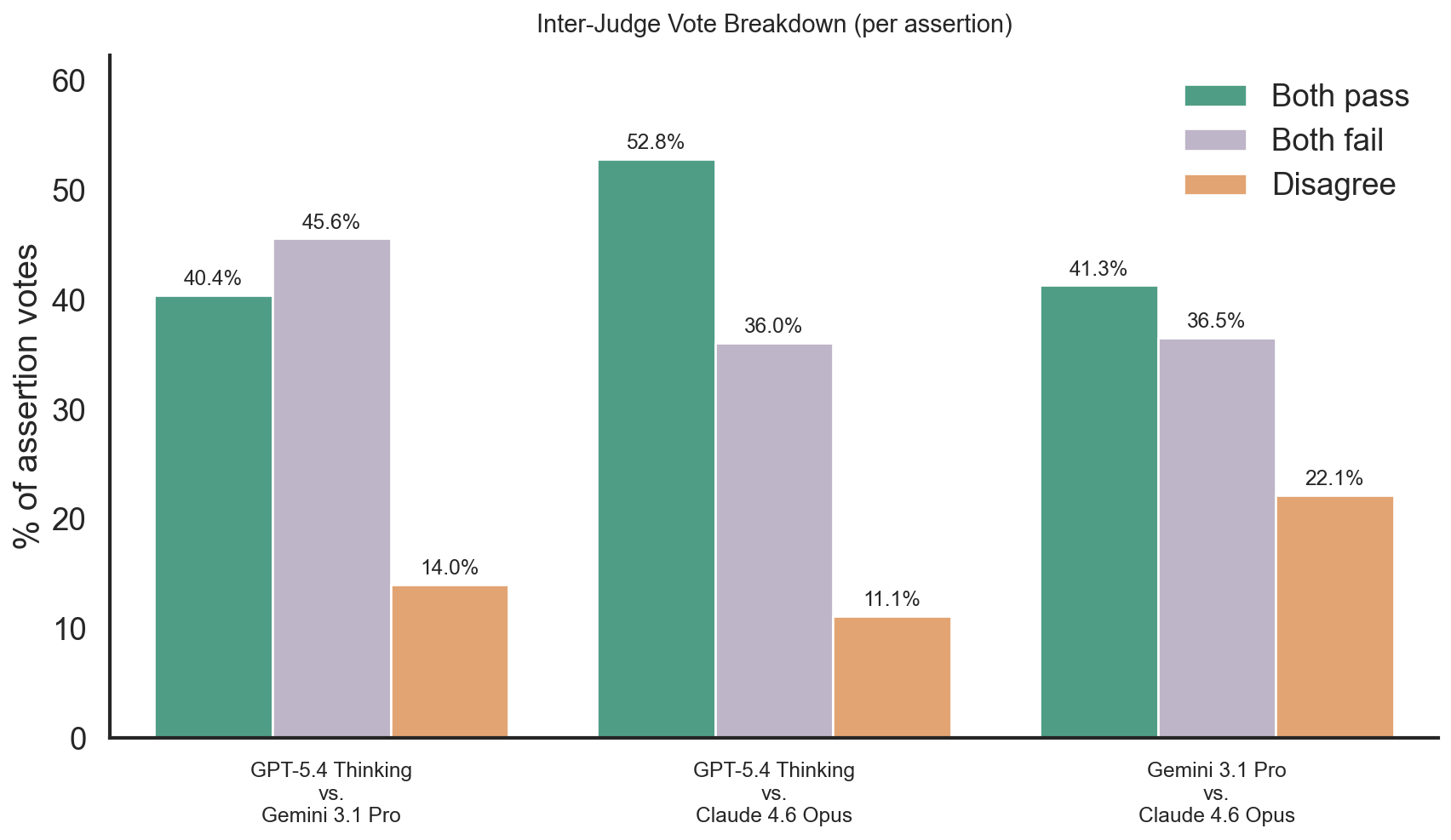}
\caption{Pairwise per-assertion vote breakdown across the three LLM judges.}
\label{fig:judge-breakdown}
\end{figure}

\paragraph{Direction of disagreement.}
Decomposing each pair's outcome into \emph{both pass} / \emph{both fail}
/ \emph{disagree} (Figure~\ref{fig:judge-breakdown}) reveals that
disagreement is not symmetric noise but reflects systematic strictness
differences. GPT-5.4 and Claude~4.6 most often agree to \emph{pass}
($52.8\%$ of judgments), while Gemini 3.1 agrees with Claude on a
\emph{fail} in $36.5\%$ of cases and disagrees with Claude in $22.1\%$,
the largest gap in the ensemble. This pattern is consistent with
Gemini 3.1 being the strictest judge of the three.

The $3$-judge majority vote reduces idiosyncratic strictness bias from
any single model. Combined with the low evaluation-noise variance
reported in \S\ref{sec:variance}, the high pairwise agreement
supports treating the majority verdict as a stable measurement.

\paragraph{Leave-one-out judge ablation.}
To quantify each judge's marginal contribution to the 3-judge majority
vote, we hold out one judge at a time and look at the verdict produced by
the remaining two. A pair is \emph{decisive} when the two remaining
judges agree (the full 3-judge majority must match them); otherwise the
pair \emph{ties} and the held-out judge is the sole tie-breaker. To
characterise how coherent the surviving pair is on its own, we also
report Cohen's $\kappa$ on the same set of assertions.

\begin{table}[h]
\centering
\scriptsize
\setlength{\tabcolsep}{4pt}
\begin{tabularx}{\linewidth}{l *{3}{>{\centering\arraybackslash}X}}
\toprule
\textbf{Held-out judge} &
\textbf{2-judge decisive} & \textbf{2-judge tie} &
\textbf{Cohen $\kappa$} \\
\midrule
Claude 4.6 Opus    & 86.0\% & 14.0\% & 0.724 \\
Gemini 3.1 Pro     & 88.9\% & 11.1\% & 0.773 \\
GPT-5.4 Thinking   & 78.0\% & 22.0\% & 0.579 \\
\bottomrule
\end{tabularx}
\caption{Leave-one-out judge ablation.}
\label{tab:loo-ablation}
\end{table}

\begin{figure}[h]
\centering
\includegraphics[width=0.9\linewidth]{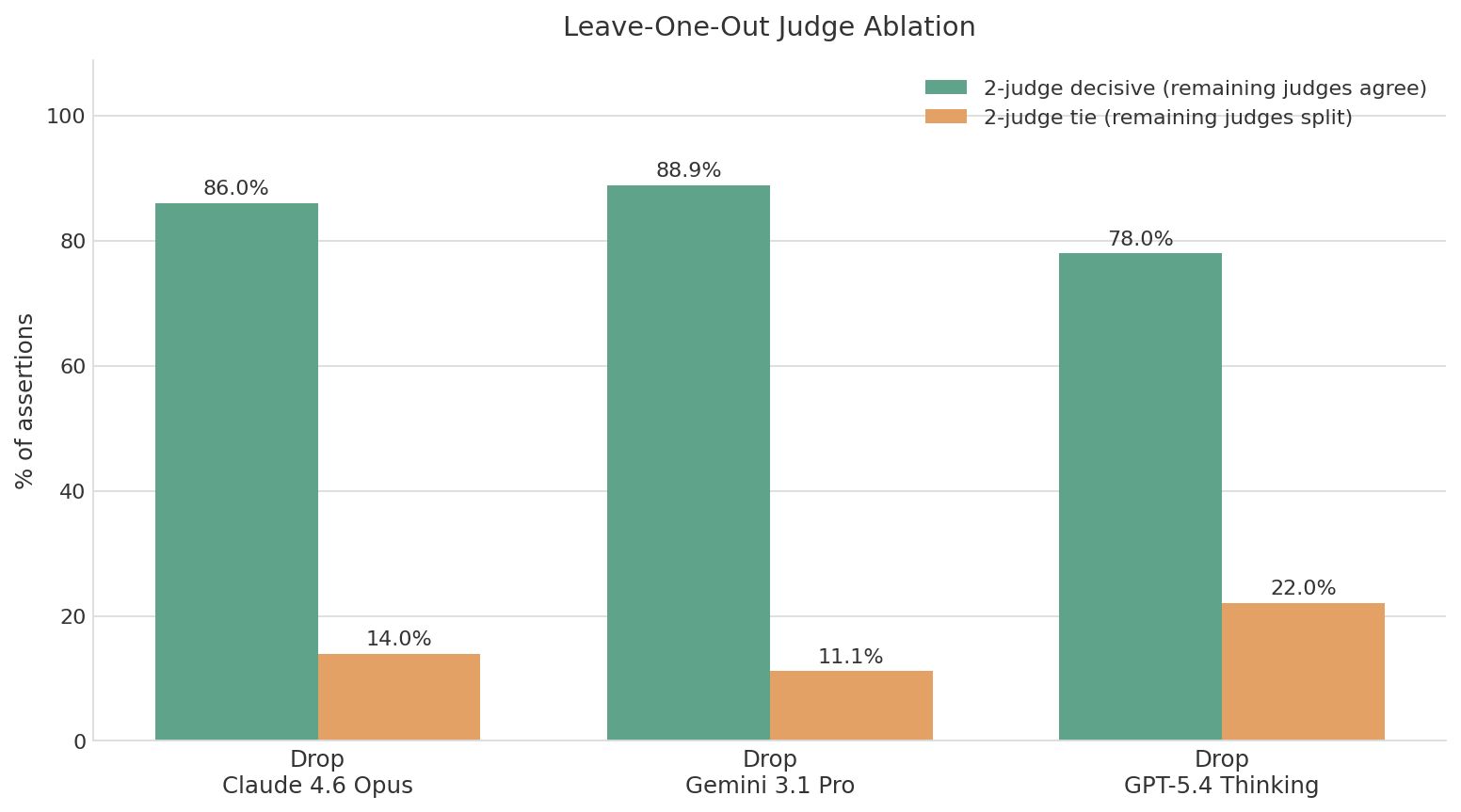}
\caption{Leave-one-out judge ablation}
\label{fig:loo-ablation}
\end{figure}

The analysis yields three main observations:
\begin{itemize}[leftmargin=*]
  \item \textbf{GPT-5.4 Thinking is the most stabilising judge.} Removing
  it leaves Claude and Gemini tied on 22.0\% of assertions and gives the
  weakest pair-level agreement ($\kappa = 0.579$) --- roughly double the
  tie rate observed when either Claude or Gemini is removed, consistent
  with GPT-5.4's central pass rate (53.6\%) sitting between Claude's
  (63.5\%) and Gemini's (41.7\%).
  \item \textbf{Claude + GPT-5.4 is the most coherent 2-judge subset.}
  Dropping Gemini yields the lowest tie rate (11.1\%) and the highest
  chance-corrected agreement ($\kappa = 0.773$), so this pair alone
  already reproduces 88.9\% of the panel's decisions.
  \item \textbf{No single pair dominates.} Even the strongest pair leaves
  $\sim$11\% of assertions undetermined, and the weakest leaves 22\%. The
  third judge therefore contributes a non-trivial deciding vote on a
  meaningful fraction of cases in every configuration.
\end{itemize}

Together these results justify keeping all three judges: every 2-judge
subset shows a materially higher tie rate than the full panel resolves,
and the spread of $\kappa$ values (0.58--0.77) confirms that the three
judges are not redundant.

\section{Query-Level Macro-Average Accuracy}
\label{app:query-level}

The headline metric in \S\ref{sec:eval} is assertion-level (micro-average)
accuracy: every assertion contributes equally, so queries with more
assertions contribute proportionally more to the score. As a robustness
check we also compute a \emph{query-level macro-average}, where each
query's per-assertion pass rate is computed first and the resulting
per-query rates are averaged with equal weight:
\begin{equation}
\mathrm{Acc}_{\mathrm{macro}} = \frac{1}{|Q|} \sum_{q \in Q} \frac{\sum_{i} y_{q,i}}{|A_q|},
\end{equation}
where $|Q|$ is the number of queries and $|A_q|$ is the assertion count
for query $q$. This weighting is more sensitive to short queries (which
contribute less under the assertion-level definition) and dampens the
influence of long queries with many assertions.

Table~\ref{tab:query-level} reports the query-level numbers with the
same scrape-level $95\%$ CI construction used in \S\ref{sec:variance}.
The model ranking is identical to the assertion-level ranking and
all qualitative findings carry over: GPT-5.5 Thinking and Claude
Opus 4.7 form a top tier with non-overlapping CIs, and Gemini 3.1 Pro
is clearly separated. Absolute scores are roughly $3$pp higher than
assertion-level across all three models, indicating that shorter
queries (with fewer assertions) tend to be answered more accurately
than longer ones on average.

\begin{table}[t]
\centering
\scriptsize
\setlength{\tabcolsep}{3pt}
\renewcommand{\arraystretch}{1.1}
\caption{Query-level (macro-average) accuracy and variance decomposition
  on Domain Q\&A ($3$ scrapes $\times$ $3$ evals = $9$ runs per model).
  Variance columns are sample standard deviations in percentage points;
  CI is the $95\%$ scrape-level confidence interval on the mean,
  computed as in \S\ref{sec:variance}.}
\label{tab:query-level}
\resizebox{\linewidth}{!}{%
\begin{tabular}{lccccc}
\toprule
\textbf{Model} & \textbf{Accuracy} & \textbf{Overall Var} & \textbf{Scrape Var} & \textbf{Eval Var} & \textbf{$95\%$ CI ($\pm$)} \\
\midrule
GPT-5.5 Thinking & $62.70\%$ & $0.42$ & $0.45$ & $0.15$ & $\pm 1.13$ \\
Claude Opus 4.7  & $59.97\%$ & $0.32$ & $0.34$ & $0.14$ & $\pm 0.84$ \\
Gemini 3.1 Pro   & $49.21\%$ & $0.80$ & $0.91$ & $0.20$ & $\pm 2.25$ \\
\bottomrule
\end{tabular}%
}
\end{table}

\section{Weighted Domain Q\&A Scores}
\label{app:weighted}
 
\S\ref{sec:domainqa} reports unweighted assertion accuracy as the headline
metric and notes that per-assertion weights are reported in this
appendix as a secondary signal. We do so for transparency: although
weights are not calibrated across raters and are not directly comparable
across rubrics, readers may want to see how scores shift when
expert-assigned importance is taken into account.
 
\paragraph{Weighting scheme.}
Weights are normalized post-hoc as described in \S\ref{sec:domainqa}:
the independent-rubric collection uses a
primary/secondary distinction normalized to 7.5/2.5 per assertion
(treating primary as 3$\times$ more important than secondary); the
other two collections have weights on a 1--10 scale; Within each query, a model's weighted score is
$\sum_i w_i y_{q,i} / \sum_i w_i$, where $w_i$ is the normalized weight
of assertion $i$ and $y_{q,i}$ is the binary judge verdict.
 
\paragraph{Weighted headline results.}
Table~\ref{tab:weighted-headline} reports weighted assertion accuracy
for the three evaluated systems on the Domain Q\&A track, alongside
the unweighted headline scores reported in the main paper
(\S\ref{sec:results}).
 
\begin{table}[h]
\centering
\small
\caption{Weighted vs.\ unweighted assertion accuracy on Domain Q\&A.
Unweighted scores are the headline metric reported in the main paper;
weighted scores apply per-assertion expert importance weights normalized
as described above.}
\label{tab:weighted-headline}
\setlength{\tabcolsep}{3pt}
\resizebox{\linewidth}{!}{%
\begin{tabular}{lcc}
\toprule
\textbf{Model} & \textbf{Unweighted (\%)} & \textbf{Weighted (\%)} \\
\midrule
OpenAI GPT-5.5 Thinking       & 59.3 & 61.0 \\
Anthropic Claude Opus 4.7     & 56.8 & 59.2 \\
Google Gemini 3.1 Pro         & 45.7 & 47.3 \\
\bottomrule
\end{tabular}%
}
\end{table}
 
\begin{figure}[h]
\centering
\includegraphics[width=\linewidth]{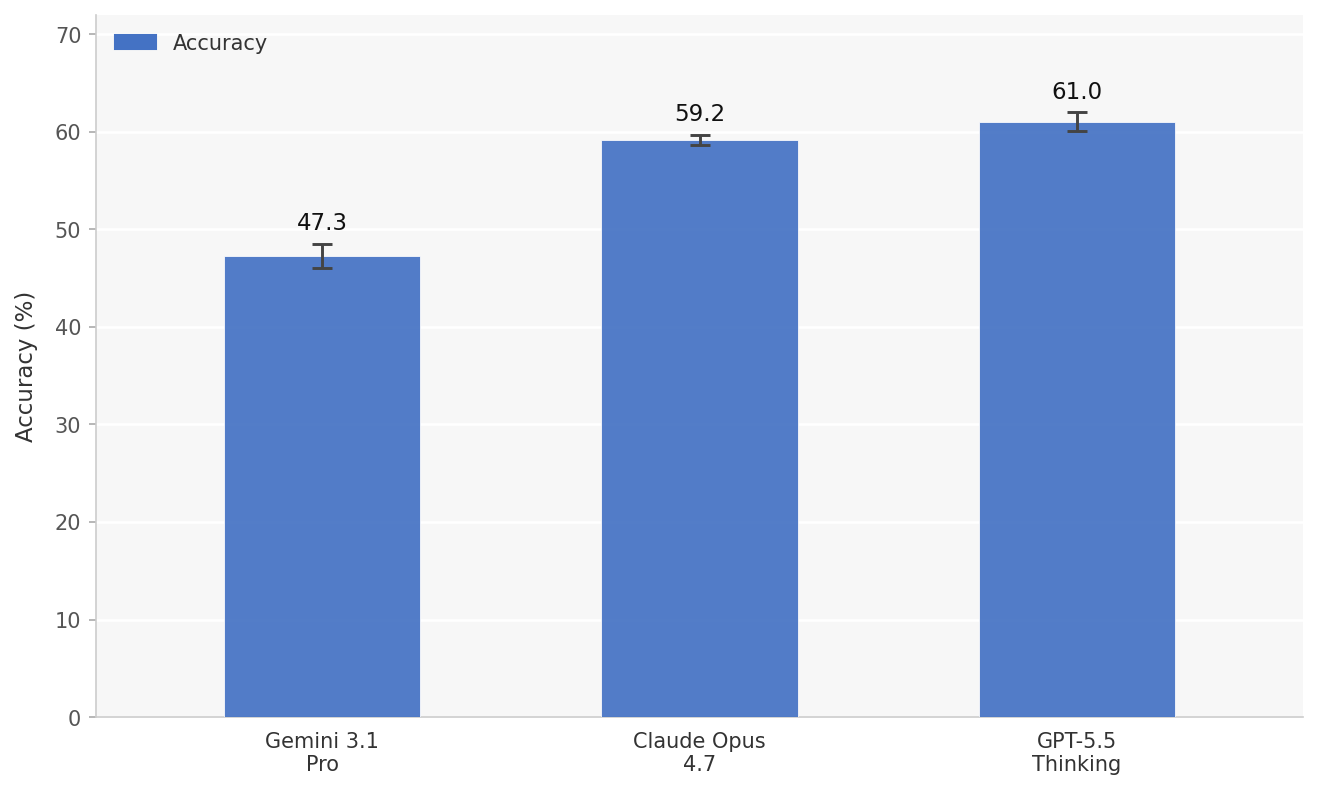}
\caption{Weighted Domain Q\&A accuracy. Equivalent to
Figure~\ref{fig:win-rate-chart} but with per-assertion expert
importance weights applied.}
\label{fig:weighted-headline}
\end{figure}

\paragraph{Comparison of rankings.}
Weighted and unweighted scores agree on model rankings: GPT-5.5
Thinking ranks first, followed by Claude Opus 4.7, with Gemini 3.1 Pro
trailing. The unweighted-to-weighted gain is uneven across systems,
$+1.7$pp for GPT-5.5 Thinking, $+2.4$pp for Claude Opus 4.7, and
$+1.6$pp for Gemini 3.1 Pro, so weighting modestly narrows the
top-two gap from $2.5$pp under unweighted scoring (59.3\% vs.\ 56.8\%)
to $1.8$pp under weighted scoring (61.0\% vs.\ 59.2\%). Expert-assigned
weights therefore favor Claude slightly relative to the per-assertion
average, but not enough to change the headline ordering.

\begin{figure}[h]
\centering
\includegraphics[width=\linewidth]{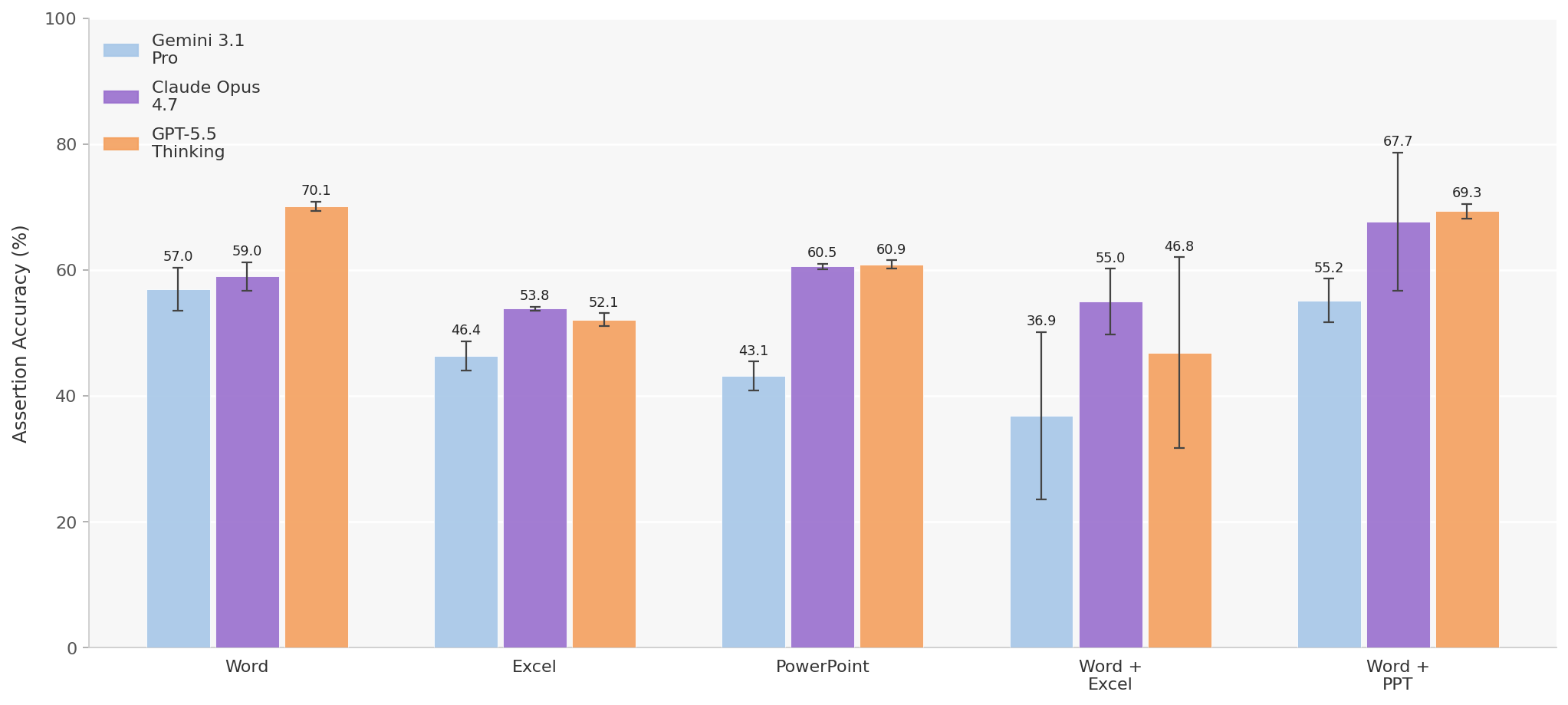}
\caption{Weighted Domain Q\&A accuracy by file type. Equivalent to
Figure~\ref{fig:filetypes} but with per-assertion expert importance
weights applied.}
\label{fig:weighted-filetypes}
\end{figure}

\paragraph{Weighted breakdown by file type.}
Figure~\ref{fig:weighted-filetypes} shows weighted accuracy across the
five file-type buckets. Per-file-type leaders are identical to the
unweighted picture in \S\ref{sec:domain-analysis}: GPT-5.5 Thinking
leads on Word (70.1\%) and Word$+$PPT (69.3\%); Claude Opus 4.7
leads on Excel (53.8\%) and the mixed Word$+$Excel input (55.0\%);
GPT-5.5 Thinking marginally leads on PowerPoint (60.9\% vs.\ 60.5\%).
Weighting shifts per-bucket point estimates by under $3$pp across the
board, and the Claude-vs-GPT gap on Word$+$Excel widens slightly from
$\sim$$7$pp under unweighted scoring to $\sim$$8$pp under weighted
scoring; otherwise weighting does not change the cross-system story
on any file-type bucket.

\begin{figure}[h]
\centering
\includegraphics[width=\linewidth]{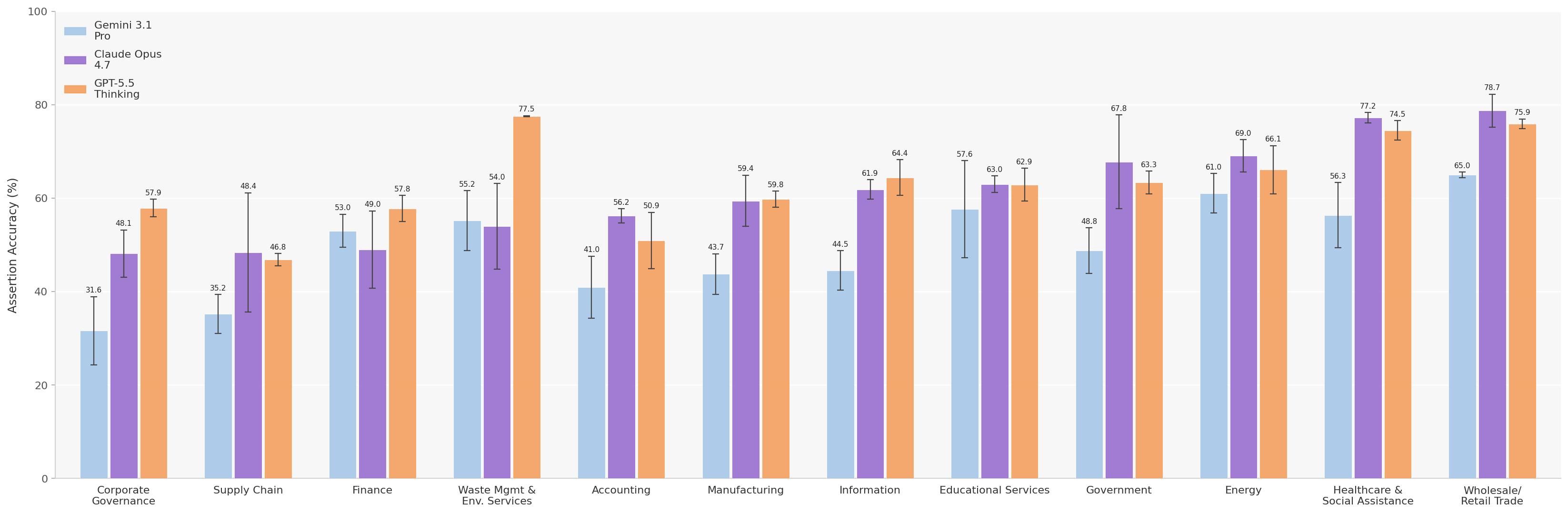}
\caption{Weighted Domain Q\&A accuracy by industry. Equivalent to
Figure~\ref{fig:industries} but with per-assertion expert importance
weights applied.}
\label{fig:weighted-industries}
\end{figure}

\paragraph{Weighted breakdown by industry.}
Figure~\ref{fig:weighted-industries} shows weighted accuracy across
the 12 industries. The broad sector-difficulty pattern is preserved,
Healthcare \& Social Assistance and Wholesale/Retail Trade remain
the easiest sectors across systems, Corporate Governance, Supply
Chain and Accounting remain the hardest. Per-industry leadership between the top
two systems is also largely preserved: GPT-5.5 Thinking leads in $5$
of $12$ sectors under weighted scoring vs.\ $6$ unweighted, and
Claude Opus 4.7 leads in $7$ vs.\ $6$. The single sector that flips
from GPT-5.5 to Claude under weighting (Educational Services,
$60.5\% \rightarrow 62.9\%$) had a within-$2$pp margin under
unweighted scoring, so the swap is consistent with weighting acting
as a small re-ordering of assertions of comparable systems rather
than a substantive shift. Gemini 3.1 Pro remains the per-industry
leader in zero sectors under either scheme.

\paragraph{Caveat: rater-level weight calibration.}
As noted in \S\ref{sec:domainqa}, different experts may apply importance weights with different levels of strictness. For example, some raters assign high importance to many assertions or mark most assertions as primary, while others reserve high weights or primary status only for the most essential claims. Since importance weights were not calibrated across raters during collection, weighted scores should be interpreted as meaningful within a given rubric or slice, rather than as fully absolute signals across the entire benchmark. Readers comparing weighted scores across \OCB{} slices should account for this potential rater-level variation. The per-industry leader shift discussed above, in particular, may partly reflect rater-level weighting practice rather than a model capability difference.

\section{Test-Time Compute Ablations --- Numerical Results}

Table~\ref{tab:ablation-accuracy} reports the full numerical results behind the
test-time compute ablation figures in \S\ref{sec:ablations}.

\label{app:ablation-table}

\begin{table*}[t]
\centering
\small
\setlength{\tabcolsep}{4pt}
\caption{Test-time compute ablations on Domain Q\&A. The two
  Standard-Thinking rows use the $3$-scrape $\times$ $3$-eval protocol
  ($9$ runs per row); all other rows use the leaner $3$-scrape
  $\times$ $1$-eval protocol ($3$ runs per row), as justified in
  \S\ref{sec:ablations}. All accuracy CIs are scrape-level $95\%$
  intervals computed as $t_{0.025,\,2}\cdot\sigma_{\mathrm{scrape}}/\sqrt{3}$
  with $t_{0.025,\,2}=4.303$. Per-query response time is the mean
  across $\sim$120 queries pooled over $3$ scrapes, with the $95\%$ CI
  computed across scrape means.}
\label{tab:ablation-accuracy}
\begin{tabular}{llcccc}
\toprule
\textbf{Model} & \textbf{Mode} & \textbf{Accuracy} & \textbf{Acc. CI ($\pm$)} & \textbf{Time (s)} & \textbf{Time CI ($\pm$)} \\
\midrule
Claude Opus 4.7 & Standard-Thinking  & $56.83\%$ & $\pm 0.61$ & $98.0$  & $\pm 13.8$ \\
Claude Opus 4.7 & Adaptive-Thinking  & $58.93\%$ & $\pm 2.05$ & $138.6$ & $\pm 6.4$  \\
\midrule
GPT-5.5 & Thinking-Light    & $59.11\%$ & $\pm 3.07$ & $53.0$  & $\pm 18.9$ \\
GPT-5.5 & Thinking-Standard & $59.35\%$ & $\pm 1.11$ & $62.3$  & $\pm 4.3$  \\
GPT-5.5 & Thinking-Extended & $59.79\%$ & $\pm 0.84$ & $91.3$  & $\pm 22.4$ \\
GPT-5.5 & Thinking-Heavy    & $59.81\%$ & $\pm 0.77$ & $105.3$ & $\pm 21.4$ \\
GPT-5.5 & Pro-Standard      & $61.55\%$ & $\pm 0.79$ & $239.6$ & $\pm 25.1$ \\
GPT-5.5 & Pro-Extended      & $\mathbf{63.09\%}$ & $\pm 3.57$ & $358.3$ & $\pm 50.0$ \\
\bottomrule
\end{tabular}
\end{table*}

\section{Human-Annotator Baseline on File Fidelity}
\label{app:human-baseline}

To anchor the File Fidelity headline numbers against a human reference,
we collected accuracy from human annotators on a substantial
subset of File Fidelity Q\&A queries. The dashed horizontal lines in
Figure~\ref{fig:file-fidelity-accuracy-chart} (\S\ref{sec:results})
report the per-application annotator accuracy: $78.6\%$ on Word,
$68.3\%$ on PowerPoint, and $45.9\%$ on Excel.

\paragraph{Annotator pool.}
Annotators are familiar with general office literacy (proficient with
Word, Excel, and PowerPoint at a working-professional level) rather
than specialized domain expertise. This is important context for
interpreting the Excel result, where many File Fidelity queries
require quantitative reasoning (formula tracing, cross-sheet
aggregations, computing aggregates over filtered ranges) that
domain-matched annotators would likely answer more accurately. The
Word and PowerPoint annotator baselines are correspondingly more
informative of an attainable single-rater ceiling than the Excel
baseline.

\paragraph{Coverage.}
Annotators answered a substantial subset of File Fidelity queries per
application: $96.6\%$ of PowerPoint queries, $91.9\%$ of Word queries,
and $76\%$ of Excel queries. The unanswered fraction reflects queries
that were added to the benchmark after the annotation pass completed,
not selective exclusion of harder items.

\paragraph{Setup.}
Annotators received the file and the question with the same single-pass
setup used for the LLM systems: the source file is provided, the
question is asked once, and the annotator produces a free-form
response. Each query was answered by one annotator. Annotator
responses were scored against the same atomic assertion rubrics used
for the LLM responses, so the human-annotator and LLM-system numbers
are directly comparable on a query-by-query, assertion-by-assertion
basis.

\paragraph{Caveats.}
The baseline has two methodological limitations, also acknowledged in
the main paper's Limitations section. First, with one rater per query,
no inter-rater agreement is measured; reported accuracy reflects a
single rater's judgment and does not bound the variance of a
multi-rater consensus. Second, the annotator pool was selected for
general office literacy rather than specialized expertise, which
particularly affects categories requiring numerical or analytical
reasoning. A larger multi-rater baseline with domain-matched
annotators is left to future work.

\section{File Fidelity Q\&A --- Per-App Performance Breakdowns}
\label{app:filefidelity-breakdowns}

This appendix expands \S\ref{sec:per-app} with the full per-application
breakdowns: accuracy decomposed across the artifact-type taxonomy of
\S\ref{app:taxonomy} and accuracy as a function of file size, separately
for each of Word, Excel, and PowerPoint. Size buckets are defined per-app
using the natural document unit, pages for Word, worksheets for Excel,
and slides for PowerPoint, with thresholds stated in each subsection
below.

\subsection{Word}

\paragraph{Accuracy by feature.}
Figure~\ref{fig:word-feature} shows per-artifact accuracy for the 15
Word artifact categories defined in Appendix~\ref{app:taxonomy}.
Claude Opus 4.7 dominates across nearly every category, scoring 100\%
on Comment, Equation, Hyperlink, and Text, and remaining above 80\% on
Table, Font, Textbox, Layout, Strikeout, Footnote, and Footer.
GPT-5.5 Thinking is competitive on text-centric artifacts (Text 100\%,
Table 96\%, Footnote 89\%, Comment 75\%) but drops sharply on
formatting and structural artifacts (Hyperlink 6\%, Font 17\%, Header
17\%, Textbox 27\%, Layout 29\%, Footer 36\%, Metadata 6\%). Gemini
3.1 Pro shows the most uneven profile: strong on plain text and tables
(Text 96\%, Table 90\%, Equation 83\%) but near-zero on structural and
metadata artifacts (Comment 8\%, Hyperlink 11\%, Font 14\%, Layout
10\%, Header 8\%, Footer 0\%, Highlight 8\%, Metadata 0\%). The
divergence on structural artifacts is the dominant source of the
overall Word performance gap between Claude Opus 4.7 and the other two
systems.

\begin{figure}[tbp]
\centering
\includegraphics[width=\linewidth]{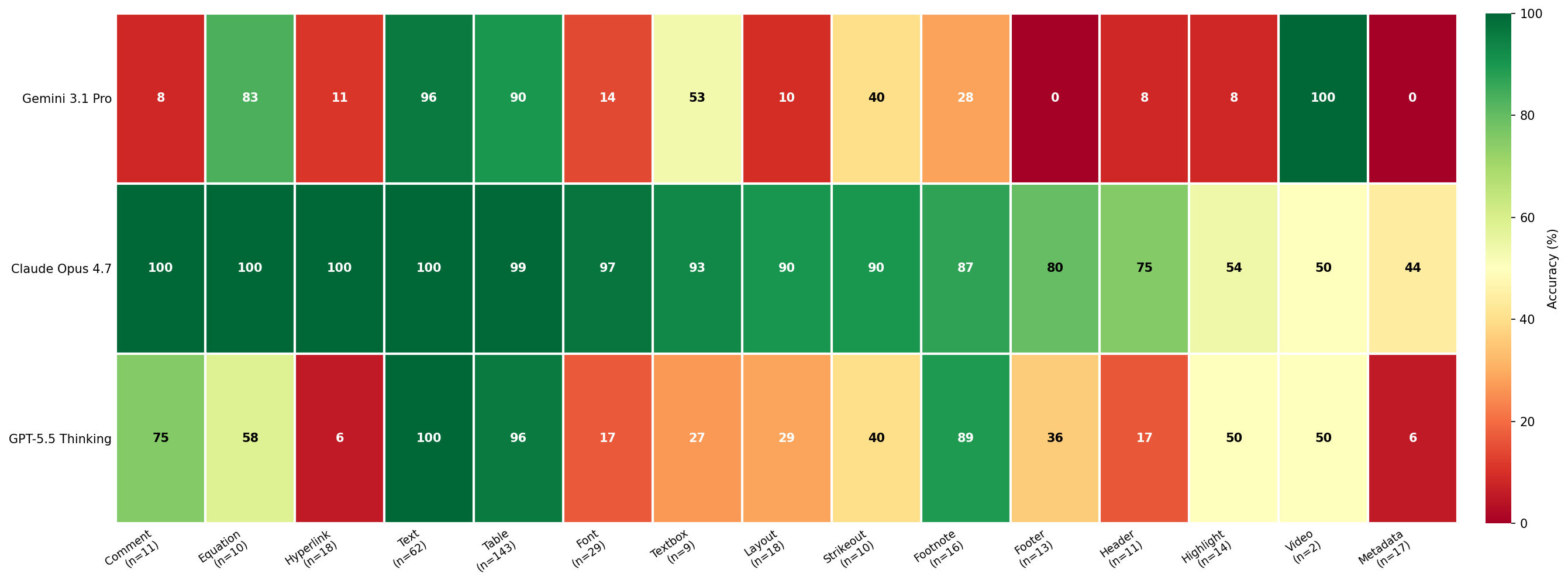}
\caption{Word File Fidelity accuracy by artifact type. Cell counts ($n$)
indicate the number of queries per category.}
\label{fig:word-feature}
\end{figure}

\begin{figure}[tbp]
\centering
\includegraphics[width=\linewidth]{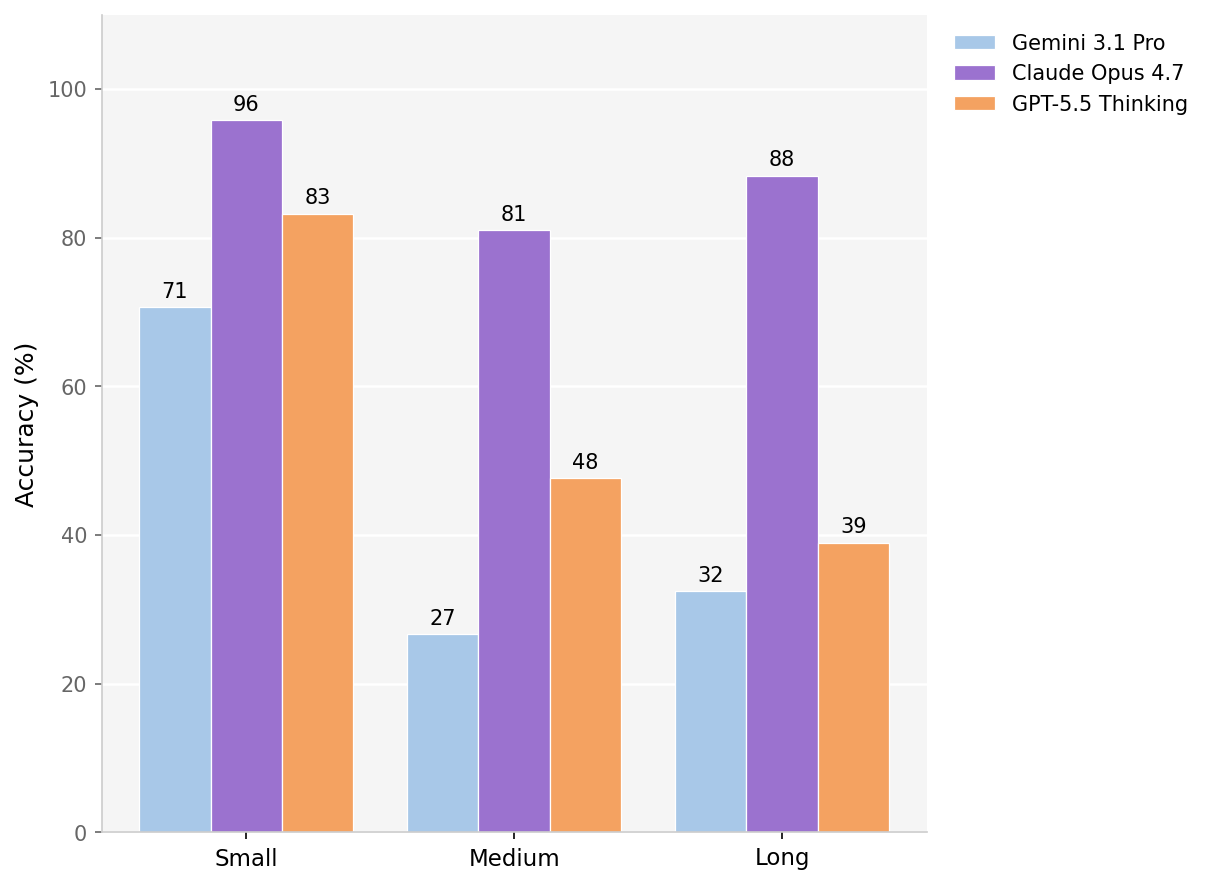}
\caption{Word File Fidelity accuracy by document size.}
\label{fig:word-size}
\end{figure}

\paragraph{Accuracy by file size.}
Documents are bucketed by page count: Small ($\leq$10 pages),
Medium (11--30 pages), and Long ($>$30 pages).
Figure~\ref{fig:word-size} shows accuracy stratified by document length
(Small / Medium / Long). Claude Opus 4.7 maintains high accuracy
across all three buckets (96\% / 81\% / 88\%), with only modest
degradation on medium documents. GPT-5.5 Thinking and Gemini 3.1 Pro
both show a pronounced cliff between Small and Medium/Long: GPT-5.5
drops from 83\% on Small to 48\% / 39\% on Medium / Long, and Gemini
3.1 Pro drops from 71\% on Small to 27\% / 32\%. Long-document
robustness is therefore a clear differentiator on the Word track.

\subsection{Excel}
\begin{figure}[tbp]
\centering
\includegraphics[width=\linewidth]{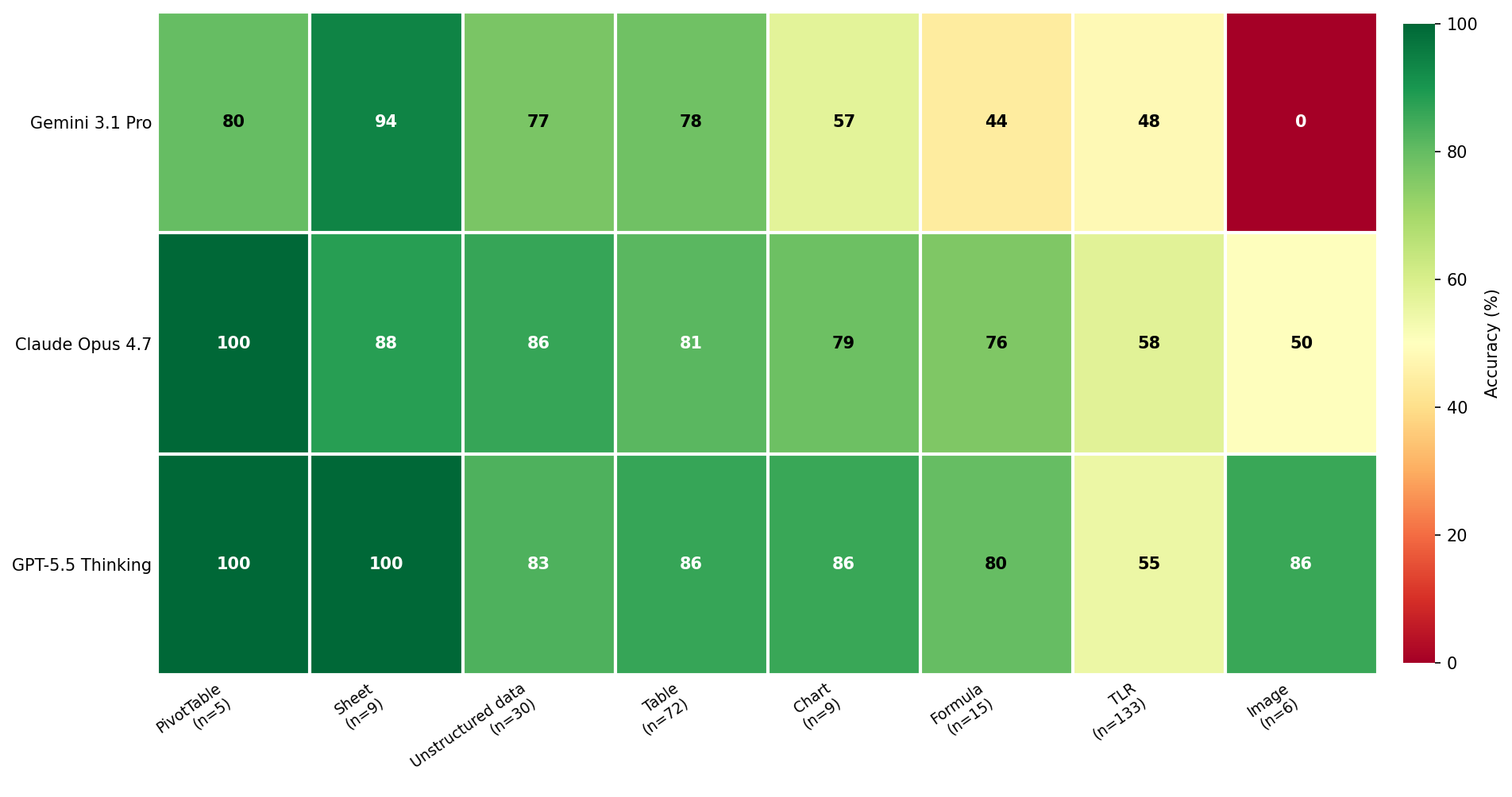}
\caption{Excel File Fidelity accuracy by artifact type.}
\label{fig:excel-feature}
\end{figure}

\paragraph{Accuracy by feature.}
Figure~\ref{fig:excel-feature} shows per-artifact accuracy across the
Excel taxonomy. The three systems perform comparably on the structured
spreadsheet objects, formal Tables (Gemini 78\%, Claude 81\%, GPT-5.5
86\%) and Sheets (94\% / 88\% / 100\%). Claude and GPT-5.5 reach 100\%
on PivotTables, while Gemini reaches 80\% ($n{=}5$, so these should be
read with caution). Differences emerge on formula-grounded queries,
where Claude Opus 4.7 (76\%) and GPT-5.5 Thinking (80\%) substantially
outperform Gemini 3.1 Pro (44\%), and on embedded images, where Gemini
3.1 Pro scores 0\% versus 50\% (Claude) and 86\% (GPT-5.5).
Table-Like Ranges (TLR), the largest category in the Excel benchmark
with 133 queries, remains the hardest artifact for all systems,
Gemini 48\%, GPT-5.5 55\%, Claude 58\%, reflecting the difficulty
of inferring row/column structure from contiguous cell ranges that
are not formally registered as Excel Tables.

\begin{figure}[tbp]
\centering
\includegraphics[width=\linewidth]{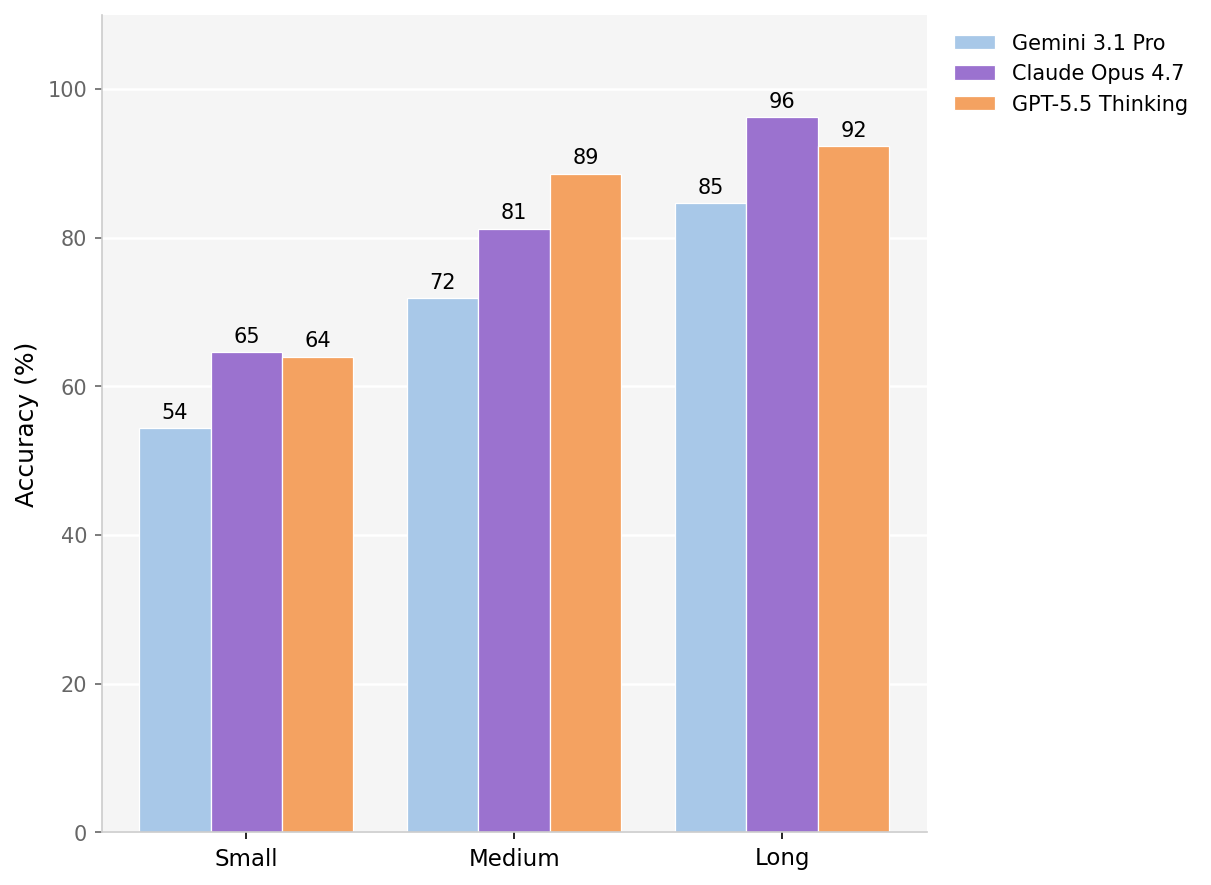}
\caption{Excel File Fidelity accuracy by workbook size.}
\label{fig:excel-size}
\end{figure}

\paragraph{Accuracy by file size.}
Workbooks are bucketed by worksheet count: Small (1--3 sheets),
Medium (4--8 sheets), and Long ($>$8 sheets). The Long bucket is
dominated by a single 87-sheet workbook contributing 12 of the
queries in that bucket, so per-system Long-bucket scores should be
interpreted as a small-sample signal rather than a robust estimate.
Figure~\ref{fig:excel-size} shows accuracy by workbook size. From
Small to Medium, all three systems improve markedly, Claude
Opus 4.7 from 65\% to 81\%, GPT-5.5 Thinking from 64\% to 89\%,
and Gemini 3.1 Pro from 54\% to 72\%. We attribute this Small-to-Medium
gain to a composition shift rather than a true size effect:
Small-bucket workbooks are dominated by sparse, loosely structured
layouts heavy in TLR and unstructured content, whereas Medium-bucket
workbooks more often contain formal Tables, named Sheets, and
well-defined formulas, which all three models handle more reliably.
The Long-bucket numbers (Claude 96\%, GPT-5.5 92\%, Gemini 85\%)
extend the apparent trend but, as noted above, reflect a single
87-sheet workbook with 12 queries; we therefore treat them as a
single-document signal rather than evidence of a continued size
effect, and we caution against reading Medium-to-Long as a robust
trajectory.

\subsection{PowerPoint}

\begin{figure}[tbp]
\centering
\includegraphics[width=\linewidth]{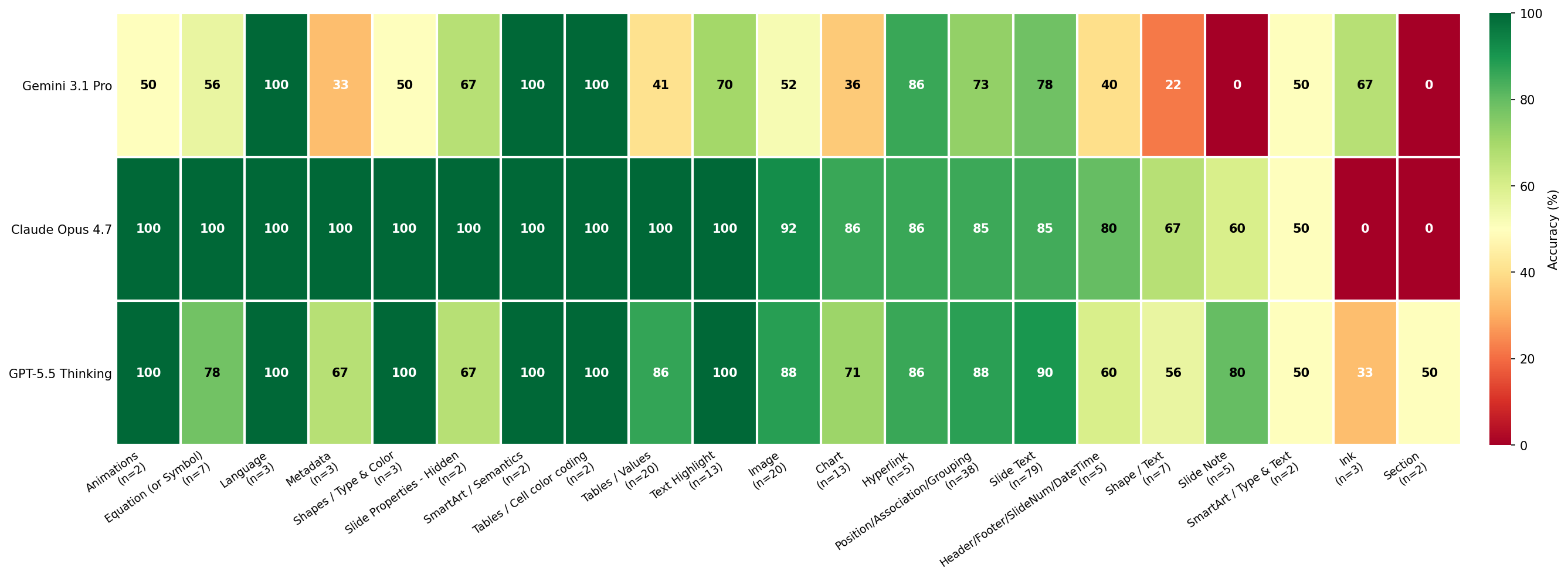}
\caption{PowerPoint File Fidelity accuracy by artifact type.}
\label{fig:ppt-feature}
\end{figure}

\paragraph{Accuracy by feature.}
Figure~\ref{fig:ppt-feature} shows per-artifact accuracy across the 21
PowerPoint categories defined in Appendix~\ref{app:taxonomy}. Claude
Opus 4.7 is the strongest system on PowerPoint, scoring 100\% on
Animations, Equation, Language, Metadata, Shapes / Type \& Color,
Slide Properties, Hidden, SmartArt / Semantics, Tables / Cell color
coding, Tables / Values, and Text Highlight, and remaining above 80\%
on the bulk of the remaining categories (Image 92\%, Chart 86\%,
Hyperlink 86\%, Position/Association/Grouping 85\%, Slide Text 85\%,
Header/Footer/SlideNum/DateTime 80\%). GPT-5.5 Thinking is close behind
on text and tabular artifacts but drops on Metadata (67\%),
Header/Footer/SlideNum/DateTime (60\%), Shape/Text (56\%), and Ink
(33\%). Gemini 3.1 Pro again shows the widest spread, with strong
results on Language (100\%), SmartArt / Semantics (100\%), Tables /
Cell color coding (100\%), and Hyperlink (86\%), but near-zero on
Slide Note (0\%) and Section (0\%), and weak performance on
Shape/Text (22\%), Metadata (33\%), Chart (36\%), and
Header/Footer/SlideNum/DateTime (40\%). Many of the smallest
categories ($n=2$ or $n=3$) should be interpreted with caution, but
the broad pattern, Claude $>$ GPT-5.5 $>$ Gemini, with the gap
widening on app-specific structural artifacts, is consistent.

\begin{figure}[tbp]
\centering
\includegraphics[width=0.78\columnwidth]{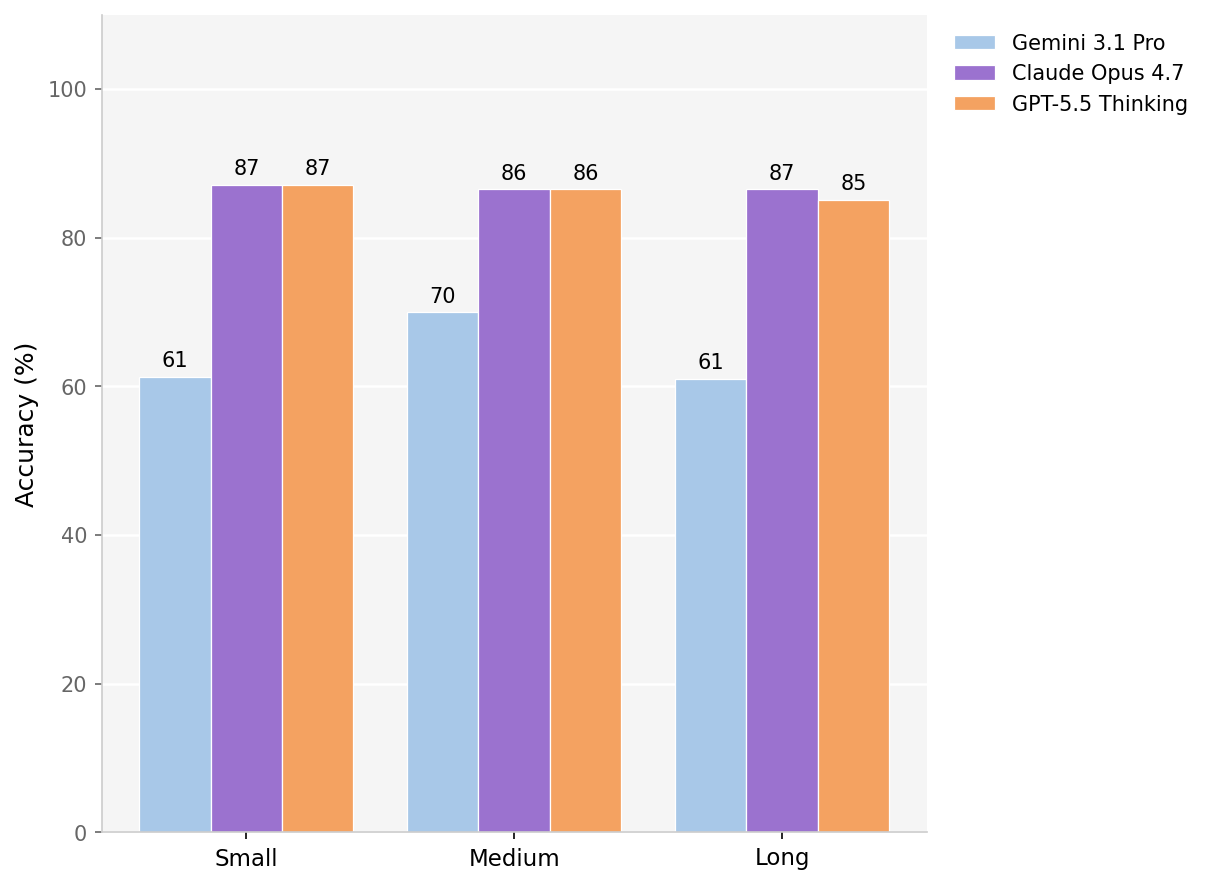}
\vspace{-0.4em}
\caption{PowerPoint File Fidelity accuracy by deck size.}
\label{fig:ppt-size}
\end{figure}

\begin{table*}[t]
\centering
\scriptsize
\caption{Differentiation across document-comprehension benchmarks. The columns capture the four design dimensions where prior work falls short of integrated office comprehension: input fidelity, application coverage, task type, and evaluation methodology.}
\label{tab:related}
\renewcommand{\arraystretch}{1.1}
\setlength{\tabcolsep}{4pt}
\resizebox{\textwidth}{!}{%
\begin{tabular}{@{}p{\dimexpr0.2344\textwidth-6.4pt\relax}p{\dimexpr0.1875\textwidth-6.4pt\relax}p{\dimexpr0.1797\textwidth-6.4pt\relax}p{\dimexpr0.1953\textwidth-6.4pt\relax}p{\dimexpr0.2031\textwidth-6.4pt\relax}@{}}
\toprule
\textbf{Benchmark} & \textbf{Input format} & \textbf{Apps covered} & \textbf{Task type} & \textbf{Evaluation} \\
\midrule
DocVQA~\citep{mathew2021docvqa}           & Document images           & Generic (industry docs)   & Extractive QA              & ANLS \\
ChartQA~\citep{masry2022chartqa}          & Chart images              & Charts only               & Visual + logical QA        & Exact match \\
SlideVQA~\citep{tanaka2023slidevqa}       & Slide images (rendered)   & Slides only               & Multi-hop, numerical QA    & Exact match \\
MP-DocVQA~\citep{tito2023mpdocvqa}        & PDF (rendered, $\le$20pp) & Generic (industry docs)   & Extractive QA + page ID    & ANLS + page-ID accuracy \\
DUDE~\citep{vanlandeghem2023dude}         & PDF (rendered, multi-pg)  & Generic, multi-industry   & Mixed extractive/abstractive & ANLS / ANLS$^{*}$~\citep{peer2024anlsstar} \\
MMLongBench-Doc~\citep{ma2024mmlongbenchdoc} & PDF (rendered, $\sim$50pp) & Generic long PDFs      & Multi-modal QA             & GPT-4 judge \\
DocBench~\citep{zou2024docbench}          & PDF (rendered)            & Academia / finance / govt / law / news & End-to-end document QA & GPT-4 judge \\
MMDocBench~\citep{zhu2024mmdocbench}      & Document images           & Generic (multi-domain)    & Fine-grained QA + bbox grounding & Accuracy + IoU \\
LongDocURL~\citep{deng2024longdocurl}     & PDF (rendered, multi-pg)  & Generic long PDFs         & Understand / reason / locate & Per-task accuracy + locating accuracy \\
FinQA / TAT-QA~\citep{chen2021finqa,zhu2021tatqa} & PDF excerpts + text & Finance only         & Numerical QA + program     & Exact match on derivations \\
FinanceBench~\citep{islam2023financebench} & PDF (10-K, 10-Q)          & Finance only              & Open-book QA               & Manual correctness w/ hallucination flags \\
SpreadsheetBench~\citep{ma2024spreadsheetbench} & Native \texttt{.xlsx} & Excel only           & Manipulation (code-gen)    & Online-judge test cases \\
OfficeBench~\citep{wang2024officebench}   & Office app environment    & Word + Excel + PDF + email/shell & Agent task completion & Pass / fail per task \\
OdysseyBench~\citep{wang2025odysseybench}   & Office app environment    & Word + Excel + PDF + email + calendar        & Long-horizon agent workflows & Task success \\
GDPval~\citep{patwardhan2025gdpval}       & Mixed work products       & Cross-occupational Word + Excel + PPT + other deliverables        & Work-product creation      & Human head-to-head \\
OfficeQA Pro~\citep{databricks2026officeqapro} & PDF (Treasury Bulletins) & Finance / govt only & Numeric multi-doc QA       & Exact match w/ tolerance \\
GDP.PDF~\citep{garre2026gdppdfbenchmarkinggroundedmultimodal}         & PDF (rendered, professional) & Cross-occupational PDFs & Grounded multimodal reasoning & Rubric mean + strict pass rate \\
\midrule
\textbf{\OCB{} (ours)} & \textbf{Native \texttt{.docx}, \texttt{.xlsx}, \texttt{.pptx} and its variants} & \textbf{Word + Excel + PPT + cross-occupational} & \textbf{Perception + analytical reasoning} & \textbf{Atomic assertions, 3-judge majority over multi-paragraph responses} \\
\bottomrule
\end{tabular}%
}
\end{table*}

\paragraph{Accuracy by file size.}
Decks are bucketed by slide count: Small ($\leq$10 slides),
Medium (11--30 slides), and Long ($>$30 slides).
Figure~\ref{fig:ppt-size} shows accuracy stratified by deck size.
Claude Opus 4.7 and GPT-5.5 Thinking are essentially flat across Small,
Medium, and Long decks (Claude: 87\% / 86\% / 87\%; GPT-5.5: 87\% /
86\% / 85\%), indicating that PowerPoint comprehension for these
systems is not strongly limited by deck length. Gemini 3.1 Pro is
weaker overall (61\% / 70\% / 61\%) and shows a non-monotonic pattern,
peaking on Medium decks.


\section{Differentiation from Other Document-Comprehension Benchmarks}
Table~\ref{tab:related} summarizes the difference between our benchmark and other existing benchmark, showcasing the gaps that we are attempting to fill.

\end{document}